%% file: main_new.tex
\documentclass[preprint,12pt,authoryear]{elsarticle}
\usepackage{lineno}

\usepackage{amsmath}
\usepackage{amsfonts}
\usepackage{enumitem}
\usepackage{amsopn}
\usepackage{bm}
\usepackage{listings}
\usepackage[ruled,vlined]{algorithm2e}
\usepackage{color}
\usepackage{url}
\usepackage{hyperref}
\usepackage{wrapfig}
\usepackage[T1]{fontenc}
\usepackage{adjustbox}
\usepackage{flushend} 
\hypersetup{pdfborder=0 0 0}
\usepackage{mathrsfs,amsmath}
\usepackage[all,cmtip]{xy}
\usepackage{graphicx}
\usepackage{multirow}
\usepackage{tabularx}
\usepackage{longtable}
\usepackage{lscape}
\usepackage{booktabs}
\usepackage{multirow}
\usepackage{rotating}
\usepackage[table]{xcolor}
\definecolor{nice-green}{HTML}{007849}
\definecolor{nice-blue}{HTML}{0375B4}
\definecolor{nice-orange}{HTML}{CC7722}
\definecolor{nice-red}{HTML}{FF5733}
\usepackage{bbding}
\usepackage{pdflscape}
\usepackage{makecell}
\usepackage{tikz}
\usetikzlibrary{positioning, arrows.meta}
\usepackage{amssymb}
\usepackage{bbm}
\usepackage{pifont}%
\usepackage{tcolorbox}
\usepackage{float}
\usepackage[normalem]{ulem}

\newcommand{\X}{\mathbf{X}}
\newcommand{\Z}{\mathbf{Z}}

\newcommand{\M}{\mathbf{M}}
\newcommand{\x}{\mathbf{x}}
\newcommand{\y}{\mathbf{y}}
\newcommand{\z}{\mathbf{z}}
\newcommand{\tr}{\mathrm{tr}}
\newcommand{\I}{\mathbf{I}}
\newcommand{\R}{\mathbb{R}}

\journal{Science of Remote Sensing}

\begin{document}

\begin{frontmatter}



\title{Feature Extraction in the Remote Sensing Data Value Chain: A Systematic Review of Methods and Applications} 



\author[uv]{Nathan~Mankovich}
\author[uv]{Kai-Hendrik~Cohrs}
\author[uv]{Homer~Durand}
\author[wag]{Vasileios~Sitokonstantinou}
\author[uv]{Tristan~Williams}
\author[uv]{Gustau~Camps-Valls}

\affiliation[uv]{organization={Image Processing Lab, Universitat de Val\`encia},
            addressline={C/ Cat. Agustín Escardino Benlloch, 9}, 
            city={Paterna},
            postcode={46980}, 
            state={Val\`encia},
            country={Spain}}

\affiliation[wag]{organization={Artificial Intelligence Group, Wageningen University},
            addressline={P.O. Box 16},
            postcode={6700 AA},
            city={Wageningen},
            country={The Netherlands}}

\begin{abstract}
Earth observation involves collecting, analyzing, and processing an ever-growing mass of data. This planetary data is crucial for addressing relevant societal, economic, and environmental challenges, ranging from environmental monitoring to urban planning and disaster management. However, its high dimensionality entails significant feature redundancy and computational overhead, limiting the effectiveness of machine learning models. Feature extraction (FE) techniques address these challenges by preserving essential data properties while reducing redundancy and enhancing tasks in Remote Sensing (RS). The landscape of FE for RS is diverse, disorganized, and rapidly evolving. We offer a practical guide for this landscape by introducing a framework of FE. Using this framework, we trace the evolution of FE across the data value chain in RS. Finally, we synthesize these trends and offer perspectives for the future of FE in RS by first characterizing this shift from single-task models to unified representations, then identifying two perspectives in the foundation model era: the need for robust and interpretable FE and the potential of bridging classical FE with modern representation learning.
\end{abstract}




\end{frontmatter}


\section{Introduction}

Advancements in Remote Sensing (RS) technologies have ushered in an era of unprecedented data availability, with modern RS platforms continuously generating high-resolution spatial, spectral, and temporal Earth observation data from local to global scales. Data volume, velocity, variety, and dimensionality are expected to grow faster as imaging systems improve~\citep{reichstein2019deep}. These datasets have revolutionized domains such as environmental monitoring~\citep{li2020review}, natural resource management~\citep{kingra2016application}, urban planning~\citep{wellmann2020remote}, agricultural activity monitoring~\citep{weiss2020remote}, and disaster management~\citep{van2000remote}, offering essential information to support timely and informed decision-making.

\begin{figure}[!ht]
    \centering
    \includegraphics[width=.9\linewidth]{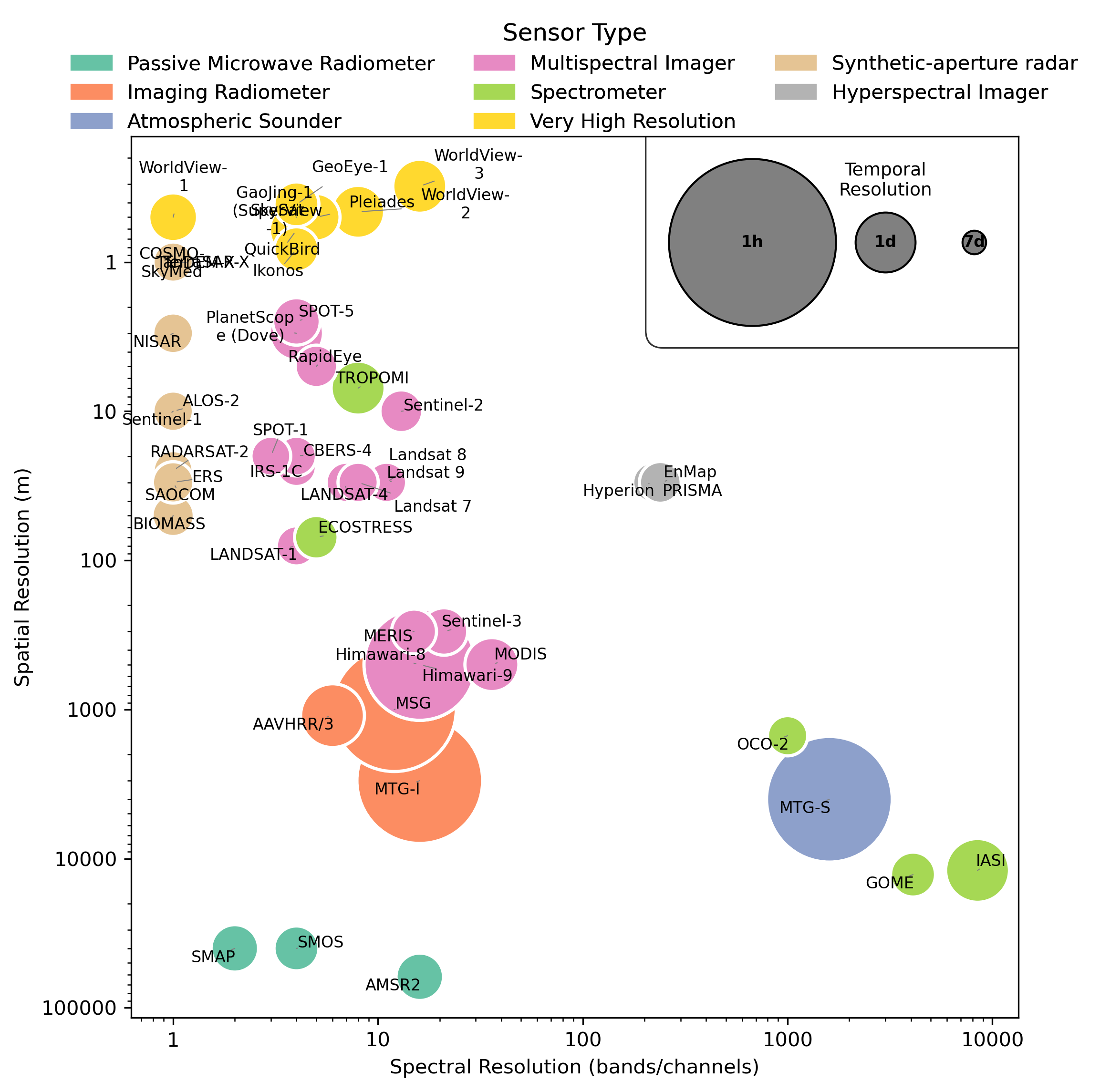}
    \caption{
    \textbf{The utility of FE for RS data from various widely used Earth observation sensors.} We show a simplified representation of the dimensions of remote sensing data, with spectral on the x-axis, spatial on the y-axis, and temporal indicated by the circle size. In the legend, any circle larger than the left circle has a revisit time exceeding an hour, and any circle smaller than the right circle has a revisit time less than weekly. Of course, many other dimensions in these data introduce redundancy (see Tab.~\ref{tab:rs_mod}). Airborne instruments (e.g., AVIRIS, HyMap, UAVSAR, and airborne LiDAR) share the same sensing modalities as their spaceborne counterparts but differ in spatial scale, revisit pattern, and acquisition flexibility.} 
    \label{fig:RS_FE_3D}
\end{figure}

Sophisticated data processing techniques are crucial for extracting actionable insights from complex datasets. These methods, including data mining and machine learning, help identify patterns, make predictions, and derive meaningful interpretations. However, high data volume and dimensionality—across spectrum, space, and time—present significant challenges. For example, satellites like the Sentinel missions produce $8$ to $12$ terabytes of synthetic aperture radar (SAR) and optical imagery daily~\citep{shurmer2018sentinels}, and the near-real-time data stream from weather satellites such as the geostationary operational environmental satellite (GOES-R) series provides continuous monitoring of atmospheric conditions~\citep{goodman2020goes}. Furthermore, the volume, variety, and complexity of RS data dimensionality are illustrated in Fig.~\ref{fig:RS_FE_3D}. As dimensions increase, many techniques become computationally impractical. Even with large datasets, high dimensionality can lead to unreliable distance metrics and an increased risk of overfitting in machine learning models. These problems are commonly summarized as \textit{the curse of dimensionality}~\citep{altman2018curse}.

\begin{figure*}[t!]
    \centering
    \includegraphics[width=\linewidth]{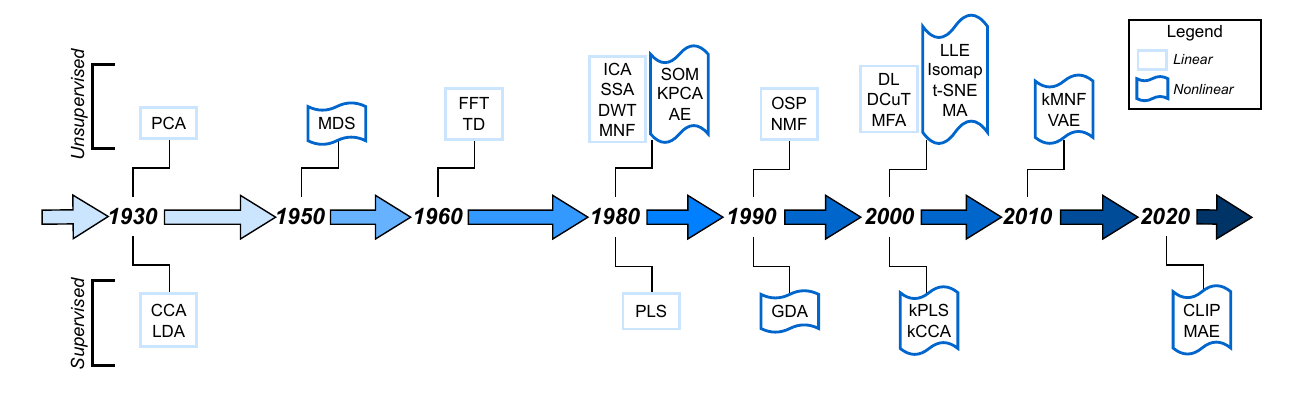}
    \caption{\textbf{A timeline of common FE methods.} FE for feature extraction began with linear multivariate analysis methods such as Principal Component Analysis (PCA)~\citep{hotelling1933analysis} in the $1930$s and became popular in the RS community in the $1970$s~\citep{lyzenga1978passive}. The nonlinear FE boom began in the late $1900$s, including manifold learning and methods such as kernel PCA (kPCA)~\citep{scholkopf1997kernel}, and was quickly adopted by the remote sensing community within $10$ years~\citep{camps2009kernel}. Increasing computing power has enabled the popularity of deep learning-based FE methods, such as the Variational Autoencoder (VAE)~\citep{kingma2013auto}. Deep learning has been quickly adopted by the RS community~\citep{zhu2017deep}. This paved the way for modern deep learning methods such as Contrastive Language-Image Pretraining (CLIP) and Masked Autoencoders (MAE). These methods are already being applied in RS (e.g., Sat-CLIP~\citep{klemmer2023satclip}). See Tab~\ref{tab:dr_glossary_standard} for a glossary of FE abbreviations.}
    \label{fig:history}
\end{figure*}

To address these issues and fully extract value from the data, feature extraction (FE) methods are crucial. Unlike feature selection techniques, FE for feature extraction analyzes the original data and \emph{extracts low-dimensional features} from high-dimensional data while preserving the essential properties needed for downstream analysis. Here, ``dimension'' denotes the number of features used by the representation at a given stage (e.g., spectral bands, spatial-temporal descriptors, or learned reduced features), rather than the width of intermediate hidden layers inside deep networks.
FE methods to extract low-dimensional spectral, spatial, and/or temporal features can enhance the value of RS data from preprocessing and analysis to the improvement of RS products. Over the last century, the field of FE has grown in popularity 
and developed a dense, fragmented landscape of FE methods, ranging from linear multivariate analysis to deep learning (see Fig.~\ref{fig:history}). Thus, the problem of high-dimensional data can be addressed by a FE method, but it is replaced by a secondary problem of selecting the optimal method from the vast FE landscape.

With the proper navigation tools for the FE landscape, families of methods can be identified for specific RS tasks at each level of the RS data value chain. Previous works navigate the landscape of FE with an eye for RS applications by either focusing on hyperspectral data~\citep{9082155, 9451654}, specific RS data tasks~\citep{dua2020comprehensive,rasti2021image,li2022deep,dey2018big,hu2022hyperspectral,maxwell2018implementation}, or by restricting to small pieces of the FE landscape~\citep{9451654,wang2023tensor,izquierdo2017advanced}. For example, some taxonomies only address linear methods~\citep{van2009dimensionality} or overlook them entirely~\citep{lee2007nonlinear}. Meanwhile, others miss a perspective on deep learning~\citep{nanga2021review}.

\clearpage
\newpage
\begin{figure}[t!]
    \centering
    \includegraphics[width=0.85\linewidth]{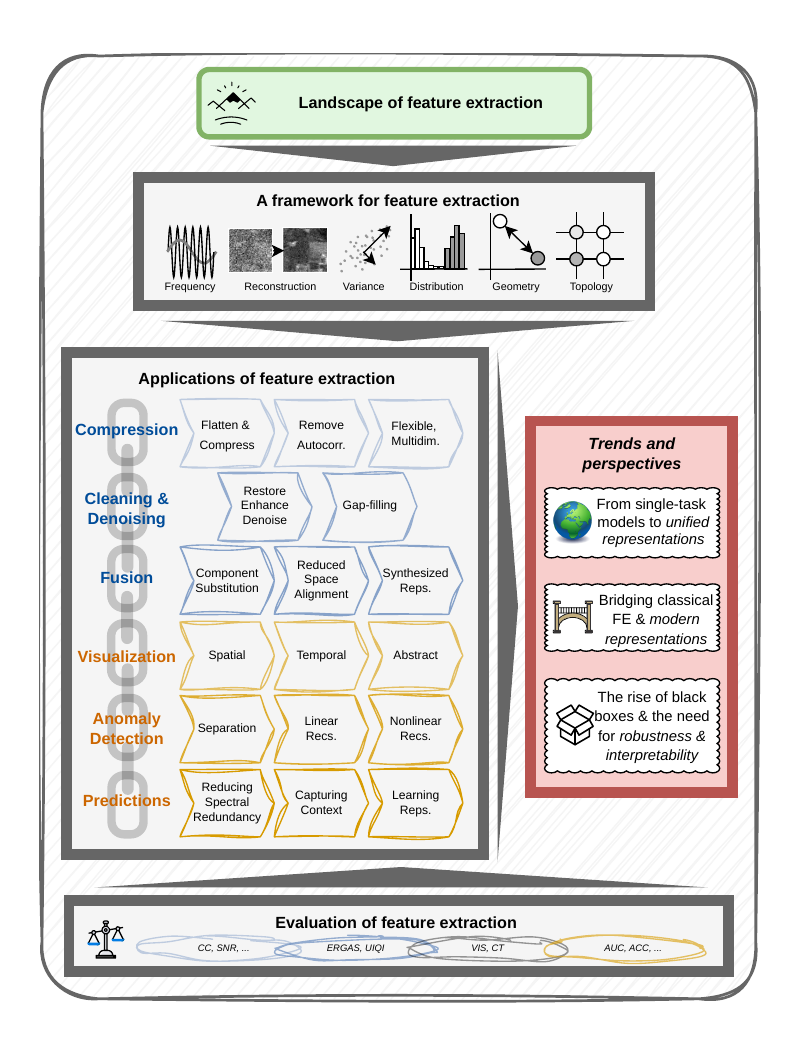}
    \caption{\textbf{A graphical abstract for FE in RS.} We provide \textit{a framework for FE} and use it to characterize standard FE methods in RS within the complex landscape of FE (see Sec.~\ref{sec:framework}). Once organized, these methods are tracked across the RS data value chain, as we traverse the \textit{applications of FE} in RS (see Sec.~\ref{sec:dr_rs}). Then, we present application-specific \textit{evaluation of FE} (see Sec.~\ref{sec:eval}). Finally, the trends of FE in RS are synthesized, yielding three key perspectives for the future of FE in the foundation model era (see \textit{trends and perspectives}, Sec.~\ref{sec:perspective}).}
    \label{fig:concept}
\end{figure}
\clearpage
\newpage

We address all three limitations through a modern, comprehensive review of FE methods applied across the entire RS data value chain, one that moves beyond hyperspectral data analysis and incorporates the transition in modern FE, moving towards deep learning and foundation models. What follows is a guide for using FE in RS, outlined in Fig.~\ref{fig:concept}. The logical flow in Fig.~\ref{fig:concept} is sequential: the framework organizes method families, which organization structures the task-wise survey, and both determine which evaluation criteria are informative for each task. Specifically, we begin with a framework for standard FE methods in RS. Then, using this framework as a map, we provide a systematic survey of FE applications for improving each task in the RS data value chain and outline standard metrics for FE evaluation in RS. Finally, we summarize the trends and perspectives and outline the way forward for FE in RS applications.

\section{A framework for feature extraction}\label{sec:framework}

\begin{figure}[H]
    \centering
    \includegraphics[trim={1.5cm 0cm .5cm 0cm
}, clip, width = .77\linewidth]{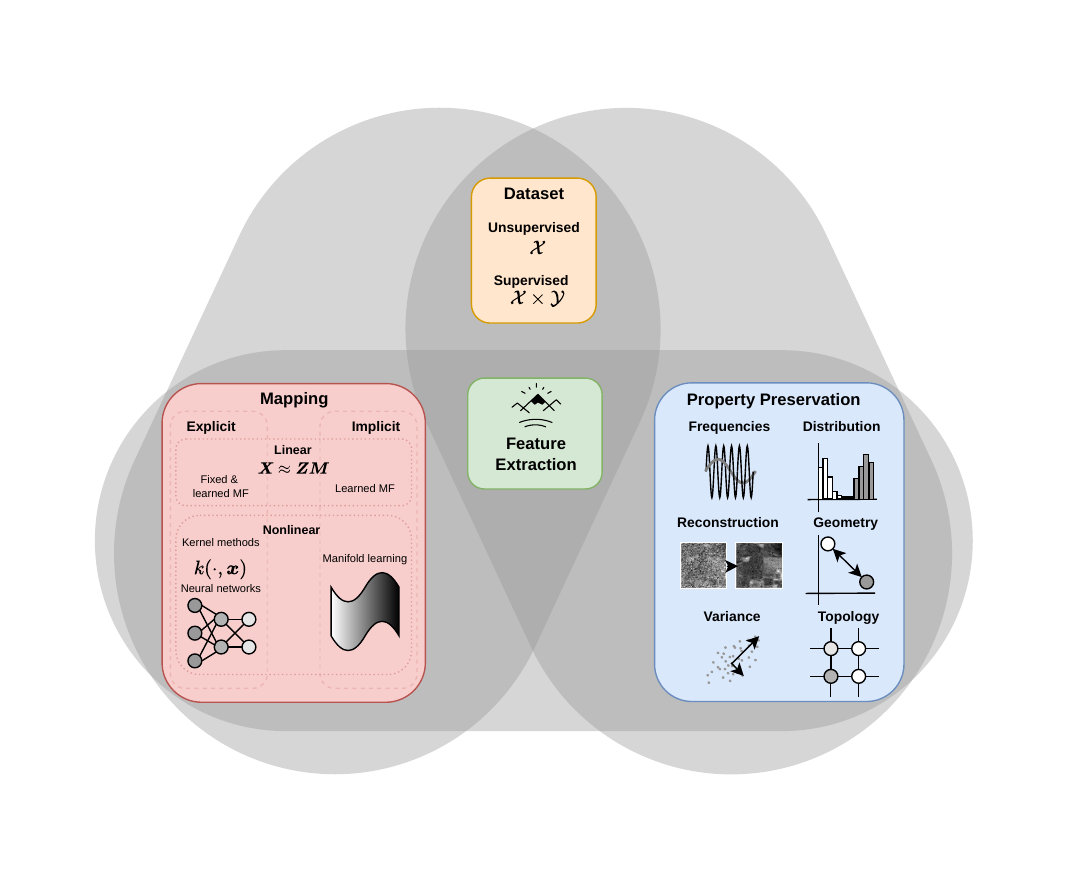}
    \caption{\textbf{A framework of FE.} FE characteristics are separated into the three pillars: mapping, dataset, and property preservation. The FE mapping can be either explicit or implicit and either linear or nonlinear. We separate classes of FE mappings by their mapping mechanisms: fixed or learned matrix factorization (MF), kernel methods, neural networks, and manifold learning. Different FE methods are used for different tasks based on the input dataset. Unsupervised FE methods just input a dataset $\mathcal{D} \subset \mathcal{X}$, whereas supervised methods take a dataset of pairs $\mathcal{D} \subset \mathcal{X} \times \mathcal{Y}$ as inputs. The property preservation for FE algorithms includes data frequencies, reconstructions, variance, distributions, geometry, and topology.}
    \label{fig:dr_properties}
\end{figure}

The field of Feature Extraction (FE) is expansive and populated with a zoo of algorithms ranging from linear multivariate analysis and manifold learning to deep learning. Selecting an appropriate FE technique for a given RS task can be challenging. Our novel framework, illustrated in Fig.~\ref{fig:dr_properties}, structures the field of FE, thus providing a practical guide for researchers. We characterize FE methods into families based on three axes: the input dataset (Sec.~\ref{sec:dataset}), the mapping (Sec.~\ref{sec:mapping}), and the properties preserved (Sec.~\ref{sec:prop_pres}). Compared with taxonomies that primarily separate methods by data modality, specific task, or linear/nonlinear type alone, this three-pillar view jointly captures supervision assumptions, mapping mechanism, and preservation objective in one schema.

\begin{figure}[H]
    \centering
    \includegraphics[width=\linewidth]{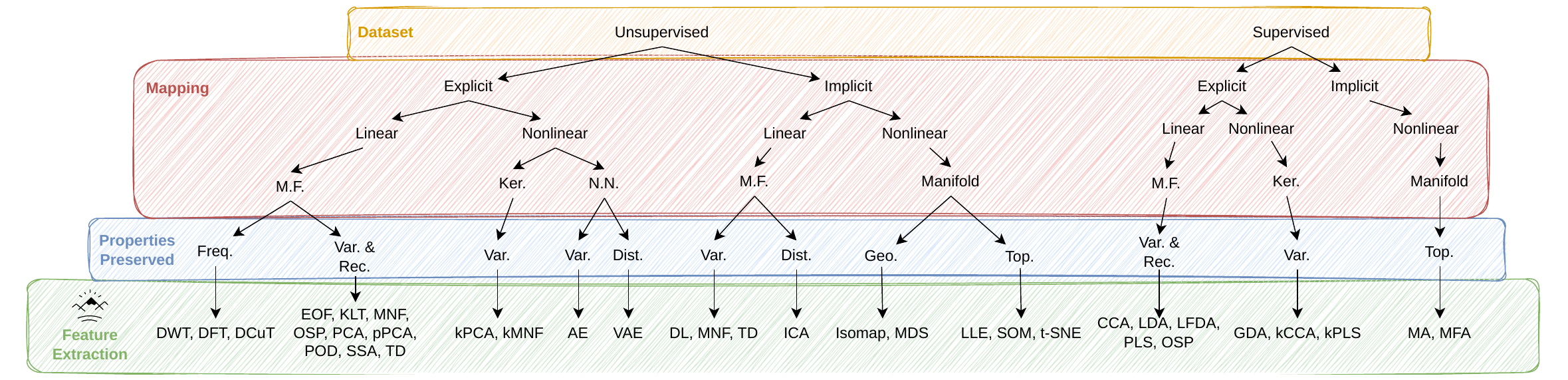}
    \caption{\textbf{Common FE in RS characterized by their dataset, mapping, and properties preserved.} We use the following abbreviations for mapping mechanisms: M.F. for matrix factorizations, Ker. for kernel methods, N.N. for neural networks, and Manifold for manifold learning. For preserved properties, we abbreviate frequencies and signal structure as Freq., and variance and reconstruction as Var. \& Rec., distribution as Dist., geometry as Geom. and topology as Top.}
    \label{fig:dr_taxonomy}
\end{figure}

Historically popular FE methods in RS are placed in our framework in Fig.~\ref{fig:dr_taxonomy}. Using this framework, a practitioner can navigate the FE landscape to identify a family of algorithms best suited for their goal.

\subsection{Dataset}\label{sec:dataset}
The \emph{dataset} $\mathcal{D}$ is the input data to FE and contains $N$ samples in the high, $P$-dimensional, ambient space. Depending on the application, samples can be anything from pixels, entire images, spectral bands, or even a collection of images. After samples are chosen, the feature dimensions are the attributes associated with each sample. For example, in pixel classification of hyperspectral imagery, samples are pixels, and features are spectra. We characterize FE methods as supervised or unsupervised based on the structure of the input dataset. For supervised FE methods, our input dataset consists of a set of $N$ paired samples $\mathcal{D} =\{(\x_1,\y_1),\ldots,(\x_N,\y_N)\} \subset \mathcal{X}\times\mathcal{Y}$, where $\mathcal{X}$ is the ambient space and $\mathcal{Y}$ contains auxiliary information (e.g., class labels) that guide the feature extraction process. In contrast, unsupervised methods do not rely on such labels, and the input dataset is only $\mathcal{D} = \{\x_1,\ldots, \x_N\}\subset \mathcal{X}$. Notably, self-supervised FE techniques do not rely on external labels; instead, they generate pseudo-labels from the data's inherent structure of the data and are therefore unsupervised.

\subsection{Mapping}\label{sec:mapping}
Regardless of whether or not a FE method is supervised, the input RS data is high-dimensional, redundant, corrupted by noise, and challenging to interpret. FE reduces the feature dimension of an input dataset from a high, $P$-dimensional ambient space to a low, $K$-dimensional reduced space. In the reduced space, the extracted features capture the essential information of the data in fewer dimensions, often eliminating redundancy and noise while enhancing interpretability. This transformation is done by the FE mapping. 

FE mappings can be either explicit or implicit. \emph{Explicit FE} mappings, denoted $\boldsymbol{\phi}$, transform the ambient features into reduced features, whereas \emph{implicit FE} outputs the reduced data without defining the mapping $\boldsymbol{\phi}$. Thus, explicit FE can be applied to new data using $\boldsymbol{\phi}$, whereas implicit FE cannot. In explicit FE, an (often approximate) inverse FE mapping (denoted $\boldsymbol{\psi}(\cdot)\approx\boldsymbol{\phi}^{-1}(\cdot)$) can be either learned or directly computed from $\boldsymbol{\phi}$. The inverse FE mapping allows us to reconstruct the data from its reduced representation, enabling tasks in RS, like denoising. 

Beyond explicit and implicit mappings, FE mappings are separated into linear and nonlinear. \emph{Linear} FE has lower computational complexity, higher interpretability, and often a closed-form solution. In contrast, \emph{nonlinear} FE captures more complex nonlinear relationships between data. This definition is compatible with deep networks that may temporarily increase hidden representation size: FE is assessed at the task-level embedding used downstream, where the final representation remains reduced relative to the original descriptor. For explicit FE, the linear vs nonlinear dichotomy refers to the structure of $\boldsymbol{\phi}$. For implicit FE, we distinguish between linear and nonlinear methods based on the technique's ability to preserve nonlinear structures in the data. Overall, while nonlinear methods showcase impressive advantages~\citep{lee2007nonlinear}, linear techniques remain valuable in various practical scenarios. 

We now characterize different linear and nonlinear FE methods by their mapping mechanism. The primary mapping mechanism for linear FE methods is \emph{matrix factorization}, meaning that they factor the high-dimensional data matrix $\X \in \R^{N \times P}$ (samples $\times$ ambient features) into a mixing matrix (sometimes called dictionary) $\M\in \R^{K \times P}$ and the matrix storing the reduced representations $\Z\in \R^{N \times K}$ as $\X \approx \Z \M$. Matrix factorizations are divided into fixed factorizations (e.g., predefined $\M$) and those with learned factorizations (e.g., learned $\M$). Respective examples of fixed factorizations and learned factorizations are the Discrete Wavelet Transform (DWT)~\citep{broughton2018discrete} and Principal Component Analysis (PCA)~\citep{hotelling1933analysis}. Fixed matrix factorization methods are often explicit, whereas learned matrix factorizations are divided into implicit methods, such as Dictionary Learning (DL)~\citep{kreutz2003dictionary}, and explicit methods, such as DWT and PCA.

The other family of mapping mechanisms, nonlinear FE, comprises three mechanisms: kernels, neural networks, and manifold learning. \emph{Kernel methods} are a large family of non-linear explicit FE methods that use a kernel mapping mechanism~\citep{CampsValls09}. These methods use the kernel trick to generalize classical linear FE methods to nonlinear methods, like kernel PCA (kPCA)~\citep{scholkopf1997kernel}, by applying them in a high-dimensional (potentially infinite) feature space $\mathcal{H}$. The foundation of kernel methods is the kernel trick, which circumvents the need to specifically compute a mapping $\boldsymbol{\varphi}$ into $\mathcal{H}$ while allowing us to compute properties in that space directly. It does this through the kernel function $k$, enabling the computation of similarities in the feature space as
\begin{equation}
    k(\x_n,\x_m) = \langle \boldsymbol{\varphi}(\x_n), \boldsymbol{\varphi}(\x_m)\rangle_{\mathcal{H}}.
\end{equation}
\noindent The most popular kernel function, the RBF Kernel, is ``universal,'' i.e., it can uniformly approximate any function~\citep {micchelli2006universal}.

A second family of explicit nonlinear FE methods uses a \emph{neural network} mapping mechanism to parameterize FE mappings. For example, Autoencoders (AEs)~\citep{bank2023autoencoders} are a flexible FE method that learns a FE mapping (encoder) and its approximate inverse (decoder) by minimizing a loss function. Design choices for the neural network methods include the number of layers and hidden units, the type of nonlinearity, and the properties preserved in the loss function. Recently, deep neural network methods have gained popularity as they enable training models with millions of parameters and the assimilation of vast amounts of data. Neural network methods offer high flexibility, albeit at the cost of reduced interpretability. 

Both kernel methods and neural networks are explicit FE mappings; most implicit nonlinear FE uses a \emph{manifold learning} mapping mechanism. These methods optimize directly for the reduced data representation, often aiming to mirror the ambient data manifold's structure in the reduced space by preserving geometric or topological properties. One of the most popular manifold learning FE methods for data visualization is t-distributed Stochastic Neighborhood Embedding (t-SNE)~\citep{van2008visualizing}.

\subsection{Property preservation}\label{sec:prop_pres}
The distinction between different property preservation goals is the core utility of our FE framework. Most families, other than frequency- and signal-structure-preserving FE, preserve properties from the ambient space in the reduced space by defining FE methods as solutions to optimization problems. The inputs to the objective function can include ambient data, reduced data, FE mapping, and/or its approximate inverse. Then the objective function returns a value that represents how well the property of interest is preserved. 

\paragraph{Frequency and signal-structure preserving}
A small family of FE methods designs a pre-defined FE mapping to \textit{preserve frequency and other signal structures} in the reduced space. These transforms are a simple and computationally efficient, task-agnostic family of FE methods. They prescribe a linear FE mapping $\boldsymbol{\Phi} \in \mathbb{R}^{K \times N}$ and approximate inverse $\boldsymbol{\Psi} \in \mathbb{R}^{N \times K}$ in
\begin{equation}\label{eq: freq signal preserving}
    \z = \boldsymbol{\Phi} \x , \quad \x \approx \boldsymbol{\Psi} \z, \quad \boldsymbol{\Phi} \text{ fixed.}
\end{equation}
These mappings are often aligned to the desired signal structure through a basis of functions. The DWT is a classic example of these methods, as it uses a mother wavelet to construct $\boldsymbol{\Phi}$ and preserves both signal location and scale in the reduced space.

\paragraph{Variance preserving}
Variance-preserving methods, like the unsupervised PCA and the supervised Partial Least Squares (PLS)~\citep{Elsevier-1986GeladiK}, use a trace-ratio objective in Eq.~\ref{eq: leared var mf} where $\mathbf{A}$ encodes the signal we want to preserve and $\mathbf{B}$ encodes the noise or irrelevant directions in the data.
\begin{equation}\label{eq: leared var mf}
    \max_{\M} \frac{\tr(\mathbf{M} \mathbf{A} \mathbf{M}^\top)}{\tr(\mathbf{M} \mathbf{B} \mathbf{M}^\top)}
\end{equation}

Classical variance-preserving methods are explicit and linear and thus incapable of capturing nonlinear structures in the reduced space. Kernel methods generalize most of these classical FE methods from trace-ratio problems to nonlinear, explicit, variance-preserving methods like kPCA~\citep{scholkopf1997kernel} and kernel PLS (kPLS)~\citep{Elsevier-1986GeladiK}.

\paragraph{Reconstruction preserving}
Some variance-preserving methods, like PCA, can also be viewed as reconstruction-preserving methods because, under certain conditions, reconstruction minimization is equivalent to trace maximization. Specifically,
\begin{equation*}
    \underset{\M \M^\top = \I}{\mathrm{argmin}} \: \: \| \X - \X \M^\top \M \|_F = \underset{\M \M^\top = \I}{\mathrm{argmax}}\: \: \tr(\M \X^\top \X \M^\top).
\end{equation*}
However, not all reconstruction preserving methods are variance-preserving. These methods include dictionary learning, tensor factorization, and autoencoders.

Thus far, we have touched on FE methods with fixed constraints (e.g., orthogonality) that lack the flexibility necessary for some RS tasks. \textit{Dictionary Learning} (DL)~\citep{kreutz2003dictionary} offers an alternative to these variants through learned matrix factorization (e.g., implicit, linear FE) that aims to preserve data reconstructions while satisfying constraints $\mathcal{C}$ on $\Z$, and/or the dictionary, $\M$ (see Eq.~\ref{eq: dl}). 
\begin{equation}\label{eq: dl}
    \min_{\M, \Z} \| \X - \M \Z \| \quad \mathrm{s.t.} \: \: \M, \Z \in \mathcal{C}.
\end{equation}
\noindent For instance, when modeling a known physical process, one may want to enforce this constraint because negative entries would contradict the physical understanding of the process (e.g., temperatures in Kelvin). Non-negative Matrix Factorization (NMF)~\citep{wang2012nonnegative} adds these hard constraints to the optimization problem by enforcing positive entries in $\M$ and $\Z$.

Although diverse and useful, these matrix-based approaches, like PCA, kPCA, and DL, are misaligned with the structure of most RS data. Most RS data arrives as $3$-dimensional (a.k.a. $3$rd-order) tensors with spatial (latitude and longitude) and spectral dimensions. \textit{Tensor Decomposition} (TD) methods~\citep{wang2023tensor} generalize matrix-based FE methods to tensors by jointly reducing multiple dimensions of RS data within a single decomposition framework. TD methods are often both variance- and reconstruction-preserving. The most common TD method is the Tucker decomposition, which generalizes the singular value decomposition (SVD) used in matrix FE methods such as PCA and DL to tensor inputs. This method factors a tensor into a core tensor and matrix factors and is not scalable to high-order tensors. The canonical polyadic decomposition factors a tensor into an outer product of vectors, removing the need for a core tensor, but it suffers from unstable optimization. The tensor train decomposition is more stable, scalable, and memory-efficient than previous TD methods, but it depends on the ordering of the tensor's dimensions. For example, a (bands $\times$ time $\times$ space) tensor and a (time $\times$ space $\times$ bands) tensor would have different decompositions. The tensor ring decomposition is a modern TD method that solves the problem of order dependence by being invariant under circular permutations of dimensions while remaining memory-efficient and scalable to high-order tensors, yet it incurs higher computational complexity.

More modern, flexible, and nonlinear FE actually learns FE mappings (encoder $\boldsymbol{\phi}$ and decoder $\boldsymbol{\psi}$) parameterized by neural networks by minimizing a loss function $\mathcal{L}$ 
\begin{equation}\label{eq: ae loss}
    \min_{\boldsymbol{\phi}, \boldsymbol{\psi}} \mathcal{L}(\X, \boldsymbol{\phi}, \boldsymbol{\psi}).
\end{equation}
\noindent using gradient descent variants~\citep{rumelhart1986learning}. A typical example of these methods is AEs. Although AEs were initially designed to minimize only reconstruction error, they can optimize any sufficiently smooth objective function $\mathcal{L}$ and thus incorporate various regularizers enforcing properties of interest (e.g., physical, causal, probabilistic, geometric, and topological)~\citep{bank2023autoencoders}. However, this flexibility of AEs comes at a cost. AEs often lack interpretability and theoretical guarantees, require many samples to fit the data properly, and are computationally expensive to train.

\paragraph{Distribution preserving}
Distribution-preserving methods partially remedy the lack of theoretical guarantees for AEs by imposing a statistical model on the reduced data distribution, yielding robust models that are statistically consistent across observations. These methods are exemplified by the Variational AE (VAE)~\citep{kingma2013auto}, which learns a generative model of the data. Specifically, the VAE models a reduced (a.k.a. latent) space prior $p(\z)$ with an encoded probability $q_\phi (\z|\x)$ and produces reconstruction probabilities $p_\psi(\x|\z)$. Then, VAEs optimize the data representations by maximizing the Evidence Lower Bound (ELBO)
\begin{equation*}
    \mathrm{ELBO}(\boldsymbol{\phi}, \boldsymbol{\psi}, \x) = \mathbb{E}[\log p_\psi(\x|\z)] - \mathrm{D}_{\mathrm{KL}}(q_\phi (\z|\x) || p (\z)).
\end{equation*}
The first term ensures reconstruction accuracy, while the Kullback–Leibler (KL) term enforces regularity of the reduced space distribution.

Related generative approaches include Generative Adversarial Networks (GANs)~\citep{rezende2015variational}, which learn a generator that maps latent variables to data through adversarial training without explicitly modeling the data likelihood, and normalizing flows~\citep{goodfellow2020generative}, which learn invertible transformations between data and latent variables that permit exact likelihood evaluation. While GANs emphasize high-fidelity sample generation, flows provide tractable density estimation and bijective latent representations.

\paragraph{Geometry preserving}
Remote sensing data is often governed by a small number of continuous parameters. Under smooth forward models, moderate noise, sufficient sampling density, and few regime changes, the data concentrate near low-dimensional manifolds embedded in the ambient space. These manifolds are locally Euclidean and thus can be partially captured locally by linear FE methods. However, linear methods miss the global structure of the nonlinear data manifold. Although some methods discussed so far are nonlinear, they neither directly preserve the manifold geometry nor its topology. Implicit, nonlinear FE, known as manifold learning, assumes that high-dimensional data lie on a low-dimensional manifold and aims to preserve either the global geometry or the local topology in the reduced space.

Early attempts, like Multidimensional Scaling (MDS)~\citep{saeed2018survey}, \textit{preserve global geometry} through matching distances between ambient and reduced spaces, e.g.,
\begin{equation}
    \min_{\{\z_n\}_{n=1}^N}\sum_{n>m}^{N} (d_{n,m} - \|\z_n - \z_m\|_2)^2.
\end{equation} 
Standard MDS fails to capture local structures because the chosen ambient space distance between points $n$ and $m$ (denoted $d_{n,m}$) is often not the true distance measure on the data manifold. Isomap~\citep{balasubramanian2002isomap} improves estimation of $d_{n,m}$ by using geodesic distances, thus better capturing the true data manifold structure.

\paragraph{Topology preserving}
The community quickly moved from global geometry-preserving FE to local topology-preserving FE to preserve neighborhood structures in the reduced space. One of the first methods in this paradigm, Locally Linear Embedding (LLE)~\citep{saul2000introduction}, uses nearest-neighbor graphs to preserve local structures. Although LLE is a theoretically sound FE method, it requires a smooth data manifold, well-sampled data, and consistency among locally linear patches.

Trading the geometric faithfulness of LLE for visual separability, t-SNE is a widely used local topology-preserving FE method. t-SNE models both the ambient and reduced data using similarity graphs based on the RBF kernel. Then it uses a probabilistic approach to align the edge distribution of the reduced graph $Q$ with that of the ambient graph $P$. Using $p_{n,m}$ as the edge weight between node $n$ and $m$ in $P$ and $q_{n,m}$ as the modeled edge weight in $Q$, t-SNE minimizes the KL divergence between $P$ and $Q$:
\begin{equation*}
\mathrm{D}_{\mathrm{KL}}(P||Q) = \sum_{n > m} p_{n,m} \log \bigg(\frac{p_{n,m}}{q_{n,m}}\bigg)
\end{equation*}
via gradient descent. Since t-SNE is an iterative method, careful treatment of the initial conditions leads to better-reduced spaces~\citep{kobak2021initialization}. Although t-SNE preserves local neighborhoods, resulting in clustered low-dimensional embeddings, it suffers from high computational cost, sensitivity to perplexity, and a non-convex objective.

\subsection{Synopsis}
Our characterization of FE methods provides a clear framework for navigating the landscape of FE techniques. It is built on three pillars: the input dataset, the mapping, and the properties preserved. While the dataset and mapping criteria create a mutually exclusive and collectively exhaustive classification, property preservation does not, as a single method can serve multiple objectives. 

This framework for FE is not merely a catalog; it is a tool designed to guide practitioners in selecting appropriate algorithms for their applications. The complete taxonomy of common FE methods in RS, as visualized in Fig.~\ref{fig:dr_taxonomy}, serves as a practical decision-making guide for any RS practitioner. For a quick reference, all common FE methods in RS are listed in the glossary; see Tab.~\ref{tab:dr_glossary_standard}. Now, armed with a structured understanding of common FE in RS, we are ready to analyze how these methods improve tasks in the RS data value chain. By applying this framework to the literature, we can extract underlying patterns that reveal why certain families of FE methods are consistently used for specific tasks in the RS data value chain.

\section{Applications of feature extraction}\label{sec:dr_rs}

Having established our FE framework, we use it now as a guide to follow an RS dataset through the data value chain. In doing so, we will move \emph{beyond HS data}, given the varied needs for FE across data types (see Fig.~\ref{fig:RS_FE_3D}). This journey begins with preprocessing tasks: compression, cleaning, and fusion of raw remote sensing data. Then we proceed to the essential analysis stage, where FE is used for visualization, anomaly detection, and ultimately for empowering predictive models to generate improved scientific insight. The content of this section is summarized in Tab.~\ref{tab:master_summary}.

\input{tables/master_summary}

\subsection{Preprocessing}
The goal of preprocessing is to prepare raw RS data by isolating the underlying signal from noise and atmospheric effects. Raw RS data is often high-volume and high-complexity, thus necessitating compression for easy transport and storage. These data also require additional cleaning and denoising steps to ensure that sensor artifacts, noise, and misregistration do not dominate the signal. Once the signal is isolated, a wide variety of RS data modalities and resolutions from different sensors (see Fig.~\ref{fig:RS_FE_3D}) are often fused to accomplish a specific task. FE techniques themselves are the primary engines of these preprocessing tasks, empowering the core tasks of compression (\ref{sec:compression}), data cleaning (\ref{sec:cleaning}), and fusion (\ref{sec:fusion}).

\subsubsection{Compression}\label{sec:compression}
Data compression reduces the dimensionality of remote sensing data while preserving important information, thereby improving downstream tasks such as parameter retrieval, unmixing, and classification~\citep{Garcia-Vilchez2011253, garcia2019improved}. Specifically, it addresses challenges involving limited transmission channel bandwidth, transmission time, and storage space by removing redundant information from data onboard platforms (e.g., satellites, drones) and on the ground. 

Data compression is divided into two tasks: lossless and lossy. Lossless compression reduces data volume while preserving perfect reconstruction, whereas lossy compression allows some information loss. \emph{We only consider FE for lossy compression} because FE reconstructions are generally imperfect. Algorithms for lossy compression consist of an encoder, a bitstream translation, and a decoder. A FE method for RS compression is almost always unsupervised using an explicit mapping to provide an encoder and decoder via $\boldsymbol{\phi}$ and $\boldsymbol{\psi}$.

\paragraph{Flatten \& compress}
RS data naturally contains spatial, spectral, and temporal dimensions. Initially, dimensions were compressed individually, or combinations of dimensions were flattened, then compressed. Originally, linear matrix factorization FE methods that preserve frequency, variance, reconstruction, and/or distributions (e.g., DFT, PCA, and ICA) were tested for the compression of RS data~\citep{benz1995comparison, kaarna2000compression}. These methods were surpassed by the foundation for RS data compression, JPEG$2000$~\citep{skodras2001jpeg}. This method breaks the image into tiles and utilizes the DWT to compress the spectral dimension. Although other, more flexible implicit FE methods, like NMF, have been tested for compression~\citep{wang2006independent}, DWT-based compression remained the baseline FE method, proving effective for compressing even ultraspectral sounder data~\citep{serra2008remote}. Overall, linear FE methods, with predefined mappings that preserve frequency and signal structure, have proven to be solid baselines for compression.

\paragraph{Removing autocorrelation}
Autocorrelation in the spectral bands of Multispectral (MS) and Hyperspectral (HS) data results in spectral redundancy that is poorly compressed by standard JPEG. Pre-processing with the linear, explicit, frequency, and reconstruction-preserving FE like $3$D DWT~\citep{penna2006progressive} and PCA variants~\citep{Du2007,penna2007transform} decorrelates these bands, thus reducing spectral redundancy and improving the effectiveness of subsequent DWT compression.

\paragraph{Flexible, multidimensional compression}
Although promising advances have been made, these methods either compress each dimension individually or flatten dimensions of spatio-temporal data cubes before applying compression, thereby missing the structure of RS datasets. An initial remedy for this issue is video compression as it is a promising method for faithful compression of both spatial and temporal dimensions~\citep{pellicer2025video}. A more general solution is TD methods, which extend traditional linear matrix factorization to handle the data cube in its natural form, thereby preserving its inherent structure. For example, the Tucker decomposition improves upon DWT-based compression~\citep{Karami2012}. In general, TD methods are highlighted as a promising research direction for HS compression FE~\citep{garcia2017statistical}.

Due to increases in computational power and, consequently, the rising popularity of deep learning, FE methods for compression no longer need to rely on rigid matrix factorizations. Consequently, learnable neural network FE shows promise for modern RS data compression. For example, AEs improve data-compression flexibility by learning optimal nonlinear transforms directly from data rather than using a fixed DWT or TD basis~\citep{xiang2024remote}.

\subsubsection{Data cleaning}\label{sec:cleaning}
Data cleaning tackles issues involving data quality during the preprocessing phase. For example, cloud cover affects the usability of optical satellite imagery~\citep{prudente2020limitations} and atmospheric interference introduces noise and reduces land cover classification rates~\citep{vanonckelen2013effect}. Data cleaning addresses these problems by either separating signal from noise via image restoration, enhancement, and denoising, or by generating new information to fill in missing data. In hyperspectral settings, this also includes physically motivated decompositions such as spectral unmixing, which separate mixed pixel observations into constituent materials and their abundances~\citep{keshava2002spectral}. In general, these tasks are performed in spatial, temporal, and/or spectral dimensions and include cloud, shadow, and haze removal, as well as sensor error correction, such as image de-striping~\citep{Review2015, CampsValls21wiley}. FE for data cleaning leverages reconstructions from the reduced space to generate uncorrupted images.

\paragraph{Image restoration, enhancement, and denoising}
Denoising can be performed individually in each dimension of the HS image or simultaneously across multiple dimensions using unsupervised FE. Linear, frequency, variance, and reconstruction preserving methods (e.g., DWT, PCA, and MNF) work by concentrating the signal structure into a few reduced components and discarding the ``noise'' in the remaining components. In this sense, hyperspectral unmixing plays a similar role by expressing each pixel in a low-dimensional, physically interpretable basis of endmembers and abundances. The DWT and PCA are combined for denoising HS data~\citep{chen2010denoising}. PCA has also been adapted for LiDAR denoising~\citep {duan2021low} and compared to MNF for denoising HS data~\citep{luo2016minimum}. Regularized matrix factorization approaches further link unmixing to FE. For example, graph Laplacian regularization produces fractional abundance maps that more precisely capture material distributions, particularly in noisy conditions~\citep{ince2020superpixel}.

As with compression, denoising FE has moved from these linear baselines to more flexible, explicit, and nonlinear deep learning methods. The untied denoising AE is designed for denoising HS data and outperforms state-of-the-art methods in high-noise regimes for spectral unmixing~\citep{qu2018udas}, where the learned representations improve both denoising and abundance estimation. Finally, a contrastive learning approach that pairs clean, noisy, and denoised images in the representation space has outperformed other deep learning approaches in denoising 3-channel images~\citep{Wang2024}. 

\paragraph{Gap-filling}
In gap-filling, there is no signal to separate. The goal is to generate data by learning from the spatial, spectral, and temporal context. We focus on two case studies for gap-filling, namely cloud/shadow replacement and temporal gap filling.

Cloud/shadow replacement is one of the most common spatial gap-filling tasks. FE methods for cloud replacement are generally supervised because they use cloudless reference images from different spatial locations or the same spatial location at other times, and/or other data modalities~\citep{shen2015missing}. In general, FE methods for this task combine reduced representations or FE mappings of cloudy and reference images to replace missing data. Supervised variants of DL-based methods align dictionaries to replace clouds in HS and MS datasets (e.g., Hyperion, OLI, Landsat, and MODIS)~\citep{li2019cloud, xu2016cloud}. 

The frontier of cloud replacement employs deep learning methods, such as AEs, that excel at learning complex contextual relationships. For example, such methods effectively replace clouds in SST measurements~\citep{dong2018inpainting} and MS data~\citep{ding2024robust}. A review of gap-filling using convolutional neural network architectures highlights the utility of these AEs architectures~\citep{qin2021image}.

In contrast with cloud and shadow replacement, FE methods for time-series gap-filling in RS have yet to adopt deep learning and continue to use linear methods that preserve variance and frequency. For example, PCA is used for reconstructing surface chlorophyll, total suspended matter, and sea surface temperature data~\citep{sirjacobs2011cloud} along with MODIS leaf area index products~\citep{kandasamy2013comparison}. Furthermore, frequency-preserving FE, like the Discrete Cosine Transform (DCT), has been incorporated into a specialized algorithm to replace missing soil moisture data~\citep{wang2012three}.

\subsubsection{Fusion}\label{sec:fusion}
The challenge of harmonizing different RS data modalities and resolutions for joint analysis is called fusion. Algorithms for fusion must be computationally efficient, preserve high resolution, and reduce color distortion. Fusion is carried out at different levels: pixel, feature, or decision level~\citep{Ghamisi2019}. We focus on pixel and feature-level fusion where FE is most beneficial. Fusion is divided into \emph{homogeneous fusion} and \emph{heterogeneous fusion}. The former uses only single-modal data and can be applied to any gridded data by matching image locations and applying pixel-level operations. The latter is more modern and flexible because it integrates a broader range of sources. FE has evolved for data fusion through $3$ eras, generally moving from homogeneous tasks to more general, heterogeneous tasks. These eras are component substitution, alignment of a shared reduced space, and learning synthesized representations with deep learning. Throughout these eras, the best models are often supervised and have moved from explicit, linear FE that preserves variance and frequency, to implicit, topology-preserving FE, and finally to flexible deep learning FE with neural network mapping mechanisms.

\paragraph{Component substitution}
The earliest strategies, such as Component Substitution (CS), are explicit, linear FE and were primarily developed for homogeneous fusion, i.e., pansharpening. CS is simple, it runs PCA on the low-resolution image, then substitutes the leading principal components with the high-resolution pan image, and finally maps the new reduced representation back to the ambient space~\citep{Chavez1991}. In applications, CS offers benefits in low color distortion but suffers from spectral distortion in MS and HS data. Various pre-defined matrix factorization FE, such as wavelet, contourlet, or support value transforms, are run before CS to address the persistent challenge of spectral distortion in the fused images~\citep{Luo2008, Yang2013, 4490068, YANG2012177}

\paragraph{Alignment of a shared reduced space}
Limitations of CS, such as its difficulty in handling spectrally diverse datasets, motivated the next step in FE for fusion, paving the way for tools for both homogeneous and heterogeneous tasks. These tools exchange the PCA-reduced space for a more general, nonlinear shared reduced space. 

In homogeneous fusion, supervised topology-preserving manifold learning techniques like Locally Linear Embedding (LLE) and semi-supervised Manifold Alignment (MA) achieve this by preserving local structure. Specifically, LLE reduces bias by capturing structural differences among image patches~\citep{6460533, XING2018165}. Semi-supervised MA builds upon LLE and has been used to align multi-temporal, multi-angle, and multi-source RS data to improve classification rates~\citep{tuia2014semisupervised}. 

Early heterogeneous fusion uses PCA as a shared feature extractor for graph-based fusion of optical-thermal-hyperspectral data~\citep{Liao2015} and HS-LiDA-R \citep{Debes2014}. kPCA improves linear, PCA-based fusion by finding a nonlinear shared space for HS-LiDAR fusion~\citep{Ghamisi2017} while supervised methods like CCA have been applied to fuse MS and LiDAR data for improved forest structure characterization~\citep{Manzanera2016}.

\paragraph{Synthesized representations with deep learning}
Nowadays, deep learning drives the data fusion paradigm through learning optimal end-to-end synthesis of remote sensing datasets. In homogenous fusion, sparse deep AEs~\citep{Huang2015} achieve high spatial resolution while mitigating spectral distortion. This was further refined by introducing independent encoders for each source~\citep{9013047}. This technique has been extended with adaptive PCA and multiscale DNNs \citep{HUANG2020115850}. In heterogeneous fusion, deep AEs have been used to integrate LiDAR, SAR, and satellite optical data to map above-ground forest biomass~\citep{Shao2017}. 

Beyond the autoencoder, a modern deep learning method, contrastive learning, has been applied to fusion tasks. When co-registered images are available, contrastive learning can encourage representations from different modalities to be similar, thereby implicitly performing fusion. This approach has shown superior performance when pretraining on Sentinel-1 and Sentinel-2 data for land cover classification~\citep{gupta2025mosaicmultimodalmultilabelsupervisionaware}.

\subsection{Analysis}
The analysis stages of the RS data value chain begin to extract real scientific meaning from the datasets. First, data visualizations explore datasets, constructing maps and identifying patterns that can be used to build hypotheses (Sec.~\ref{sec:visualization}). Then, anomaly detection identifies unexpected patterns and outliers, such as extreme events or crop failures (Sec.~\ref{sec:anomalies}). Finally, RS data is quantified in predictions, the ultimate goal of the RS data value chain, turning data understanding into actionable forecasts and scientific conclusions (Sec.~\ref{sec:predictions}). For each of these pursuits, FE is a critical tool for improving these tasks by extracting essential low-dimensional representations from complex high-dimensional data. In this review, preprocessing denotes operations that improve data quality, compactness, or cross-modal compatibility before modeling, whereas analysis denotes operations that produce interpretation, detection, or prediction from prepared data.

\subsubsection{Visualization}\label{sec:visualization}
As a picture is worth a thousand words (or, in the era of big data, even a million), visualization aims to summarize data, reveal patterns and structures, and thus extract information in a way that is easy for the human eye to interpret. We categorize FE algorithms for visualization based on the axes they reduce and the information they aim to preserve: creating interpretable maps, identifying dynamic patterns, and uncovering hidden spatial structures.

\paragraph{Spatial visualizations} In FE, spatial visualizations reduce spectral features to $1$ to $3$ bands and display them as a map, which can evolve to track changes. This enables analysis of spatial patterns, such as vegetation changes, urbanization, or cloud cover. The most straightforward approach to reducing the spectral domain is PCA, which helps generate informative color maps that outperform traditional False Color Composites (FCCs), particularly as satellite sensors become more advanced~\citep{Canas1985}. 

Spatial visualizations quickly moved beyond explicit linear FE to implicit FE that preserves topology. At first, SOMs were a common approach for visualizing HS data and have since been extended to produce a three-dimensional cube, that maps the data into an RGB subspace for enhanced visualization~\citep{gross1993visualization, tasdemir2009exploiting}. However, both SOM and PCA fail to coherently preserve both local and global structures and thus struggle with larger scenes.

Advancements in manifold learning for RS data visualization have overcome this challenge, ensuring more coherent visual representations by capturing local and global structures in the data. For example, Najim et al. demonstrate that nonlinear LLE improves cluster separation, thereby producing a more meaningful spatial visualization. Due to computational constraints, an HSI must be divided into smaller tiles, after which FCCs can be computed for each tile~\citep{Najim2023}. Finally, these FCCs must be aligned to produce one coherent FCC. Groundbreaking work by Bachmann et al. uses Isomap to produce FCCs for each tile, then uses LLE to align these tiles, thus creating coherent, structure-preserving maps of large areas~\citep{bachmann2005exploiting}.

\paragraph{Temporal visualization}  
Many applications (e.g., climate and atmospheric sciences) focus on temporal changes. FE to enable the identification of these changes often reduces the spatial and/ or spectral dimension of RS data. For example, PCA variants are standard techniques in climate science for extracting modes of climate variability—time series representing complex spatiotemporal phenomena—and identifying teleconnections — statistical dependencies between modes~\citep{Horel1981, Barnston1987}. Although ubiquitous in fields such as climate science, PCA cannot capture the nonlinear dynamics of the Earth system.

More recent approaches have extended PCA to application-specific methods that capture more subtle, complex climate variability. Variants such as the non-linear ROCK-PCA~\citep{bueso2020nonlinear} and rotated Varimax PCA~\citep{runge2015} have been used to decompose spatiotemporal datasets of different climate variables, extracting seasonality and modes of variability. By enabling greater flexibility and capturing nonlinearities, deep learning techniques have also been applied to improve climate indices. For example,~\citep{Ibebuchi} demonstrated that VAEs explain more variability in the North Atlantic Oscillation (NAO) than traditional PCA-based approaches. 

\begin{figure}[H]
    \centering
    \includegraphics[width=\linewidth]{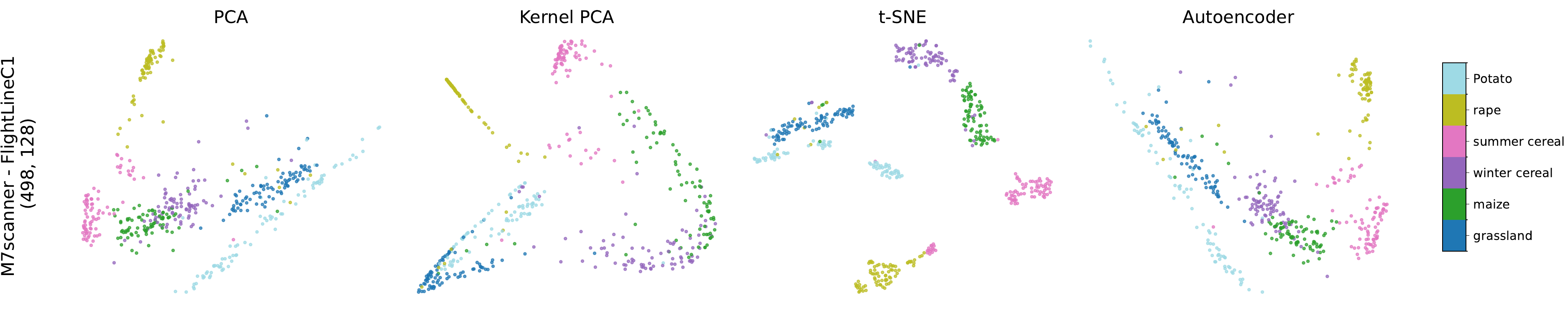}
    \caption{\textbf{Two-dimensional embeddings of spectral data generated using various feature extraction techniques.} The data originates from the HyperLabelme dataset~\citep{hyperlabelme}, specifically from the FlightLineC1 site and the M7scanner sensor. The feature extraction algorithms were trained on 498 samples, each with a spectral dimensionality of 128.}
    \label{fig:vis}
\end{figure}

\paragraph{Visualization in an abstract space}
In this setting, FE reduces one or a combination of RS data dimension samples to a $2$-$3$ dimensional space to form hypotheses about the data. We compare FE methods for visualization in an abstract space in Fig.~\ref{fig:vis}. Although FE for visualization began with unsupervised, explicit, linear FE that preserves variance, it quickly shifted to nonlinear methods that preserve the topology of the data manifold. For example, Song et al. enhanced t-SNE by integrating it with a Gaussian Mixture Model, improving its ability to represent HS data~\citep{Song2019}. 

Now, in the age of general and black-box abstract feature spaces from neural-network-parameterized deep learning models, FE can be used to interpret these features. For example, applying t-SNE to the feature spaces learned by deep learning models improves interpretability, enabling the development of hypotheses for how the model distinguishes between classes~\citep{zhang2020three, russwurm2020self}.

\subsubsection{Anomaly detection}\label{sec:anomalies}

Anomaly detection involves identifying samples that differ substantially from the majority of data within a dataset. We classify anomalies into point, collective, or contextual. Point anomalies are single instances that deviate from the rest of the data; collective anomalies consist of multiple related instances that are anomalous when combined; and contextual anomalies depend on the surrounding context for their abnormality~\citep{Chandola2009anomaly}. In RS, anomaly detection includes change detection, which focuses on identifying statistically significant differences between multitemporal observations of the same scene~\citep{Lu2004, zhu2017deep}. This connection is particularly important for spatiotemporal RS applications, where abrupt land-surface changes, infrastructure damage, or environmental disturbances appear as temporally contextual anomalies~\citep{Coppin2004}. Flach et al. demonstrated that effective feature extraction via FE can be more crucial for detecting spatiotemporal extremes than the choice of detection algorithm~\citep{Flach2017anom}. FE techniques can improve anomaly detection across spectral, spatial, and temporal dimensions because they may only become apparent in the reduced space or because of poor reconstructions. FE for anomaly detection in RS follows two distinct models: the separation model and the reconstruction model.

\paragraph{The separation model}
The separation model explicitly separates the data into background and anomaly components and is generally used for spectral anomalies. In this model, reconstruction-preserving FE methods with implicit mappings impose mathematical constraints that directly decompose the data into background and anomaly components. For example, low-rank DL models decompose HS data into a structured low-rank component and a sparse anomaly component. Spatial constraints refine this process by enforcing local consistency, ensuring that anomalies align with expected spatial patterns rather than appearing as isolated noise~\citep{TanLRR2019}. DL adapts basis functions to HS data, improving feature separation compared to PCA~\citep{NiuLRR2016}. Discriminative metric learning optimizes DL to maximize spectral contrast, enhancing robustness~\citep{Du2014anom}. Sparse representation models extend this concept by applying DL to anomaly detection in HS data~\citep{Ma2018DFL}. 

Nonlinear FE is also used in the separation model. For example, geometry-preserving manifold learning, like MDS, has successfully quantified earthquake damage by comparing pre-event optical images with post-event SAR images~\citep{Touati2018}. Other nonlinear FE directly built on DL, resulting in hybrid models like Low-Rank and Sparse Matrix Decomposition (LRaSMD) and Graph and Total Variance Regularized Low-Rank Representation (GTVLRR). These models incorporate sparse coding and structured low-rank constraints to enhance anomaly separation, making them highly effective for HS imagery~\citep{LRSM2014, Cheng2020}. Thus, in the separation model, methods have moved from DL to flexible, hybrid models that adopt deep learning components. 

\paragraph{The linear reconstruction model}
The reconstruction model is the most common anomaly detection strategy and is used for both spectral and spatio-temporal anomaly detection, including many forms of change detection.) Essentially, this model uses an explicit FE mapping and its inverse to reconstruct data, identifying anomalies as the samples with high reconstruction error. Simple, variance-preserving linear methods like PCA~\citep{Jablonski2015pca} and a combination of PCA and JPEG-2000~\citep{Du2007} highlight spectral anomalies. 

In spatio-temporal settings, these approaches are widely used to detect temporally localized deviations, including change detection. Here, FE is applied to paired or multitemporal observations to suppress nuisance variation while preserving structurally meaningful temporal differences~\citep{Lu2004}. PCA-based change detection has been used extensively~\citep{Nielsen1998,Lu2004,Turgay2009}, especially for SAR, where PCA helps mitigate speckle and isolate genuine anomalies, such as
infrastructure changes~\citep{YousifChange2013}, and land deformations\citep{Festa2023}. For temporal anomaly detection, this approach has helped define extreme weather events across European eco-regions~\citep{Mahecha2017Extremes} and spatially to isolate genuine spatial anomalies in LiDAR point clouds~\citep{duan2021low}. However, the linearity of PCA-based approaches limits their ability to model nonlinear backgrounds, and they perform poorly for more complex anomaly detection tasks.

\paragraph{The nonlinear reconstruction model}
Initial steps toward overcoming this complex background challenge include nonlinear, variance-preserving FE methods such as kPCA. Specifically, kPCA is shown to improve spectral anomaly detection in complex spectral environments~\citep{Gu2008anom}. The field quickly moved beyond such methods, adopting more flexible deep learning approaches. In this paradigm, many deep learning and AE-based approaches have emerged~\citep {bengio2013representation, Shi2024change}. However, standard AEs often generalize too well, reducing the reconstruction error for anomalies~\citep{xie2019spectral}. 

To mitigate this, these methods employ sophisticated regularizers that increase the reconstruction error for anomalies. Sparse and manifold-constrained AEs enforce feature selectivity and preserve local geometric structures, reducing redundant background reconstruction~\citep{lu2020manifold}. Transformer-based AEs model long-range dependencies through self-attention, improving feature representation in complex spectral environments~\citep{wu2024taef}. The Regularized Graph AE embeds spatial relationships via superpixel-based regularization to maintain spectral-spatial consistency~\citep{fan2021rgae}. Memory-augmented architectures leverage stored background prototypes to suppress anomaly reconstruction, improving contrast~\citep{huo2024maae}. Guided AEs incorporate spectral similarity constraints to reinforce background structure, while fully convolutional networks adjust feature learning dynamically through adaptive loss functions~\citep{xiang2021guidedAE, wang2022autoAD}.

\subsubsection{Predictions}\label{sec:predictions}

The prediction task often serves as the primary output of RS data analysis. The prediction task often serves as the primary output of RS data analysis, including classification, regression, and physically meaningful parameter retrieval (e.g., biophysical or geophysical variables inferred from observed spectra). Classical FE for predictions involved reducing the spectral dimension on a pixel-by-pixel basis, then incorporating contextual information such as the spatial distribution of pixels, and finally performing end-to-end representation learning to extract optimal spectral-spatial-contextual features.

\paragraph{Reducing spectral redundancy} 
The high spectral dimensionality of RS data (especially HS data), combined with the limited number of samples, makes classification and parameter retrieval challenging due to the redundancy of adjacent bands and pixels and the resulting ill-posedness of inverse mappings. Although PCA is a standard spectral FE method, Harsanyi and Chang's seminal work built upon it by introducing Orthogonal Subspace Projection (OSP) for simultaneous FE and classification of HS data through enhancing the signal-to-noise ratio for a desired spectral signature~\citep{harsanyi1994}. Other linear matrix factorization FE methods have been used alone or combined to perform FE to improve HS predictions~\citep{penna2007transform}. For example, unsupervised methods, like probabilistic PCA, DWT, and DCuT, and supervised methods like LDA, reduce redundancy and improve classification rates and parameter retrieval accuracy~\citep{vaddi2020probabilistic, bruce2002, garcia2017statistical, qiao2016effective, li2018discriminant}. However, these methods fail to capture nonlinear patterns in the reduced space. This is particularly important for parameter retrieval tasks, where reducing spectral redundancy improves the conditioning and stability of the inverse problem.

Nonlinear FE methods, specifically manifold learning, capture these complex features through approximating the local structure of the data manifold. For example, injecting local spectral information into LDA reduces HS spectral redundancy, thereby improving classification~\citep{li2011locality}. MFA also builds upon LDA and was further modified into a FE method called Local Geometric Structure Fisher Analysis (LGSFA), which extracts discriminatory features for improved HS classification by injecting local geometric structures~\citep{luo2017local}. Hybrid nonlinear FE methods also exist; for example, combining Isomap and LLE enhances discrimination among spectrally similar classes compared to traditional methods~\citep{bachmann2005exploiting}.

\paragraph{Capturing context} 
Considering only pixel-based information imposes a fundamental ceiling on predictive performance. To achieve the next step in FE for predictions in RS, researchers looked beyond the single pixel. They incorporated contextual information into FE, thereby improving predictive tasks that require an understanding of spatial relationships, such as object recognition or classification. At first, context was derived from spatial information, then from other sensors, and finally, from different times.

The most immediate form of context is the spatial arrangement of neighboring pixels in a single image. This spatial context was used by Liu et al. to improve object recognition in SAR images by addressing speckle-induced image distortion using a locality-preserving algorithm~\citep{liu2016synthetic}. This context is also encoded directly in the RS data cube. Thus, methods that do not flatten spatial dimensions (e.g., TD) automatically incorporate this structure~\citep{Karami2012}. 

Moving beyond single-image contexts, data fusion can serve as a feature-engineering tool to incorporate context from other images and even from different data modalities. For example, fusion competitions evaluate new FE for RS data fusion via landcover classification using the fused reduced features~\citep{Debes2014, Liao2015, Ghamisi2017}. And case studies on specific FE methods, like supervised MA, use multimodal feature fusion to improve pixel classification rates~\citep{tuia2014semisupervised}. 

Although data fusion encompasses a wide range of contexts, it overlooks the key temporal context of most RS data cubes. FE helps capture temporal information in remote sensing tasks and improve predictive capacity by extracting biophysical variables through parameter retrieval~\citep{rivera2017hyperspectral} and handling missing data in time-series~\citep{brooks2012fitting}. Rivera et al.~\citep{rivera2017hyperspectral} compare various linear FE methods and their kernel formulations for extracting features to be used as inputs to multivariate regression algorithms. Finally, when restricted to specific frequencies, the DFT can predict NDVI~\citep{brooks2012fitting}. 

Even the time dimension is considered by deep learning methods. For example, RS foundation models can take a full time series of remote sensing images as input and are therefore especially suitable for dynamic tasks such as change detection. It was shown that this enables the construction of much smaller models with similar performance~\citep{tseng2024lightweightpretrainedtransformersremote}.

\paragraph{Learning the representation}
Previous FE methods rely on hand-crafting features. The modern paradigm learns the representation itself, moving from task-specific modes to general-purpose embeddings. Early deep learning methods combine features for a single task. For example, the enhanced hybrid-graph discriminant learning (EHGDL) method builds upon LDA to improve classification accuracy by enhancing class homogeneity and reducing inter-class heterogeneity~\citep{luo2020}. By incorporating spatial context, deep learning architectures have also improved HS classification~\citep{abdi2017spectral}. Unsupervised sparse AE is used to fuse LiDAR and optical data, improving maps of forest above-ground biomass~\citep{Shao2017}. Although these were steps in the right direction for learning reduced representations, they remain largely task-specific, motivating a shift toward general-purpose representations that can support multiple downstream prediction tasks.

\subsection{From task-specific features to foundation models}

To address this limitation, deep learning has moved FE into the broader realm of \textit{representation learning}~\citep{Payandeh2023}, where the goal is no longer task-specific feature design but the learning of general-purpose embeddings. In this field, the focus shifts from merely reducing dimensions to extracting general, useful, often equally high-dimensional features that disentangle factors of variation. These rich representations are typically learned via self-supervised learning and can then be applied to various downstream tasks via transfer learning~\citep{Wang2022}. 

An important subclass of representation learning is contrastive representation learning. Unlike autoencoders, which are usually trained with a reconstruction loss, these are trained with a contrastive loss~\citep{Chen2020}. For a similarity function $\text{sim}:\mathcal{Z}\times\mathcal{Z}\rightarrow \mathbb{R}$ (e.g., the cosine similarity) and a positive pair $(\z_i,\z_j)\in \mathcal{Z}\times\mathcal{Z}$ that we want to be similar in the representation (a.k.a. reduced) space, it is defined by
\begin{align*}
    l(\z_i,\z_j) = \log\frac{\exp(\text{sim}(\z_i,\z_j)/\tau)}{\sum_{k=1}^{2N}\mathbbm{1}_{[k\neq i]}\exp(\text{sim}(\z_i,\z_k)/\tau)},
\end{align*}
with $N\in \mathbb{N}_{\geq 1}$ and $\tau>0$. All other $\z_k$ with $k\in\{1,...,2N\}\backslash\{j\}$ are chosen as negative examples with respect to $z_i$. This loss essentially encourages representations of datapoints that, in some sense, belong together to be similar, and those of points that do not belong together to be pushed apart.

These contrastive learning approaches are a major turning point because they can learn geographic context without fully labeled datasets, thereby enabling the ingestion of vast archives of unlabeled data. For RS, one forms positive pairs by different augmentations of a scene, e.g., cropped tiles~\citep{Kang2021} and different seasons~\citep{Manas_2021_ICCV}. Further extensions include geolocations to ensure that semantically similar nearby images are treated as positive pairs~\citep{Ayush2021}. Satellite contrastive location-image pretraining (SatCLIP), for instance, matches visual patterns in satellite imagery with geographic coordinates. This improves tasks such as temperature prediction and population density estimation~\citep{klemmer2023satclip}. SatCLIP is an example of a general-purpose or foundation model (FM), given its comprehensive self-supervised pretraining and potential applicability to a multitude of downstream tasks.

Nowadays, \textit{Foundation Models} (FMs) produce massive, pretrained, ready-to-use, state-of-the-art embeddings. For example, such self-supervised representation learning techniques, which utilize large neural networks and are trained on vast amounts of data, have been instrumental to the success of large language models (LLMs) for language tasks~\citep{zhao2025surveylargelanguagemodels} and are also widely adopted for vision tasks~\citep{Awais2025}. Different self-supervised learning tasks, such as MAE~\citep{Szwarcman2025}, contrastive learning~\citep{fuller2023croma}, and self-distillation~\citep{waldmann2025}, have emerged as common pretraining tasks, with some studies demonstrating their correspondence to established FE techniques~\citep{balestriero2022contrastive}. 

Such FMs are quickly and enthusiastically being adopted as FE methods to improve RS predictions~\citep{Lu2025}. Since they can be designed to inject a wide variety of data, they can produce embeddings that capture spatial, multimodal, and temporal context simultaneously. For example, powerful pre-trained representations such as the collection of FM embeddings called Major TOM~\citep{czerkawski2024} or the \href{https://developers.google.com/earth-engine/datasets/catalog/GOOGLE_SATELLITE_EMBEDDING_V1_ANNUAL}{Google Satellite Embedding} from the AlphaEarth FM~\citep{brown2025alphaearth} provide readily available, robust features. Even more recent foundation models include AnySat~\citep{astruc2025anysat} and Copernicus FM~\citep{wang2025unifiedcopernicusfoundationmodel}. Multiple benchmarks focus on evaluating the representations provided by foundation models pre-trained for multiple downstream tasks simultaneously, including burn scar, flood, and crop mapping, land use and land cover classification, and biomass estimation~\citep{Marsocci2025}. 

Although extremely promising, FMs are certainly not the solution for all FE in RS. Firstly, FMs are only as good as their input data, meaning that poor data results in poor FMs. The success of FMs over supervised deep learning baselines depends substantially on the resolution, sampling, and modalities of the pretraining data. Overall, FMs are largely black boxes, so the embeddings they produce are generally less interpretable than those from standard FE methods, leading to a dangerous lack of trustworthiness.

\subsection{Synopsis}
As we traveled through the uses of FE in the RS data value chain, it became clear that RS tasks have shifted from specific to more general. To support more general, complex tasks, FE methods have moved from linear to nonlinear to extremely general trained embeddings, often obtained by FMs. This begs the question: does traditional FE still have a place in RS? We will address this in Sec.~\ref{sec:perspective}.

\section{Evaluation metrics for feature extraction}\label{sec:eval}

A common thread through FE in RS is the evaluation of FE methods, an essential step to determining the optimal FE method for a given task. Although FE in RS has evolved, FE evaluation in RS has largely remained the same, as it depends heavily on the downstream RS task. Thus, we provide a collection of the most common metrics from the works surveyed in this review, sorted by RS task in Tab.~\ref{tab:evaluation_metrics_summary} and an organized bibliography linking each article to its RS task and evaluation metrics in Tab.~\ref{tab:metrics_cites}.

\begin{figure}[H]
    \centering
    \includegraphics[width=\linewidth]{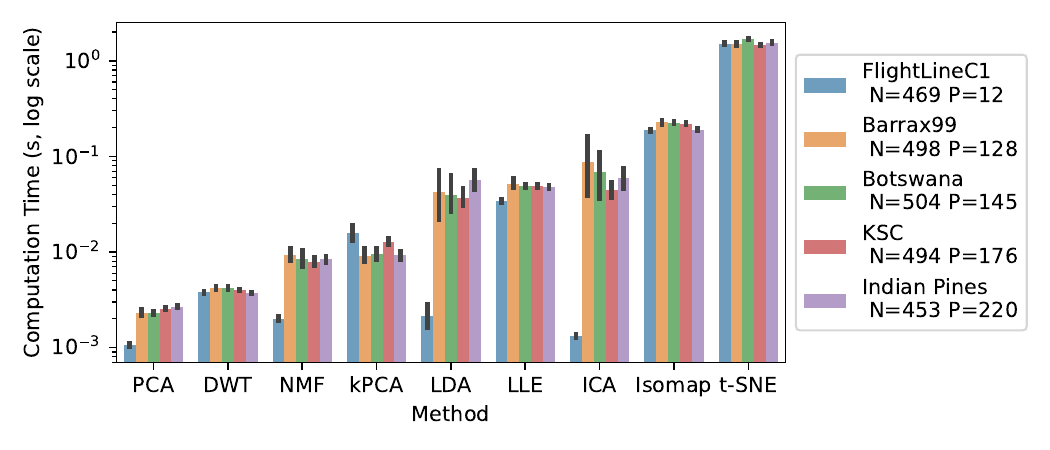}
    \caption{\textbf{Computation times for common FE algorithms for different HyperLabelMe datasets~\citep{hyperlabelme}.} We evaluate unsupervised methods and one supervised method for classification datasets (LDA) to reduce from P to $K=2$ dimensions. All methods are run on a $2020$ MacBook Pro with M1 chip and $16$GB of memory.}
    \label{fig:times}
\end{figure}

There are two universal metrics: visualization and computation time. Visualization is a useful qualitative metric. For example, suppose a reconstructed image is shifted to the right by one pixel, yielding low correlation and a higher mean squared error with the original image. Still, visually, it might be an acceptable reconstruction that captures the original image's structure. On the other hand, computation time provides a practical understanding of how quickly FE methods can be executed relative to one another. We compare the computation time of the most common FE methods to reduce the spectrum of various HS images in Fig.~\ref{fig:times}—the more complex the FE method, the higher the computational cost. Specifically, supervised methods (LDA) are slower than unsupervised methods, and nonlinear methods (kPCA, Isomap, and t-SNE) are slower than linear methods.
For deployment-focused evaluation, computation should be interpreted jointly with memory footprint, hardware constraints, and inference latency, especially for onboard and edge remote sensing settings.

Outside of computation time, evaluation metrics for compression and denoising evaluation metrics compare an original sample $\x \in \R^P$ to its reconstruction $\hat{\x} = \boldsymbol{\psi} \circ \boldsymbol{\phi} (\x) \in \R^P$ to assess reconstruction quality. These metrics assess the reconstruction shape and/or scale and are separated into proxies for similarity (e.g., correlation) and proxies for error (e.g., mean squared error). Many of the aforementioned metrics are also used to evaluate data fusion. However, in the surveyed works, it was found that specific tools like Universal Image Quality Index (UIQI)~\citep{wang2002universal} and Relative Dimensionless Global Error in Synthesis (ERGAS)~\citep{wald1997fusion} are 

\input{tables/metrics}

\noindent used almost exclusively for evaluating FE for data fusion.

FE for compression, denoising, and fusion is frequently evaluated by the capacity of the reduced or reconstructed data to perform anomaly detection and/or prediction. Anomaly detection is essentially a binary classification problem and is evaluated by plotting the Receiver Operator Characteristic (ROC) curve and measuring the Area Under the ROC Curve (AUC). Predictions are partitioned into classification and regression, and each is evaluated differently, depending on the data characteristics.

\subsection{Synopsis}
Evaluation of FE in the RS data value chain is highly task-specific. So, either there is a need for metrics for general FE evaluation in RS, or we must accept that there is no sufficiently general FE evaluation method for all RS tasks.

\section{Trends and perspectives}\label{sec:perspective}

Across the data value chain, we observe that earlier works began using linear unsupervised methods, such as PCA, for FE. Recently, FE in RS has shifted towards deep learning and foundation models for complex, nonlinear, and general features. We outline this paradigm shift in Sec.~\ref{sec:fm_perspective}. Then, we identify three perspectives on FE in this new era, arising from trends extracted from our survey. Specifically, we identify a bridge between deep learning and classical FE in Sec.~\ref{sec:classical_perspective}, and then we emphasize the need for interpretable embeddings in Sec.~\ref{sec:interp_perspective}. A glossary of the FE methods discussed in this section is in Tab.~\ref{tab:dr_glossary_perspectives}.

\subsection{From single-task models to unified representations}\label{sec:fm_perspective}
Deep learning for specific tasks and simple multimodal fusion are becoming state-of-the-art for learning joint representations of RS data. This trend points toward a grander future carried on the shoulders of large-scale, multi-modal Foundation Models (FMs) that may provide the ultimate unification of RS data.

Standard methods for RS compression, such as JPEG-2000, are multi-step pipelines that include FE steps, such as the DWT. JPEG-AI outperforms JPEG-2000 and unifies data compression into a single step using advances in deep learning~\citep{ascenso2023jpeg}. Although JPEG-AI remains largely untested in RS, deep learning-based compression methods for Earth observation, such as

\input{tables/dr_perspective}

\noindent TerraCodec~\citep{costa2025terracodec}, are already revolutionizing RS data compression. 

Deep learning methods have been adapted to a wide range of data structures, such as images, time series, and graphs. They are thus a clear candidate for a structure that can unify different RS modalities. Contrastive learning leverages pairings of different modalities such as co-located optical and SAR imagery~\citep{fuller2023croma}, and Masked Autoencoders (MAE) reconstruct masked modalities from the remaining ones to learn joint representations ~\citep{Szwarcman2025}. For instance, MAEs could reconstruct a masked optical patch from its SAR counterpart. In general, we can use these methods as an implicit form of fusion. 

The goal of FMs is to bring this unification to its final form, integrating as many modalities as possible, which could eventually lead to a fusion of RS, climate, weather, and in situ data~\citep{zhu2024foundationsearthclimatefoundation}. 
Recent FMs integrate, for instance, SAR with optical satellite imagery~\citep{xiong2024} and use diffusion to generate missing modalities from the unified representation~\citep{jakubik2025terramindlargescalegenerativemultimodality}.

Despite their promise, the deep learning-derived representations of FMs have not yet fully lived up to expectations in RS applications~\citep{ramospollan2024}. They are often beaten by standard deep-learning baselines in segmentation or regression tasks~\citep{Marsocci2025}, which can be attributed to a mismatch in resolution and modalities or poor sampling of the pretraining data. Recent work has developed a mechanism using FMs for RS to flag potential failures in advance~\citep{SHRUG-FM}. However, the question of why these failures exist remains open and a promising avenue for future research.

\subsection{Bridging classical FE and modern representation learning}\label{sec:classical_perspective}
Another issue with FM embeddings is that we no longer struggle with the raw sensor dimensions; we struggle to understand and interpret the complex, high-dimensional embeddings produced by these models. Thus, FMs essentially trade the original raw RS dimensionality problem with another. Luckily, we suggest addressing this issue via bridging classical FE with modern representation learning. We propose two bridges: one that uses classical FE on FM embeddings, and a second bridge for developing hybrid classical FE-FM methods.

FE can be used to interpret FM embeddings and provide task-specific representations when such embeddings are too general. These roles are most evident in visualization, where methods such as t-SNE have already been applied to feature spaces of deep learning models~\citep{zhang2020three, russwurm2020self}. Other nonlinear topology-preserving FE methods, such as Uniform Manifold Approximation and Projection (UMAP), can also be used to explore these latent spaces, for instance, to identify distinct crop types within a supposedly monolithic agricultural class or to track the phenological evolution of a forest over time~\citep{mcinnes2018umap}. Recently, an accelerated version of t-SNE, negative or mean affinity discrimination (NOMAD) projection~\citep{duderstadt2025nomad}, enables the visualization of extremely large-scale RS datasets and FM embeddings. 

Two glaring limitations of t-SNE and UMAP are that they are computationally expensive and implicit, meaning they cannot be applied to test data or used to reconstruct data. In general, FE families like manifold learning suffer from implicit FE mappings, meaning that these methods are not readily transferable to test or out-of-distribution datasets. To remedy this issue, we consider a simple yet ambitious framework for reformulating implicit FE mappings into explicit ones. Specifically, we propose hybrid FE methods that trade off between optimizing a reduced-space embedding and optimizing neural network parameters to explicitly map to the embedding. Damrich et al. explored this concept with t-SNE and UMAP, finding that both can be formulated as explicit FE~\citep{damrich2022t}. Another perspective of hybrid FE-DL methods is a variant of t-SNE called t-SimCNE~\citep{bohm2023unsupervised}. AEs have also been adapted to respect non-Euclidean structures in data through graphs~\citep{kipf2016variational} and hypergraphs~\citep{fan2021heterogeneous}. The most modern developments in this direction include simplicial and combinatorial complexes.~\citep{papamarkou2024position} Research continuing in this direction will certainly produce more scalable and interpretable visualizations of large-scale RS datasets, facilitating exploratory analysis and downstream applications.
 
\subsection{The Rise of Black Boxes and the Need for Robustness and Interpretability}\label{sec:interp_perspective}
The pursuit of performance with black box models has created two critical failures: a lack of robustness to imperfect data and a lack of scientific understanding of the models and their predictions. We can address the prior problem by going back to the basics with simple, robust, and principled variations of PCA that are still under-explored on RS datasets, which may improve both denoising and anomaly detection. For example, Robust PCA~\citep{candes2011robust}, applied to tasks analogous to foreground/background separation, such as cloud and shadow removal or unsupervised change detection, shows potential for both denoising in the presence of outliers and anomaly detection. Furthermore, Robust subspace recovery~\citep{lerman2018overview} and dual principal component pursuit~\citep{tsakiris2018dual} are used for outlier rejection and robust model fitting in computer vision, and are rarely used in RS. 

Moving past model robustness, a modern line of research aims for result \textit{explainability}, which focuses on explaining a specific outcome based on the extracted features~\cite{hohl2024opening}. A practical workflow for explainable AI involves applying feature importance propagation methods to a downstream model's predictions. Techniques like SHAP or LIME can quantify the contribution of each extracted feature to a particular outcome, providing local, instance-specific explanations. For a land cover classification model, this could reveal which latent features contributed most to labeling a specific pixel as `wetland,' allowing a hydrologist to verify the model's reasoning. Complementing this, \textit{uncertainty quantification} (UQ) provides another critical layer of explainability. By assessing the model's confidence, UQ helps identify unreliable predictions, effectively explaining that a result should not be trusted. Together, these post-hoc explanations and uncertainty measures form a robust framework for building trust and enabling scientific validation of features extracted from complex RS data, even when the FE model itself remains a black box.

Our next step is to build \textit{interpretable} FE methods~\citep{Carvalho2019}. For example, a curse of isotropy has been uncovered in PCA, indicating that often PCs should be grouped into principal subspaces~\citep{szwagier2024curseisotropyprincipalcomponents}. This means that single principal components from PCA used in climate science as proxies for climate indices (e.g., ENSO) may oversimplify complex climate variability. Rotating a single vector in this subspace to align with the physical process yields more interpretable modes of variability than those derived from a single principal component.

Although robust, explainable, and interpretable FE methods are surely valuable, the gold standard of modern FE should achieve \textit{physical and causal understanding} of the models and their extracted dimensions~\citep{camps2026ai}. Most feature extraction methods lack any grounding in real physical processes. Hybrid modeling and physics-informed machine learning have emerged as promising research directions in which physical relations can be learned or incorporated into machine learning models~\citep{karniadakis2021physics}. FMs and traditional FE methods alike fail to separate causal features from spurious correlations, limiting generalization under domain shifts, compromising robustness and explainability~\citep{Tuia21perspective}. Causality-aware FE methods address this by disentangling actual signals from biases, thereby ensuring robust and transferable representations~\citep{buhlmann2020invariance}. In applications, this makes RS tasks, such as prediction, more robust and improves generalization. For example, a crop yield prediction model trained on data from one region would generalize better to a new region if it learned the causal link between soil moisture and growth, rather than just correlating yield with the number of satellite overpasses. For example, linear causality-aware FE adapts PCA to detect Granger causal directions via Granger PCA (gPCA)~\citep{varando2022learning}. Furthermore, deep learning FE has been injected with causal reasoning, resulting in methods like Causal Feature Learning (CFL)~\citep{chalupka2017causal} and Texture-Aware Causal Feature Extraction Network (TeACFNet)~\citep{xu2024texture}. Using these methods, we can detect causal features that improve the generalization of RS models under domain shifts (e.g., climate change and data from new sensors).

\subsection{Synopsis}
Although FMs show promise as large general FE, their embeddings are often still high-dimensional, and their black-box nature makes them difficult to interpret. In the future, FE in RS will remain task-specific by using FE on FM embeddings, building hybrid FE-FM methods, and advancing the use of robust, interpretable, and potentially causal FE methods in RS.

\section{Conclusions}
After providing a framework of standard FE methods in RS, we have traversed the entire RS data value chain and exposed the current utility of FE for each challenge. This voyage revealed the shift in FE in RS from unsupervised linear methods to the foundation-model era. As researchers facing this bold new paradigm, there is a real possibility of unifying all RS data into a cohesive representation. However, these general FM embeddings risk scientific opacity by prioritizing benchmark performance over physical structure, and potentially decouple RS models from true physical meaning. Thus, we conclude with three claims: FM embeddings are not sufficient for RS science, classical FE is not obsolete in RS, but complementary to FMs, and finally, robustness and interpretability are essential pillars of FE for RS.

In summary, we posit that the FM era should not be a full replacement of classical FE, but a reconfiguration, a synergistic incorporation of classical FE with modern foundation models. FE for RS should focus on building hybrid pipelines that couple standard FE with FMs for extracting compact, task-relevant features. This direction is necessary to achieve more efficient, trustworthy, and explainable feature representations of Earth system data.

\section*{Acknowledgements}
N.M. and G.C-V. acknowledge the support of Generalitat Val\`encia and the Conselleria d'Innovaci\'o, Universitats, Ci\`encia i Societat Digital, through the project ``AI4CS: Artificial Intelligence for complex systems: Brain, Earth, Climate, Society'' (CIPROM/2021/56). 
K-H. C. \& G.C-V. acknowledge the support from the European Research Council (ERC) under the ERC Synergy Grant USMILE (grant agreement 855187).
G.C-V. acknowledges the support from the HORIZON program under the AI4PEX project (grant agreement 101137682). 

\section*{Declaration of generative AI and AI-assisted technologies in the manuscript preparation process}

During the preparation of this work, the authors used ChatGPT and Google Gemini in order to edit and summarize content. After using this tool/service, the authors reviewed and edited the content as needed and take full responsibility for the content of the published article.

\clearpage
\newpage


\bibliographystyle{elsarticle-harv}
\bibliography{references.bib}

\clearpage
\newpage

\appendix
\section{Supplementary Material}
\label{app1}
A list of standard Feature Extraction (FE) methods in Remote Sensing (RS) is in Tab.~\ref{tab:dr_glossary_standard}. 
The dimensionality sources for different RS sensors are in Tab.~\ref{tab:rs_mod}. We provide three look-up tables that organize our references. Tab~\ref{tab:master_summary_cites} lists references sorted by their corresponding stage in the RS data value chain. Tab.~\ref{tab:task_method_matching_new} sorts references for FE in RS by the standard FE methods used to address each challenge, sorted by property preservation. Next, Tab.~\ref{tab:metrics_cites} organized the references of evaluation metrics for FE in RS by RS task.

\input{tables/rs_mod}

\input{tables/dr_glossary_standard.tex}

\input{tables/master_summary_cites}

\input{tables/task_method_matching_new}
\clearpage
\newpage
\input{tables/metrics_applications}

\end{document}

%% file: tables/master_summary.tex
\begin{table}[H]
    \centering
    \caption{\textbf{A summary of how, within our framework, FE addresses each challenge in the RS data value chain.} First, we decompose each RS challenge into stages. Then we use our framework for FE in RS to identify the common FE characteristics for each stage, namely the dataset, mapping, and properties preserved. Next, we list common FE methods at each stage. We abbreviate the following mapping mechanisms as: matrix factorization (MF), kernel (Ker), neural network (NN), and manifold learning (Man). A list of references for each stage in the RS data value chain can be found in Tab.~\ref{tab:master_summary_cites}.}
    \resizebox{\textwidth}{!}{\begin{tabular}{>{\raggedright\arraybackslash}p{2.5cm}>{\raggedright\arraybackslash}p{3.5cm}>{\raggedright\arraybackslash}p{2.5cm}>{\raggedright\arraybackslash}p{4cm}>{\raggedright\arraybackslash}p{4cm}>{\raggedright\arraybackslash}p{3.5cm}}
        \toprule
        \textbf{Challenge} & \textbf{Stage} & \textbf{Dataset} & \textbf{Mapping} & \textbf{Properties preserved} & \textbf{FE Methods} \\
        \midrule
        Compression & \cellcolor{black!10} Flatten \& Compress & \cellcolor{black!10} Unsupervised &\cellcolor{black!10} Linear, MF & \cellcolor{black!10} Frequency, variance, reconstruction, distribution & \cellcolor{black!10} DFT, DWT, ICA, NMF, PCA \\ 
        & Removing autocorrelation & Unsupervised & Explicit, linear, MF & Frequency, variance & DWT, PCA \\
        & \cellcolor{black!10} Flexible, multidimensional compression & \cellcolor{black!10} Unsupervised & \cellcolor{black!10} Explicit, linear, nonlinear, MF, NN & \cellcolor{black!10} Reconstruction & \cellcolor{black!10} AE, TD \\
        
        \midrule
        Data Cleaning & Image restoration, enhancement and denoising & Unsupervised & Explicit, implicit, linear, nonlinear, MF, NN & Frequency, variance, reconstruction & DWT, MNF, PCA, AE \\ 
        & \cellcolor{black!10} Gap Filling   & \cellcolor{black!10} Unsupervised & \cellcolor{black!10} Explicit, implicit, linear, nonlinear, MF, NN & \cellcolor{black!10} Frequencies, variance, reconstruction & \cellcolor{black!10} AE, DCT, DL, PCA  \\
        
        \midrule
        
        Fusion & Component Substitution & Unsupervised & Explicit, linear, MF & Frequency, variance &  DWT, PCA \\ 
        & \cellcolor{black!10} Alignment of a Shared Reduced Space & \cellcolor{black!10} Unsupervised, supervised & \cellcolor{black!10} Explicit, implicit, linear, nonlinear, MF, Ker, Man & \cellcolor{black!10} Variance, reconstruction, topology & \cellcolor{black!10} CCA, kPCA, LLE, PCA, MA \\
        & Synthesized Representations with Deep Learning & Unsupervised, supervised & Explicit, nonlinear, NN & Reconstruction & AE \\

        \midrule
        
        Visualization  & \cellcolor{black!10} Spatial & \cellcolor{black!10} Unsupervised & \cellcolor{black!10} Explicit, implicit, linear, nonlinear, MF, Man & \cellcolor{black!10} Variance, reconstruction, topology & \cellcolor{black!10} PCA, SOM, LLE \\ 
        & Temporal & Unsupervised & Explicit, linear, nonlinear, MF, NN & Variance, reconstruction, distribution & PCA, VAE \\ 
        & \cellcolor{black!10} Abstract & \cellcolor{black!10} Unsupervised & \cellcolor{black!10} Implicit, nonlinear, Man & \cellcolor{black!10} Geometry, topology & \cellcolor{black!10} Isomap, LLE, t-SNE\\
        
        \midrule
        
        Anomaly Detection & Separation & Supervised & Implicit, linear, nonlinear, MF, Man & Reconstruction, geometry & DL, MDS \\
        & \cellcolor{black!10} Linear Reconstruction & \cellcolor{black!10} Unsupervised & \cellcolor{black!10} Explicit, linear, MF & \cellcolor{black!10} Frequency, variance & \cellcolor{black!10} DWT, PCA \\
        & Nonlinear Reconstruction & Unsupervised & Explicit, implicit, nonlinear, NN, Ker& Variance, reconstruction& AE, kPCA, MDS, CCA \\
        
        \midrule
        
        Predictions & 
        \cellcolor{black!10} Reducing Spectral Redundancy & \cellcolor{black!10} Unsupervised, supervised & \cellcolor{black!10} Explicit, implicit, linear, nonlinear, MF, Man & \cellcolor{black!10} Frequency, variance, reconstruction, geometry, topology & \cellcolor{black!10} DCuT, DWT, Isomap, LFDA, LLE, PCA\\ 
        & Capturing Context & Unsupervised, supervised & Explicit, implicit, inear, nonlinear, MF, Ker, Man & Frequency, variance, reconstruction, topology & DFT, kPCA, MA, TD \\ 
        & \cellcolor{black!10} Learning Representations & \cellcolor{black!10} Unsupervised, supervised & \cellcolor{black!10} Explicit, nonlinear, NN & \cellcolor{black!10} Reconstruction & \cellcolor{black!10} AE\\
        \bottomrule
    \end{tabular}}
    \label{tab:master_summary}
\end{table}

%% file: tables/metrics.tex
\clearpage
\newpage
\begin{table}[H]
\centering
\caption{\textbf{A task-specific guide to evaluation metrics in FE for RS.} This table summarizes the most common metrics, categorized by their purpose and the primary RS task they serve. A table linking RS tasks to evaluation metrics is in Tab.~\ref{tab:metrics_cites}.}
\label{tab:evaluation_metrics_summary}
\renewcommand{\arraystretch}{1.3}\resizebox{\textwidth}{!}{\begin{tabular}{>
    {\raggedright\arraybackslash}p{3cm}>
    {\raggedright\arraybackslash}p{3cm}>
    {\raggedright\arraybackslash}p{5cm}>
    {\raggedright\arraybackslash}p{2.5cm}>
    {\raggedright\arraybackslash}p{4cm}}
\toprule
\textbf{RS Task} & \textbf{Metric} & \textbf{What it Measures} & \textbf{Category} & \textbf{Primary Goal}\\
\midrule

\multirow{5}{=}{\textbf{Universal (All Tasks)}} & \cellcolor{black!10} Visualization (VIS) & \cellcolor{black!10} Qualitative assessment of results (e.g., embeddings, reconstructions) & \cellcolor{black!10} Qualitative & \cellcolor{black!10} Human interpretation \& sanity checks \\
& Computation Time (CT) & Algorithmic efficiency and resource usage & Quantitative & Assess practical feasibility \& scalability \\
\midrule

\multirow{13}{=}{\textbf{Compression \& Denoising}} & \cellcolor{black!10} Correlation Coeff. (CC) & \cellcolor{black!10} Shape similarity between two signals, invariant to scale & \cellcolor{black!10} Similarity (Shape) & \cellcolor{black!10} Check pattern/structure preservation \\
& Signal-to-Noise Ratio (SNR) & Ratio of signal power to noise power & Similarity (Quality) & Assess reconstruction fidelity/scale \\
& \cellcolor{black!10} Peak SNR (PSNR) & \cellcolor{black!10}Distortion relative to the maximum possible signal value & \cellcolor{black!10} Similarity (Quality) & \cellcolor{black!10} Standard for reconstruction quality \\
& Mean Squared Error (MSE) & Average squared magnitude difference between pixels & Error (Magnitude) & Fundamental reconstruction error \\
& \cellcolor{black!10} Spectral Angle Dist. (SAD) & \cellcolor{black!10} Angle between two spectra, invariant to brightness & \cellcolor{black!10} Error (Shape) & \cellcolor{black!10} Evaluate spectral signature fidelity \\
& Rate Distortion (RD/BR) & Curve of reconstruction quality vs. compression level (bit rate) & Performance Curve & Compare algorithm compression efficiency \\
\midrule

\multirow{5}{=}{\textbf{Data Fusion}} & \cellcolor{black!10} ERGAS & \cellcolor{black!10} Relative global error, assessing radiometric and spectral quality & \cellcolor{black!10} Error (Global Quality) & \cellcolor{black!10} Standardized quality score for fused products \\
& UIQI & Combination of structural, luminance, and contrast similarity & Similarity (Structural) & Assess perceptual/visual quality of fusion \\
\midrule

\multirow{2}{=}{\textbf{Anomaly Detection}} & \cellcolor{black!10} ROC / AUC & \cellcolor{black!10} Trade-off between true positive rate and false positive rate & \cellcolor{black!10} Classification Perf. & \cellcolor{black!10} Evaluate detection sensitivity vs. false alarms \\
\midrule

\multirow{7}{=}{\textbf{Predictions}} & Accuracy (ACC) & Overall percentage of correct classifications & Classification Perf. & Simple baseline (can mislead on imbalanced data) \\
& \cellcolor{black!10} F1-score / Cohen's Kappa ($\kappa$) & \cellcolor{black!10} Metrics robust to class imbalance (precision/recall, agreement) & \cellcolor{black!10} Classification Perf. & \cellcolor{black!10} Robust evaluation of classifier performance \\
& R-squared ($R^2$) & Proportion of variance in the target variable explained by the model & Regression Perf. & Evaluate regression model fit and performance \\

\bottomrule
\end{tabular}}
\end{table}
\clearpage
\newpage

%% file: tables/dr_perspective.tex
\begin{table}[H]
    \centering
    \caption{\textbf{Perspective FE methods for RS.} These methods are organized by our three perspectives for FE in RS.}
\resizebox{\textwidth}{!}{\begin{tabular}{>{\raggedright\arraybackslash}p{3cm} > {\raggedright\arraybackslash}p{2.5cm} >{\raggedright\arraybackslash}p{6.5cm} >{\raggedright\arraybackslash}p{4cm}}
        \toprule
        \textbf{Perspective} & \textbf{Abbreviation} & \textbf{Method} & \textbf{Reference}\\
        \toprule
        \multirow{14}{=}{From single-task models to unified representations} & \cellcolor{black!10} AlphaEarth & \cellcolor{black!10} Google Satellite Embedding & \cellcolor{black!10} \citep{brown2025alphaearth} \\
        & AnySat & AnySat & \citep{astruc2025anysat} \\
        & \cellcolor{black!10} CL & \cellcolor{black!10} Contrastive Learning & \cellcolor{black!10} \citep{fuller2023croma}\\
        & Copernicus FM & Copernicus Foundation Model & \citep{wang2025unifiedcopernicusfoundationmodel}\\ 
        & \cellcolor{black!10} JPEG-AI & \cellcolor{black!10} JPEG-AI  & \cellcolor{black!10} \citep{ascenso2023jpeg} \\ 
        & MAE & Masked Autoencoder & \citep{Szwarcman2025} \\
        & \cellcolor{black!10} Major-TOM & \cellcolor{black!10} Terrestrial Observation Metaset & \cellcolor{black!10} \citep{czerkawski2024}\\
        & MoCo & Momentum Contrast & \citep{Kang2021}  \\
        & \cellcolor{black!10} Sat-CLIP & \cellcolor{black!10} Satellite Contrastive Location-Image Pretraining  & \cellcolor{black!10} \citep{klemmer2023satclip}\\
        & SD & Self-distillation & \citep{waldmann2025}\\
        & \cellcolor{black!10} TEC & \cellcolor{black!10} TerraCodec & \cellcolor{black!10} \citep{costa2025terracodec} \\
        \midrule
        \multirow{2}{=}{Bridging classical FE and modern representation learning} & NOMAD & Negative or mean affinity discrimination & \citep{duderstadt2025nomad}\\
        & \cellcolor{black!10} t-SimCNE & \cellcolor{black!10} t-SimCNE & \cellcolor{black!10} \citep{bohm2023unsupervised}\\
        & UMAP & Uniform Manifold Approximation and Projection & \citep{mcinnes2018umap}\\
        \midrule
        \multirow{16}{=}{The rise of black boxes and the need for robustness and interpretability} &  \cellcolor{black!10} CFL &  \cellcolor{black!10} Causal Feature Learning &  \cellcolor{black!10} \citep{chalupka2017causal}\\
        & DPCP & Dual PC Pursuit & \citep{tsakiris2018dual} \\ 
        &  \cellcolor{black!10} EHGDL &  \cellcolor{black!10} Enhanced Hybrid-Graph Discriminant Learning  &  \cellcolor{black!10} \citep{luo2020}\\
        & gPCA & Granger Principal Component Analysis & \citep{varando2022learning}\\
        &  \cellcolor{black!10} LGSFA &  \cellcolor{black!10} Local Geometric Structure Fisher Analysis   &  \cellcolor{black!10} \citep{luo2017local}\\
        & PAA & Piecewise Aggregate Approximation & \citep{keogh2001dimensionality}\\
        &  \cellcolor{black!10} PSA &  \cellcolor{black!10} Principal Subspace Analysis &  \cellcolor{black!10} \citep{szwagier2024curseisotropyprincipalcomponents}\\
        & RSR & Robust Subspace Recovery & \citep{lerman2018overview}\\
        &  \cellcolor{black!10} TeACFNet &  \cellcolor{black!10} Texture-Aware Causal Feature Extraction Network &  \cellcolor{black!10} \citep{xu2024texture} \\
        \bottomrule
    \end{tabular}}
    \label{tab:dr_glossary_perspectives}
\end{table}


%% file: tables/rs_mod.tex
\begin{table}[!ht]
\centering
\caption{\textbf{Sources of data dimensionality across common remote-sensing sensors.}}
\footnotesize{
\begin{tabular}{ll}
\toprule
\textbf{Sensor Type} & \textbf{Dimensionality Sources}\\
\toprule

\multirow{3}{*}{Optical / Multispectral/ Hyperspectral} & \cellcolor{black!10}spatial resolution \\
& spectral bands \\
& \cellcolor{black!10}temporal revisits \\
\hline

\multirow{4}{*}{Thermal Infrared Imager}
& spatial resolution \\
& \cellcolor{black!10}thermal bands \\
& temperature sensitivity \\
& \cellcolor{black!10}temporal revisits \\
\hline

\multirow{4}{*}{Passive Microwave Radiometer}
& footprint size \\
& \cellcolor{black!10}multiple centre frequencies \\
& polarisation \\
& \cellcolor{black!10}temporal coverage \\
\hline

\multirow{4}{*}{Atmospheric Spectrometer / Sounder}
& vertical profile levels \\
& \cellcolor{black!10}spectral resolution \\
& along-track sampling \\
& \cellcolor{black!10}temporal coverage \\
\hline

\multirow{5}{*}{SAR / Radar}
& frequency band \\
& \cellcolor{black!10}polarisation \\
& phase/coherence \\
& \cellcolor{black!10}incidence angle \\
& temporal stacks \\
\hline

\multirow{4}{*}{LiDAR} 
& \cellcolor{black!10}point density \\
& 3D geometry \\
& \cellcolor{black!10}multiple returns \\
& waveform samples\\
\bottomrule
\end{tabular}}
\label{tab:rs_mod}
\end{table}

%% file: tables/dr_glossary_standard.tex
\begin{table}[H]
    \centering
    \caption{\textbf{Standard FE methods for RS with abbreviations and references.}}
\resizebox{\textwidth}{!}{\begin{tabular}{>{\raggedright\arraybackslash}p{1.2cm} >{\raggedright\arraybackslash}p{6.5cm} >{\raggedright\arraybackslash}p{7cm}}
        \rowcolor{gray!10} AE    & Autoencoder                                 & \citep{bank2023autoencoders}                                                     \\
        CCA   & Canonical Correlation Analysis             & \citep{yang2019survey}                                                            \\
        \rowcolor{gray!10} DCT & Discrete Cosine Transform & \citep{ahmed1974discrete} \\
        CLIP  & Contrastive Language Image Pre-training    & \citep{radford2021learning}                                                       \\
        \rowcolor{gray!10} DCuT  & Discrete Curvelet Transform              & \citep{candes1999curvelets}                                                         \\
        DFT   & Discrete Fourier Transform                  & \citep{duhamel1990fast}                                                          \\
        \rowcolor{gray!10} DL    & Dictionary Learning                         & \citep{kreutz2003dictionary}                                                     \\
        DWT   & Discrete Wavelet Transform                  & \citep{broughton2018discrete}                                                    \\  
        \rowcolor{gray!10} EOF   & Empirical Orthogonal Functions              & \citep{Hannachi2007}                                                             \\
        GDA   & Generalized Discriminant Analysis           & \citep{baudat2000generalized}                                                    \\
        \rowcolor{gray!10} ICA   & Independent Component Analysis              & \citep{jutten1991blind}                                                          \\
        Isomap & Isometric Feature Mapping                   & \citep{balasubramanian2002isomap}                                                \\
        \rowcolor{gray!10} kCCA & Kernel Canonical Correlation Analysis        & \citep{akaho2006kernel}                                                          \\
        kMNF  & Kernel Maximum Noise Fraction               & \citep{nielsen2010kernel}                                                        \\
        \rowcolor{gray!10} kPCA  & Kernel Principal component analysis         & \citep{scholkopf1997kernel}                                                      \\
        kPLS & Kernel Partial Least Squares                 & \citep{rosipal2001kernel}                                                        \\
        \rowcolor{gray!10} LDA  & Linear Discriminant Analysis                 & \citep{bandos2009classification}                                                 \\
        LLE & Locally Linear Embedding                      & \citep{saul2000introduction}                                                     \\
        \rowcolor{gray!10} MA  & Manifold Alignment                            & \citep{ham2005semisupervised}                                                    \\
        MDS & Multidimensional Scaling                      & \citep{saeed2018survey}                                                          \\
        \rowcolor{gray!10} MFA & Marginal Fisher Analysis                      & \citep{yan2006graph}                                                             \\
        MNF  & Maximum Noise Fraction                       & \citep{green1988transformation}                                                  \\
        \rowcolor{gray!10} NMF  & Non-negative Matrix Factorization            & \citep{wang2012nonnegative}                                                      \\
        OSP  & Orthogonal Subspace Projection               & \citep{harsanyi1994}                                                             \\
        \rowcolor{gray!10} PCA  & Principal Component Analysis                 & \citep{hotelling1933analysis}                                                    \\
        PLS  & Partial Least Squares                        & \citep{haenlein2004beginner}                                                     \\
        \rowcolor{gray!10} POD  & Proper Orthogonal Decomposition              & \citep{chatterjee2000introduction}                                               \\
        SOM & Self-organizing Maps                          & \citep{kohonen1990self}                                                          \\
        \rowcolor{gray!10} SSA  & Singular Spectrum Analysis                   & \citep{vautard1989singular}                                                      \\
        TD    & Tensor Decomposition                        & \citep{wang2023tensor}                                                           \\
        \rowcolor{gray!10} t-SNE & t-Distributed Stochastic Neighbor Embedding & \citep{van2008visualizing}                                                       \\
        VAE   & Variational Autoencoder                     & \citep{kingma2013auto}                                                           \\
    \end{tabular}}
    \label{tab:dr_glossary_standard}
\end{table}


%% file: tables/master_summary_cites.tex
\begin{table}[H]
    \centering
    \caption{\textbf{References at each stage of the RS data value chain.}}
    \resizebox{\textwidth}{!}{\begin{tabular}{>{\raggedright\arraybackslash}p{2.5cm}>{\raggedright\arraybackslash}p{3.5cm}>{\raggedright\arraybackslash}p{14cm}}
        \toprule
        \textbf{Challenge} & \textbf{Stage} & \textbf{References}\\
        \midrule
        Compression & \cellcolor{black!10} Flatten \& Compress & \cellcolor{black!10} \citep{benz1995comparison, kaarna2000compression, skodras2001jpeg, wang2006independent, serra2008remote}\\ 
        & Removing autocorrelation & \citep{penna2006progressive, Du2007, penna2007transform}\\
        & \cellcolor{black!10} Flexible, multidimensional compression & \cellcolor{black!10} \citep{Karami2012,garcia2017statistical,xiang2024remote,pellicer2025video}\\
        
        \midrule
        Data Cleaning & Image restoration, enhancement and denoising & \citep{chen2010denoising,duan2021low, luo2016minimum,qu2018udas, Wang2024, ince2020superpixel}\\ 
        & \cellcolor{black!10} Gap Filling   & \cellcolor{black!10} \citep{shen2015missing, xu2016cloud, li2019cloud,ding2024robust, dong2018inpainting, qin2021image,sirjacobs2011cloud, kandasamy2013comparison, wang2012three}\\
        
        \midrule
        
        Fusion & Component Substitution & \citep{Chavez1991, Luo2008, Yang2013, 4490068, YANG2012177}\\ 
        & \cellcolor{black!10} Alignment of a Shared Reduced Space & \cellcolor{black!10} \citep{6460533, XING2018165, tuia2014semisupervised,Liao2015, Debes2014, Ghamisi2017, Manzanera2016}\\
        & Synthesized Representations with Deep Learning & \citep{Huang2015, 9013047, HUANG2020115850, Shao2017, gupta2025mosaicmultimodalmultilabelsupervisionaware}\\

        \midrule
        
        Visualization  & \cellcolor{black!10} Spatial & \cellcolor{black!10} \citep{Canas1985, gross1993visualization, tasdemir2009exploiting, Najim2023, bachmann2005exploiting} \\ 
        & Temporal & \citep{Horel1981, Barnston1987, bueso2020nonlinear, runge2015, Ibebuchi}\\ 
        & \cellcolor{black!10} Abstract & \cellcolor{black!10} \citep{Song2019, zhang2020three, russwurm2020self}\\
        
        \midrule
        
        Anomaly Detection & Separation & \citep{TanLRR2019, NiuLRR2016, Du2014anom, Ma2018DFL, LRSM2014, Touati2018, Cheng2020}\\
        & \cellcolor{black!10} Linear Reconstruction & \cellcolor{black!10} \citep{Jablonski2015pca, Du2007, Nielson1998,Lu2004,Turgay2009, Mahecha2017Extremes, duan2021low, YousifChange2013, Festa2023}\\
        & Nonlinear Reconstruction & \citep{Gu2008anom, bengio2013representation, Shi2024change, xie2019spectral, lu2020manifold, wu2024taef, fan2021rgae, huo2024maae, xiang2021guidedAE, wang2022autoAD}\\
        
        \midrule
        
        Predictions & 
        \cellcolor{black!10} Reducing Spectral Redundancy & \cellcolor{black!10} \citep{harsanyi1994, penna2007transform, vaddi2020probabilistic, bruce2002, garcia2017statistical, qiao2016effective, li2018discriminant, li2011locality, luo2017local, bachmann2005exploiting}\\ 
        & Capturing Context & \citep{liu2016synthetic, Karami2012, Debes2014, Liao2015, Ghamisi2017, tuia2014semisupervised, rivera2017hyperspectral, brooks2012fitting, tseng2024lightweightpretrainedtransformersremote}\\ 
        & \cellcolor{black!10} Learning Representations & \cellcolor{black!10} \citep{luo2020, abdi2017spectral, Shao2017, Kang_2021, Manas_2021_ICCV, Ayush2021, klemmer2023satclip, Lu2025, czerkawski2024, Marsocci2025}\\
        \bottomrule
    \end{tabular}}
    \label{tab:master_summary_cites}
\end{table}

%% file: tables/task_method_matching_new.tex
\begin{table}[H]    
    \centering
    \caption{\textbf{We align FE methods with their corresponding RS tasks.} Each row represents an FE method, each column an RS task, and each cell lists representative papers where the method has been applied to the task. For clarity, papers using variants, combinations, or improvements of an FE method are listed under the base method they extend.}
    \resizebox{\textwidth}{!}{%
    \begin{tabular}{>{\raggedright\arraybackslash}p{.9cm} >{\raggedright\arraybackslash}p{4.6cm} >{\raggedright\arraybackslash}p{4.6cm} >{\raggedright\arraybackslash}p{4.6cm} >{\raggedright\arraybackslash}p{4.6cm} >{\raggedright\arraybackslash}p{4.6cm} >{\raggedright\arraybackslash}p{4.6cm}}
    \toprule
\multicolumn{7}{c}{\textbf{Signal and Structure Preserving}}\\
\midrule
 & \textbf{Compression}  & \textbf{Data Cleaning} & \textbf{Fusion} & \textbf{Visualization} & \textbf{Anomaly Detection} & \textbf{Prediction} \\
  \rowcolor{gray!10}\textbf{CuT} &   &   &  \citep{4490068} &   &   &  \citep{qiao2016effective}\\
\textbf{DFT} & \citep{benz1995comparison} &  &  &  &  & \citep{brooks2012fitting}\\
  \rowcolor{gray!10}\textbf{DWT} &  \citep{benz1995comparison, kaarna2000compression,skodras2001jpeg, penna2006progressive, serra2008remote, garcia2017statistical,dua2020comprehensive, xiang2024remote} &  \citep{chen2010denoising} &  \citep{Luo2008} &   &  \citep{Du2007} &  \citep{bruce2002, Karami2012, garcia2017statistical, penna2007transform}\\
\toprule
\multicolumn{7}{c}{\textbf{Linear Variance and Reconstruction Preserving}}\\
\midrule
& \textbf{Compression}  & \textbf{Data Cleaning} & \textbf{Fusion} & \textbf{Visualization} & \textbf{Anomaly Detection} & \textbf{Prediction} \\
 \rowcolor{gray!10}\textbf{CCA} &  &  & \citep{Manzanera2016} &  & \citep{Nielson1998} & \citep{rivera2017hyperspectral}\\
\textbf{DL} &   &  \citep{xu2016cloud, li2019cloud} &   &   &  \citep{TanLRR2019, NiuLRR2016, Du2014anom, Ma2018DFL, LRSM2014, Cheng2020} &   \\
 \rowcolor{gray!10}\textbf{LDA} & & & & & &\citep{li2011locality, li2018discriminant, luo2017local, luo2020}\\
 \textbf{MNF} &   &  \citep{luo2016minimum} &   &   &   &  \citep{rivera2017hyperspectral}\\
 \rowcolor{gray!10}\textbf{OSP} & & & & & &\citep{harsanyi1994} \\
 \textbf{PCA} &  \citep{kaarna2000compression, Du2007, penna2007transform, serra2008remote, dua2020comprehensive} &  \citep{chen2010denoising, sirjacobs2011cloud, luo2016minimum, duan2021low, kandasamy2013comparison} &  \citep{Luo2008, Yang2013, 4490068, YANG2012177, Liao2015}& \citep{Canas1985, Horel1981, Barnston1987, bueso2020nonlinear, runge2015, czerkawski2024} &   \citep{duan2021low, Jablonski2015pca, YousifChange2013, Festa2023, Mahecha2017Extremes, Flach2017anom, Du2007} &  \citep{rivera2017hyperspectral, Nalepa2019, vaddi2020probabilistic, penna2007transform} \\
 \rowcolor{gray!10}\textbf{PLS} &  & \citep{wang2012three} &  &  &  & \citep{rivera2017hyperspectral}\\
 \textbf{POT} &  \citep{benz1995comparison, garcia2017statistical} &   &   &   &   &  \citep{garcia2017statistical} \\
 \rowcolor{gray!10}\textbf{TD} & \citep{dua2020comprehensive, Karami2012} &  &  &  &  & \citep{Karami2012} \\
  \toprule
\multicolumn{7}{c}{\textbf{Nonlinear Variance and Reconstruction Preserving}}\\
\midrule
 & \textbf{Compression}  & \textbf{Data Cleaning} & \textbf{Fusion} & \textbf{Visualization} & \textbf{Anomaly Detection} & \textbf{Prediction} \\
  \rowcolor{gray!10}\textbf{AE} & &  \citep{qin2021image, qu2018udas} &  \citep{Huang2015, 9013047, HUANG2020115850, XING2018165,Ghamisi2017,Shao2017} &  \citep{gross1993visualization, Ibebuchi} &  \citep{xie2019spectral, lu2020manifold, wu2024taef, fan2021rgae, huo2024maae, xiang2021guidedAE, wang2022autoAD, Shi2024change} &  \citep{hao2023review, song2021learning, abdi2017spectral, russwurm2020self, Shao2017, Ghamisi2017}\\
\textbf{kPCA} &  &  &  &  & \citep{Gu2008anom} & \citep{rivera2017hyperspectral}\\
  \toprule
\multicolumn{7}{c}{\textbf{Distribution Preserving}}\\
\midrule
 & \textbf{Compression}  & \textbf{Data Cleaning} & \textbf{Fusion} & \textbf{Visualization} & \textbf{Anomaly Detection} & \textbf{Prediction} \\
  \rowcolor{gray!10}\textbf{ICA} &  \citep{kaarna2000compression, wang2006independent} &   &   &   &  \citep{Flach2017anom} &   \\
\textbf{VAE} &  & \citep{ding2024robust} &  &  &  & \citep{lalitha2022review, hao2023review} \\
\toprule
\multicolumn{7}{c}{\textbf{Geometry \& Topology Preserving}}\\
\midrule
 & \textbf{Compression}  & \textbf{Data Cleaning} & \textbf{Fusion} & \textbf{Visualization} & \textbf{Anomaly Detection} & \textbf{Prediction} \\
  \rowcolor{gray!10}\textbf{Isomap} &   &   &   &  \citep{bachmann2005exploiting} &   &  \citep{bachmann2005exploiting}\\
\textbf{LE} &  &  & \citep{Debes2014, Liao2015} &  &  &  \\
  \rowcolor{gray!10}\textbf{LLE} &   &   &  \citep{6460533} &  \citep{bachmann2005exploiting} &   &  \citep{bachmann2005exploiting} \\
\textbf{MA} &  &  & \citep{tuia2014semisupervised} &  &  & \citep{tuia2014semisupervised} \\
  \rowcolor{gray!10}\textbf{MDS} &   &   &   &   &  \citep{Touati2018}&   \\
\textbf{SOM} &   &  &  & \citep{tasdemir2009exploiting, Najim2023} &  &  \\
  \rowcolor{gray!10}\textbf{t-SNE} &   &   &   &  \citep{Song2019, zhang2020three, russwurm2020self} &   &   \\
    \bottomrule
    \end{tabular}}
    \label{tab:task_method_matching_new}
\end{table}

%% file: tables/metrics_applications.tex
\begin{table}[H]
    \centering
    \caption{\textbf{The metrics used by the articles surveyed for feature extraction in remote sensing.}}
    \resizebox{\textwidth}{!}{\begin{tabular}{>{\raggedright\arraybackslash}p{1.5cm} >{\raggedright\arraybackslash}p{5.5cm} >{\raggedright\arraybackslash}p{5.5cm} >{\raggedright\arraybackslash}p{5.5cm} >{\raggedright\arraybackslash}p{5.5cm} >{\raggedright\arraybackslash}p{5.5cm} >{\raggedright\arraybackslash}p{5.5cm}}
        \toprule
                & \textbf{Compression} & \textbf{Denoising} & \textbf{Fusion} & \textbf{Visualization} & \textbf{Anom. Dect.} & \textbf{Predictions}\\
        \midrule
        \rowcolor{gray!10} \textbf{CC}      & \citep{Karami2012, dua2020comprehensive} & \citep{sirjacobs2011cloud, wang2012three, shen2015missing, li2019cloud} & \citep{Chavez1991, Luo2008, YANG2012177, liu2012, Yang2013, Huang2015, Manzanera2016,Shao2017, HUANG2020115850} & \citep{horel1984complex, bueso2020nonlinear, Najim2023, Ibebuchi} & \citep{Du2007} & \citep{Manzanera2016, song2021learning} \\
        \textbf{MSE }    & \citep{kaarna2000compression, dua2020comprehensive} & \citep{sirjacobs2011cloud, wang2012three, kandasamy2013comparison, shen2014effective, xu2016cloud, li2019cloud, dong2018inpainting, duan2021low} & \citep{YANG2012177, Huang2015, Shao2017, XING2018165, HUANG2020115850} & \citep{Najim2023} & \citep{Cheng2020} & \citep{penna2007transform, brooks2012fitting, garcia2017statistical, Shao2017, rivera2017hyperspectral}\\
        \rowcolor{gray!10} \textbf{ACC }    &  & \citep{duan2021low} & \citep{liu2012, tuia2014semisupervised, Debes2014, Ghamisi2017, Liao2015} &  & \citep{Touati2018, Shi2024change} & \citep{Nalepa2019, vaddi2020probabilistic, bruce2002, qiao2016effective, li2011locality, luo2017local, li2018discriminant, luo2020, liu2016synthetic, klemmer2023satclip, song2021learning, penna2007transform, li2022dimensionality, Karami2012, Liao2015, Debes2014, tuia2014semisupervised} \\
        \textbf{VIS }    & \citep{benz1995comparison, xiang2024remote} & \citep{chen2010denoising, sirjacobs2011cloud, wang2012three, dong2018inpainting, qu2018udas, li2019cloud, duan2021low, ding2024robust} & \citep{Yang2013, 4490068, YANG2012177, Huang2015, HUANG2020115850, liu2012, XING2018165, tuia2014semisupervised, Liu2021} & \citep{Song2019, zhang2020three, Barnston1987, bueso2020nonlinear, Ibebuchi} & \citep{Gu2008anom, TanLRR2019, NiuLRR2016, Du2014anom, Ma2018DFL, Cheng2020, xie2019spectral, lu2020manifold, wu2024taef, fan2021rgae, huo2024maae, xiang2021guidedAE, wang2022autoAD, wang2022adaptive, YousifChange2013, Touati2018, Shi2024change, Mahecha2017Extremes, Festa2023} & \citep{qiao2016effective, li2011locality, luo2017local, bachmann2005exploiting, li2018discriminant, luo2020, klemmer2023satclip, song2021learning, harsanyi1994, li2022dimensionality}\\
        \rowcolor{gray!10} CT      & \citep{Karami2012, dua2020comprehensive, kaarna2000compression, xiang2021guidedAE} & \citep{qu2018udas, li2019cloud} & \citep{tuia2014semisupervised, HUANG2020115850} &  & \citep{Jablonski2015pca, fan2021rgae, huo2024maae, Shi2024change} & \citep{bachmann2005exploiting, rivera2017hyperspectral, li2018discriminant} \\
        $\boldsymbol{\kappa}$&  &  & \citep{tuia2014semisupervised, Debes2014, Liao2015, Ghamisi2017} &  & \citep{Shi2024change}& \citep{Nalepa2019, vaddi2020probabilistic, li2018discriminant, penna2007transform, li2022dimensionality, Liao2015, Debes2014, tuia2014semisupervised}\\
        \rowcolor{gray!10} \textbf{SAD/ SAM }    & \citep{Karami2012, dua2020comprehensive} & \citep{shen2015missing, xu2016cloud, qu2018udas} & \citep{4490068, YANG2012177, liu2012, Huang2015, XING2018165, LIU2019SAR, HUANG2020115850} &  &  &  \\
        \textbf{SNR}     & ~\citep{benz1995comparison, penna2006progressive, penna2007transform, Du2007, Karami2012, garcia2017statistical, dua2020comprehensive} & ~\citep{sirjacobs2011cloud, chen2010denoising} &  &  &  &  \\
        \rowcolor{gray!10} \textbf{PSNR}    & ~\citep{penna2006progressive, penna2007transform, dua2020comprehensive} & ~\citep{shen2015missing, xu2016cloud, li2019cloud, xiang2021guidedAE, ding2024robust} &  &  &  &  \\
        \textbf{BR}      & ~\citep{kaarna2000compression, penna2006progressive, penna2007transform, Du2007, Karami2012, garcia2017statistical, dua2020comprehensive} &  &  &  &  &  \\
        \rowcolor{gray!10} \textbf{RD}      & ~\citep{penna2006progressive, Du2007, garcia2017statistical, xiang2024remote} &  &  &  &  &  \\
        \textbf{ERGAS}   &  &  & \citep{4490068, YANG2012177, Huang2015, XING2018165, LIU2019SAR, HUANG2020115850, XING2018165}& & & \\
        \rowcolor{gray!10} \textbf{Q}       &  & & \citep{4490068, YANG2012177, liu2012, Huang2015, XING2018165, HUANG2020115850} &  &  &  \\
        \textbf{AUC} &  &  &  &  & \citep{Jablonski2015pca, Gu2008anom, TanLRR2019, NiuLRR2016, Du2014anom, Ma2018DFL, Cheng2020, xie2019spectral, lu2020manifold, wu2024taef, fan2021rgae, huo2024maae, xiang2021guidedAE, wang2022adaptive, wang2022autoAD, YousifChange2013, Shi2024change, Flach2017anom, Du2007} &  \\
        \rowcolor{gray!10} \textbf{R$^2$} &  &  &  &  &  & \citep{klemmer2023satclip, rivera2017hyperspectral, brooks2012fitting}\\
    \end{tabular}}
    
    \label{tab:metrics_cites}
\end{table}

%% file: references.bib
@article{Tuia21perspective,
	title        = {{Towards a Collective Agenda on {AI} for {E}arth Science Data Analysis}},
	author       = {Tuia, D. and Roscher, R. and Wegner, J.D. and Jacobs, N. and Zhu, X.X. and Camps-Valls, G.},
	year         = 2021,
	month        = {jun},
	journal      = {IEEE Geosci. Remote Sens. Mag.},
	volume       = 9,
	number       = 2,
	pages        = {88--104},
	doi          = {https://doi.org/10.1109/MGRS.2020.3043504}
}

@article{brown2025alphaearth,
	title        = {Alpha{E}arth {F}oundations: {A}n embedding field model for accurate and efficient global mapping from sparse label data},
	author       = {Brown, Christopher F and Kazmierski, Michal R and Pasquarella, Valerie J and Rucklidge, William J and Samsikova, Masha and Zhang, Chenhui and Shelhamer, Evan and Lahera, Estefania and Wiles, Olivia and Ilyushchenko, Simon and others},
	year         = 2025,
	journal      = {arXiv preprint arXiv:2507.22291}
}

@article{costa2025terracodec,
	title        = {Terra{C}odec: {C}ompressing {E}arth Observations},
	author       = {Costa-Watanabe, Julen and Wittmann, Isabelle and Blumenstiel, Benedikt and Schindler, Konrad},
	year         = 2025,
	journal      = {arXiv preprint arXiv:2510.12670}
}

@inproceedings{psnr,
	title        = {Image Quality Metrics: {PSNR} vs. {SSIM}},
	author       = {Horé, Alain and Ziou, Djemel},
	year         = 2010,
	booktitle    = {Int. Conf. Pattern Recognit.},
	volume       = {},
	number       = {},
	pages        = {2366--2369},
	doi          = {10.1109/ICPR.2010.579}
}

@book{CampsValls21wiley,
	title        = {Deep learning for the {E}arth Sciences: {A} comprehensive approach to remote sensing, climate science and geosciences},
	author       = {Camps-Valls, G. and Tuia, D. and Zhu, X.X. and Reichstein, M. (Editors)},
	year         = 2021,
	publisher    = {Wiley \& Sons},
	isbn         = 9781119646143,
	url          = {https://github.com/DL4ES}
}

@article{Review2015,
	title        = {Missing Information Reconstruction of Remote Sensing Data: {A} Technical Review},
	author       = {Shen, Huanfeng and Li, Xinghua and Cheng, Qing and Zeng, Chao and Yang, Gang and Li, Huifang and Zhang, Liangpei},
	year         = 2015,
	journal      = {IEEE Geosci. Remote Sens. Mag.},
	volume       = 3,
	number       = 3,
	pages        = {61--85},
	doi          = {10.1109/MGRS.2015.2441912}
}

@article{garcia2017statistical,
	title        = {Statistical atmospheric parameter retrieval largely benefits from spatial--spectral image compression},
	author       = {Garc{\'\i}a-Sobrino, Joaqu{\'\i}n and Serra-Sagrist{\`a}, Joan and Laparra, Valero and Calbet, Xavier and Camps-Valls, Gustau},
	year         = 2017,
	journal      = {IEEE Trans. Geosci. Remote Sens.},
	publisher    = {IEEE},
	volume       = 55,
	number       = 4,
	pages        = {2213--2224}
}

@article{reichstein2019deep,
	title        = {Deep learning and process understanding for data-driven {E}arth system science},
	author       = {Reichstein, Markus and Camps-Valls, Gustau and Stevens, Bjorn and Jung, Martin and Denzler, Joachim and Carvalhais, Nuno and Prabhat, F},
	year         = 2019,
	journal      = {Nature},
	publisher    = {Nature Publishing Group UK London},
	volume       = 566,
	number       = 7743,
	pages        = {195--204}
}

@article{9082155,
	title        = {Feature Extraction for Hyperspectral Imagery: The Evolution From Shallow to Deep: Overview and Toolbox},
	author       = {Rasti, Behnood and Hong, Danfeng and Hang, Renlong and Ghamisi, Pedram and Kang, Xudong and Chanussot, Jocelyn and Benediktsson, Jon Atli},
	year         = 2020,
	journal      = {IEEE Geosci. Remote Sens. Mag.},
	volume       = 8,
	number       = 4,
	pages        = {60--88},
	doi          = {10.1109/MGRS.2020.2979764}
}

@incollection{izquierdo2017advanced,
	title        = {Advanced feature extraction for {E}arth observation data processing},
	author       = {Izquierdo-Verdiguier, E and Laparra, V and Mar{\'\i}, J Mu{\~n}oz and Chova, L G{\'o}mez and Camps-Valls, G},
	year         = 2017,
	booktitle    = {Compreh. Remote Sens.},
	publisher    = {Elsevier},
	pages        = {108--133}
}

@article{micchelli2006universal,
	title        = {Universal Kernels.},
	author       = {Micchelli, Charles A and Xu, Yuesheng and Zhang, Haizhang},
	year         = 2006,
	journal      = {J. Mach. Learn. Res.},
	volume       = 7,
	number       = 12
}

@article{9451654,
	title        = {Low-Rank and Sparse Representation for Hyperspectral Image Processing: A review},
	author       = {Peng, Jiangtao and Sun, Weiwei and Li, Heng-Chao and Li, Wei and Meng, Xiangchao and Ge, Chiru and Du, Qian},
	year         = 2022,
	journal      = {IEEE Geosci. Remote Sens. Mag.},
	volume       = 10,
	number       = 1,
	pages        = {10--43},
	doi          = {10.1109/MGRS.2021.3075491}
}

@article{van2009dimensionality,
  title={Dimensionality reduction: A comparative review},
  author={Van Der Maaten, Laurens and Postma, Eric O and Van Den Herik, H Jaap and others},
  journal={Journal of Machine Learning Research},
  volume={10},
  number={1},
  pages={1--41},
  year={2009}
}

@book{lee2007nonlinear,
	title        = {Nonlinear dimensionality reduction},
	author       = {Lee, John A and Verleysen, Michel},
	year         = 2007,
	publisher    = {Springer Science \& Business Media}
}

@article{nanga2021review,
	title        = {Review of dimension reduction methods},
	author       = {Nanga, Salifu and Bawah, Ahmed Tijani and Acquaye, Benjamin Ansah and Billa, Mac-Issaka and Baeta, Francis Delali and Odai, Nii Afotey and Obeng, Samuel Kwaku and Nsiah, Ampem Darko},
	year         = 2021,
	journal      = {J. Data Anal. Inf. Process.},
	publisher    = {Scientific Research Publishing},
	volume       = 9,
	number       = 3,
	pages        = {189--231}
}

@article{abdi2017spectral,
	title        = {Spectral-spatial feature learning for hyperspectral imagery classification using deep stacked sparse autoencoder},
	author       = {Abdi, Ghasem and Samadzadegan, Farhad and Reinartz, Peter},
	year         = 2017,
	journal      = {J. Appl. Remote Sens.},
	publisher    = {Society of Photo-Optical Instrumentation Engineers},
	volume       = 11,
	number       = 4,
	pages        = {042604--042604}
}

@article{4490068,
	title        = {An Efficient Pan-Sharpening Method via a Combined Adaptive PCA Approach and Contourlets},
	author       = {Shah, Vijay P. and Younan, Nicolas H. and King, Roger L.},
	year         = 2008,
	journal      = {IEEE Trans. Geosci. Remote Sens.},
	volume       = 46,
	number       = 5,
	pages        = {1323--1335},
	doi          = {10.1109/TGRS.2008.916211}
}

@article{Luo2008,
	title        = {Fusion of remote sensing image base on the {PCA}+{ATROUS} wavelet transform},
	author       = {Luo, Yan and Liu, Rong and Zhu, Yu},
	year         = 2008,
	month        = {01},
	journal      = {Int. Soc. Photogramm. Remote Sens.},
	volume       = 37,
	pages        = {}
}

@article{wald1997fusion,
	title        = {Fusion of satellite images of different spatial resolutions: {A}ssessing the quality of resulting images},
	author       = {Wald, Lucien and Ranchin, Thierry and Mangolini, Marc},
	year         = 1997,
	journal      = {Photogramm. Eng. Remote Sens.},
	volume       = 63,
	number       = 6,
	pages        = {691--699}
}

@article{wang2002universal,
	title        = {A universal image quality index},
	author       = {Wang, Zhou and Bovik, Alan C},
	year         = 2002,
	journal      = {IEEE Signal Process. Lett.},
	publisher    = {IEEE},
	volume       = 9,
	number       = 3,
	pages        = {81--84}
}

@article{YANG2012177,
	title        = {Fusion of multispectral and panchromatic images based on support value transform and adaptive principal component analysis},
	author       = {Shuyuan Yang and Min Wang and Licheng Jiao},
	year         = 2012,
	journal      = {Inf. Fusion},
	volume       = 13,
	number       = 3,
	pages        = {177--184},
	doi          = {https://doi.org/10.1016/j.inffus.2010.09.003},
	issn         = {1566-2535},
	url          = {https://www.sciencedirect.com/science/article/pii/S1566253510000898}
}

@article{Chavez1991,
	title        = {Comparison of Three Different Methods to Merge Multiresolution and Multispectral Data: {L}andsat TM and SPOT Panchromatic},
	author       = {Chavez, Jr, Pat and Sides, Stuart and Anderson, Jeffrey},
	year         = 1991,
	month        = {03},
	journal      = {Photogramm. Eng. Remote Sens.},
	volume       = 57,
	pages        = {265--303}
}

@article{skodras2001jpeg,
	title        = {The {JPEG} 2000 still image compression standard},
	author       = {Skodras, Athanassios and Christopoulos, Charilaos and Ebrahimi, Touradj},
	year         = 2001,
	journal      = {IEEE Signal Process. Mag.},
	publisher    = {IEEE},
	volume       = 18,
	number       = 5,
	pages        = {36--58}
}

@article{hotelling1933analysis,
	title        = {Analysis of a complex of statistical variables into principal components.},
	author       = {Hotelling, Harold},
	year         = 1933,
	journal      = {J. Educ. Psychol.},
	publisher    = {Warwick \& York},
	volume       = 24,
	number       = 6,
	pages        = 417
}

@article{bank2023autoencoders,
	title        = {Autoencoders},
	author       = {Bank, Dor and Koenigstein, Noam and Giryes, Raja},
	year         = 2023,
	journal      = {Mach. Learn. Data Sci. Handb.: Data Min. Knowl. Discov. Handb.},
	publisher    = {Springer},
	pages        = {353--374}
}

@article{kingma2013auto,
	title        = {Auto-encoding variational bayes},
	author       = {Kingma, Diederik P and Welling, Max},
	year         = 2013,
	journal      = {arXiv preprint arXiv:1312.6114}
}

@article{van2008visualizing,
	title        = {Visualizing data using t-{SNE}.},
	author       = {Van der Maaten, Laurens and Hinton, Geoffrey},
	year         = 2008,
	journal      = {J. Mach. Learn. Res.},
	volume       = 9,
	number       = 11
}

@article{saul2000introduction,
	title        = {An introduction to locally linear embedding},
	author       = {Saul, Lawrence K and Roweis, Sam T},
	year         = 2000,
	journal      = {unpublished. Available at: http://www. cs. toronto. edu/\~{} roweis/lle/publications. html}
}

@book{CampsValls09,
	title        = {Kernel methods for remote sensing data analysis},
	author       = {},
	year         = 2009,
	month        = {Dec},
	publisher    = {Wiley \& Sons},
	address      = {UK},
	isbn         = {978-0-470-72211-4},
	editor       = {Camps-Valls, G. and Bruzzone, L.}
}

@article{runge2015,
	title        = {Identifying causal gateways and mediators in complex spatio-temporal systems},
	author       = {Runge, Jakob and Petoukhov, Vladimir and Donges, Jonathan F. and Hlinka, Jaroslav and Jajcay, Nikola and Vejmelka, Martin and Hartman, David and Marwan, Norbert and Paluš, Milan and Kurths, Jürgen and et al.},
	year         = 2015,
	month        = {Jul},
	journal      = {Nat. Commun.},
	volume       = 6,
	number       = 1
}

@article{russwurm2020self,
	title        = {Self-attention for raw optical satellite time series classification},
	author       = {Ru{\ss}wurm, Marc and K{\"o}rner, Marco},
	year         = 2020,
	journal      = {ISPRS J. Photogramm. Remote Sens.},
	publisher    = {Elsevier},
	volume       = 169,
	pages        = {421--435}
}

@article{altman2018curse,
	title        = {The curse(s) of dimensionality},
	author       = {Altman, Naomi and Krzywinski, Martin},
	year         = 2018,
	journal      = {Nature Methods},
	volume       = 15,
	number       = 6,
	pages        = {399--400}
}

@article{dua2020comprehensive,
	title        = {Comprehensive review of hyperspectral image compression algorithms},
	author       = {Dua, Yaman and Kumar, Vinod and Singh, Ravi Shankar},
	year         = 2020,
	journal      = {Optical Engineering},
	publisher    = {Society of Photo-Optical Instrumentation Engineers},
	volume       = 59,
	number       = 9,
	pages        = {090902--090902}
}

@article{dey2018big,
	title        = {Big data for remote sensing: Visualization, analysis and interpretation},
	author       = {Dey, Nilanjan and Bhatt, Chintan and Ashour, Amira S},
	year         = 2018,
	journal      = {Cham: Springer},
	publisher    = {Springer},
	volume       = 104
}

@article{wang2023tensor,
	title        = {Tensor decompositions for hyperspectral data processing in remote sensing: {A} comprehensive review},
	author       = {Wang, Minghua and Hong, Danfeng and Han, Zhu and Li, Jiaxin and Yao, Jing and Gao, Lianru and Zhang, Bing and Chanussot, Jocelyn},
	year         = 2023,
	journal      = {IEEE Geosci. Remote Sens. Mag.},
	publisher    = {IEEE},
	volume       = 11,
	number       = 1,
	pages        = {26--72}
}

@article{bandos2009classification,
	title        = {Classification of hyperspectral images with regularized linear discriminant analysis},
	author       = {Bandos, Tatyana V and Bruzzone, Lorenzo and Camps-Valls, Gustavo},
	year         = 2009,
	journal      = {IEEE Trans. Geosci. Remote Sens.},
	publisher    = {IEEE},
	volume       = 47,
	number       = 3,
	pages        = {862--873}
}

@article{hu2022hyperspectral,
	title        = {Hyperspectral anomaly detection using deep learning: {A} review},
	author       = {Hu, Xing and Xie, Chun and Fan, Zhe and Duan, Qianqian and Zhang, Dawei and Jiang, Linhua and Wei, Xian and Hong, Danfeng and Li, Guoqiang and Zeng, Xinhua and others},
	year         = 2022,
	journal      = {Remote Sens.},
	publisher    = {MDPI},
	volume       = 14,
	number       = 9,
	pages        = 1973
}

@article{maxwell2018implementation,
	title        = {Implementation of machine-learning classification in remote sensing: An applied review},
	author       = {Maxwell, Aaron E and Warner, Timothy A and Fang, Fang},
	year         = 2018,
	journal      = {Int. J. Remote Sens.},
	publisher    = {Taylor \& Francis},
	volume       = 39,
	number       = 9,
	pages        = {2784--2817}
}

@article{li2022dimensionality,
	title        = {Dimensionality reduction and classification of hyperspectral remote sensing image feature extraction},
	author       = {Li, Hongda and Cui, Jian and Zhang, Xinle and Han, Yongqi and Cao, Liying},
	year         = 2022,
	journal      = {Remote Sens.},
	publisher    = {MDPI},
	volume       = 14,
	number       = 18,
	pages        = 4579
}

@inproceedings{shurmer2018sentinels,
	title        = {{S}entinels optical communications payload (OCP) operations: {F}rom test to in-flight experience},
	author       = {Shurmer, Ian and Marchese, Franco and Morales-Santiago, Jose-M and Emanuelli, Pier Paolo},
	year         = 2018,
	booktitle    = {2018 SpaceOps Conference},
	pages        = 2654
}

@incollection{goodman2020goes,
	title        = {{GOES}-{R} series introduction},
	author       = {Goodman, Steven J},
	year         = 2020,
	booktitle    = {The GOES-R Series},
	publisher    = {Elsevier},
	pages        = {1--3}
}

@article{prudente2020limitations,
	title        = {Limitations of cloud cover for optical remote sensing of agricultural areas across {S}outh {A}merica},
	author       = {Prudente, Victor Hugo Rohden and Martins, Vitor Souza and Vieira, Denis Corte and e Silva, Nildson Rodrigues de Fran{\c{c}}a and Adami, Marcos and Sanches, Ieda Del’Arco},
	year         = 2020,
	journal      = {Remote Sensing Applications: Society and Environment},
	publisher    = {Elsevier},
	volume       = 20,
	pages        = 100414
}

@article{vanonckelen2013effect,
	title        = {The effect of atmospheric and topographic correction methods on land cover classification accuracy},
	author       = {Vanonckelen, Steven and Lhermitte, Stefaan and Van Rompaey, Anton},
	year         = 2013,
	journal      = {Int. J. Appl. Earth Obs. Geoinf.},
	publisher    = {Elsevier},
	volume       = 24,
	pages        = {9--21}
}

@article{kingra2016application,
	title        = {Application of remote sensing and {GIS} in agriculture and natural resource management under changing climatic conditions.},
	author       = {Kingra, Pavneet Kaur and Majumder, Debjyoti and Singh, Som Pal},
	year         = 2016,
	journal      = {Agricultural Research Journal},
	volume       = 53,
	number       = 3
}

@article{wellmann2020remote,
	title        = {Remote sensing in urban planning: {C}ontributions towards ecologically sound policies?},
	author       = {Wellmann, Thilo and Lausch, Angela and Andersson, Erik and Knapp, Sonja and Cortinovis, Chiara and Jache, Jessica and Scheuer, Sebastian and Kremer, Peleg and Mascarenhas, Andr{\'e} and Kraemer, Roland and others},
	year         = 2020,
	journal      = {Landscape and urban planning},
	publisher    = {Elsevier},
	volume       = 204,
	pages        = 103921
}

@article{weiss2020remote,
	title        = {Remote sensing for agricultural applications: {A} meta-review},
	author       = {Weiss, Marie and Jacob, Fr{\'e}d{\'e}ric and Duveiller, Grgory},
	year         = 2020,
	journal      = {Remote Sens. Environ.},
	publisher    = {Elsevier},
	volume       = 236,
	pages        = 111402
}

@article{van2000remote,
	title        = {Remote sensing for natural disaster management},
	author       = {Van Westen, CJ},
	year         = 2000,
	journal      = {International archives of photogrammetry and remote sensing},
	publisher    = {International Society For Photogrammetry \& Remote},
	volume       = 33,
	number       = {B7/4; PART 7},
	pages        = {1609--1617}
}

@article{li2020review,
	title        = {A review of remote sensing for environmental monitoring in {C}hina},
	author       = {Li, Jun and Pei, Yanqiu and Zhao, Shaohua and Xiao, Rulin and Sang, Xiao and Zhang, Chengye},
	year         = 2020,
	journal      = {Remote Sens.},
	publisher    = {MDPI},
	volume       = 12,
	number       = 7,
	pages        = 1130
}

@inproceedings{vaddi2020probabilistic,
	title        = {Probabilistic {PCA} based hyper spectral image classification for remote sensing applications},
	author       = {Vaddi, Radhesyam and Manoharan, Prabukumar},
	year         = 2020,
	booktitle    = {Intelligent Systems Design and Applications: 18th International Conference on Intelligent Systems Design and Applications (ISDA 2018) held in Vellore, India, December 6-8, 2018, Volume 2},
	pages        = {863--869},
	organization = {Springer}
}

@article{liu2012,
	title        = {Functional factor analysis for periodic remote sensing data},
	author       = {Liu, Chong and Ray, Surajit and Hooker, Giles and Friedl, Mark},
	year         = 2012,
	journal      = {The Annals of Applied Statistics},
	volume       = 6,
	number       = 2,
	pages        = {601–624}
}

@article{Hannachi2007,
	title        = {Empirical \Uppercase{O}rthogonal \Uppercase{F}unctions and related techniques in atmospheric science: A review},
	author       = {Hannachi, A. and Jolliffe, I. T. and Stephenson, D. B.},
	year         = 2007,
	journal      = {Int. J. Climatol.},
	volume       = 27,
	number       = 9,
	pages        = {1119--1152}
}

@article{luo2020,
	title        = {Dimensionality Reduction With Enhanced Hybrid-Graph Discriminant Learning for Hyperspectral Image Classification},
	author       = {Luo, Fulin and Zhang, Liangpei and Du, Bo and Zhang, Lefei},
	year         = 2020,
	journal      = {IEEE Trans. Geosci. Remote Sens.},
	volume       = 58,
	number       = 8,
	pages        = {5336--5353},
	doi          = {10.1109/TGRS.2020.2963848}
}

@article{chen2010denoising,
	title        = {Denoising of hyperspectral imagery using principal component analysis and wavelet shrinkage},
	author       = {Chen, Guangyi and Qian, Shen-En},
	year         = 2010,
	journal      = {IEEE Trans. Geosci. Remote Sens.},
	publisher    = {IEEE},
	volume       = 49,
	number       = 3,
	pages        = {973--980}
}

@book{candes1999curvelets,
	title        = {Curvelets: {A} surprisingly effective nonadaptive representation for objects with edges},
	author       = {Candes, Emmanuel Jean and Donoho, David Leigh and others},
	year         = 1999,
	publisher    = {Department of Statistics, Stanford University Stanford, CA, USA}
}

@article{vautard1989singular,
	title        = {Singular spectrum analysis in nonlinear dynamics, with applications to paleoclimatic time series},
	author       = {Vautard, Robert and Ghil, Michael},
	year         = 1989,
	journal      = {Physica D},
	publisher    = {Elsevier},
	volume       = 35,
	number       = 3,
	pages        = {395--424}
}

@article{qiao2016effective,
	title        = {Effective denoising and classification of hyperspectral images using curvelet transform and singular spectrum analysis},
	author       = {Qiao, Tong and Ren, Jinchang and Wang, Zheng and Zabalza, Jaime and Sun, Meijun and Zhao, Huimin and Li, Shutao and Benediktsson, J{\'o}n Atli and Dai, Qingyun and Marshall, Stephen},
	year         = 2016,
	journal      = {IEEE Trans. Geosci. Remote Sens.},
	publisher    = {IEEE},
	volume       = 55,
	number       = 1,
	pages        = {119--133}
}

@incollection{serra2008remote,
	title        = {Remote sensing data compression},
	author       = {Serra-Sagrist{\`a}, Joan and Aul{\'\i}-Llin{\`a}s, Francesc},
	year         = 2008,
	booktitle    = {Computational Intelligence for Remote Sensing},
	publisher    = {Springer},
	pages        = {27--61}
}

@article{Elsevier-1986GeladiK,
	title        = {Partial least-squares regression: {A} tutorial},
	author       = {Geladi, Paul and Kowalski, Bruce R},
	year         = 1986,
	journal      = {Analytica chimica acta},
	publisher    = {Elsevier},
	volume       = 185,
	pages        = {1--17}
}

@article{jutten1991blind,
	title        = {Blind separation of sources, Part I: An adaptive algorithm based on neuromimetic architecture},
	author       = {Jutten, Christian and Herault, Jeanny},
	year         = 1991,
	journal      = {Signal Process.},
	publisher    = {Elsevier},
	volume       = 24,
	number       = 1,
	pages        = {1--10}
}

@article{bruce2002,
	title        = {Dimensionality reduction of hyperspectral data using discrete wavelet transform feature extraction},
	author       = {Bruce, L.M. and Koger, C.H. and Jiang Li},
	year         = 2002,
	journal      = {IEEE Trans. Geosci. Remote Sens.},
	volume       = 40,
	number       = 10,
	pages        = {2331--2338},
	doi          = {10.1109/TGRS.2002.804721}
}

@article{mcinnes2018umap,
	title        = {{UMAP}: {U}niform manifold approximation and projection for dimension reduction},
	author       = {McInnes, Leland and Healy, John and Melville, James},
	year         = 2018,
	journal      = {arXiv preprint arXiv:1802.03426}
}

@inproceedings{varando2022learning,
	title        = {Learning causal representations with {G}ranger {PCA}},
	author       = {Varando, Gherardo and Fern{\'a}ndez-Torres, Miguel-{\'A}ngel and Mu{\~n}oz-Mar{\'\i}, Jordi and Camps-Valls, Gustau},
	year         = 2022,
	booktitle    = {UAI Workshop Causal Represent. Learn.}
}

@article{bueso2020nonlinear,
	title        = {Nonlinear {PCA} for spatio-temporal analysis of {E}arth observation data},
	author       = {Bueso, Diego and Piles, Maria and Camps-Valls, Gustau},
	year         = 2020,
	journal      = {IEEE Trans. Geosci. Remote Sens.},
	publisher    = {IEEE},
	volume       = 58,
	number       = 8,
	pages        = {5752--5763}
}

@article{li2022deep,
	title        = {Deep learning in multimodal remote sensing data fusion: A comprehensive review},
	author       = {Li, Jiaxin and Hong, Danfeng and Gao, Lianru and Yao, Jing and Zheng, Ke and Zhang, Bing and Chanussot, Jocelyn},
	year         = 2022,
	journal      = {Int. J. Appl. Earth Obs. Geoinf.},
	publisher    = {Elsevier},
	volume       = 112,
	pages        = 102926
}

@book{camps2009kernel,
	title        = {Kernel methods for remote sensing data analysis},
	author       = {Camps-Valls, Gustau and Bruzzone, Lorenzo},
	year         = 2009,
	publisher    = {John Wiley \& Sons}
}

@article{zhu2017deep,
	title        = {Deep learning in remote sensing: A comprehensive review and list of resources},
	author       = {Zhu, Xiao Xiang and Tuia, Devis and Mou, Lichao and Xia, Gui-Song and Zhang, Liangpei and Xu, Feng and Fraundorfer, Friedrich},
	year         = 2017,
	journal      = {IEEE Geosci. Remote Sens. Mag.},
	publisher    = {IEEE},
	volume       = 5,
	number       = 4,
	pages        = {8--36}
}

@article{lyzenga1978passive,
	title        = {Passive remote sensing techniques for mapping water depth and bottom features},
	author       = {Lyzenga, David R},
	year         = 1978,
	journal      = {Appl. Opt.},
	publisher    = {Optical Society of America},
	volume       = 17,
	number       = 3,
	pages        = {379--383}
}

@article{Turgay2009,
  author={Celik, Turgay},
  journal={IEEE Geoscience and Remote Sensing Letters}, 
  title={Unsupervised Change Detection in Satellite Images Using Principal Component Analysis and $k$-Means Clustering}, 
  year={2009},
  volume={6},
  number={4},
  pages={772-776},
  doi={10.1109/LGRS.2009.2025059}}

@article{Nielson1998,
title = {Multivariate Alteration Detection ({MAD}) and {MAF} Postprocessing in Multispectral, Bitemporal Image Data: New Approaches to Change Detection Studies},
journal = {Remote Sensing of Environment},
volume = {64},
number = {1},
pages = {1-19},
year = {1998},
issn = {0034-4257},
doi = {https://doi.org/10.1016/S0034-4257(97)00162-4},
url = {https://www.sciencedirect.com/science/article/pii/S0034425797001624},
author = {Allan A. Nielsen and Knut Conradsen and James J. Simpson}
}

@article{Coppin2004,
  title = {Review ArticleDigital change detection methods in ecosystem monitoring: a review},
  volume = {25},
  ISSN = {1366-5901},
  url = {http://dx.doi.org/10.1080/0143116031000101675},
  DOI = {10.1080/0143116031000101675},
  number = {9},
  journal = {International Journal of Remote Sensing},
  publisher = {Informa UK Limited},
  author = {Coppin,  P. and Jonckheere,  I. and Nackaerts,  K. and Muys,  B. and Lambin,  E.},
  year = {2004},
  month = may,
  pages = {1565–1596}
}

@article{Lu2004,
author = {Lu, Dengsheng and Mausel, Paul and Brondízio, Eduardo and Moran, Emilio},
year = {2004},
month = {01},
pages = {},
title = {Change Detection Techniques},
volume = {25},
journal = {International Journal of Remote Sensing}
}

@article{bengio2013representation,
	title        = {Representation Learning: {A} Review and New Perspectives},
	author       = {Bengio, Yoshua and Courville, Aaron and Vincent, Pascal},
	year         = 2013,
	journal      = {IEEE Trans. Pattern Anal. Mach. Intell.},
	volume       = 35,
	number       = 8,
	pages        = {1798--1828},
	doi          = {10.1109/TPAMI.2013.50}
}

@article{xie2019spectral,
	title        = {Spectral Constraint Adversarial Autoencoder for Hyperspectral Anomaly Detection},
	author       = {Xie, Yuan and Tao, Ran and Peng, Xing},
	year         = 2019,
	journal      = {Neural Networks},
	volume       = 118,
	pages        = {107--120},
	doi          = {10.1016/j.neunet.2019.07.001}
}

@article{Lu2020Manifold,
	title        = {Exploiting Embedding Manifold of Autoencoders for Hyperspectral Anomaly Detection},
	author       = {Lu, Xiaoqiang and Zhang, Wuxia and Huang, Ju},
	year         = 2020,
	journal      = {IEEE Trans. Geosci. Remote Sens.},
	volume       = 58,
	number       = 3,
	pages        = {1527--1537},
	doi          = {10.1109/TGRS.2019.2944419}
}

@article{wang2022adaptive,
	title        = {Adaptive Loss Function-Based Autoencoder for Hyperspectral Anomaly Detection},
	author       = {Wang, Tian and Wang, Hao and Sun, Hongqi},
	year         = 2022,
	journal      = {IEEE Trans. Image Process.},
	volume       = 31,
	pages        = {2234--2247},
	doi          = {10.1109/TIP.2022.3142547}
}

@article{Flach2017anom,
	title        = {Multivariate anomaly detection for {E}arth observations: {A} comparison of algorithms and feature extraction techniques},
	author       = {Flach, M. and Gans, F. and Brenning, A. and Denzler, J. and Reichstein, M. and Rodner, E. and Bathiany, S. and Bodesheim, P. and Guanche, Y. and Sippel, S. and Mahecha, M. D.},
	year         = 2017,
	journal      = {Earth Syst. Dyn.},
	volume       = 8,
	number       = 3,
	pages        = {677--696},
	doi          = {10.5194/esd-8-677-2017},
	url          = {https://esd.copernicus.org/articles/8/677/2017/}
}

@article{Ma2018DFL,
	title        = {Hyperspectral Anomaly Detection via Discriminative Feature Learning with Multiple-Dictionary Sparse Representation},
	author       = {Ma, Dandan and Yuan, Yuan and Wang, Qi},
	year         = 2018,
	journal      = {Remote Sens.},
	volume       = 10,
	number       = 5,
	doi          = {10.3390/rs10050745},
	issn         = {2072-4292},
	url          = {https://www.mdpi.com/2072-4292/10/5/745},
	article-number = 745
}

@article{Mahecha2017Extremes,
	title        = {Detecting impacts of extreme events with ecological in~situ monitoring networks},
	author       = {Mahecha, M. D. and Gans, F. and Sippel, S. and Donges, J. F. and Kaminski, T. and Metzger, S. and Migliavacca, M. and Papale, D. and Rammig, A. and Zscheischler, J.},
	year         = 2017,
	journal      = {Biogeosciences},
	volume       = 14,
	number       = 18,
	pages        = {4255--4277},
	doi          = {10.5194/bg-14-4255-2017},
	url          = {https://bg.copernicus.org/articles/14/4255/2017/}
}

@article{Jablonski2015pca,
	title        = {Principal Component Reconstruction Error for Hyperspectral Anomaly Detection},
	author       = {Jablonski, James A. and Bihl, Trevor J. and Bauer, Kenneth W.},
	year         = 2015,
	journal      = {IEEE Geosci. Remote Sens. Lett.},
	volume       = 12,
	number       = 8,
	pages        = {1725--1729},
	doi          = {10.1109/LGRS.2015.2421813}
}

@article{Du2014anom,
	title        = {A Discriminative Metric Learning Based Anomaly Detection Method},
	author       = {Du, Bo and Zhang, Liangpei},
	year         = 2014,
	journal      = {IEEE Trans. Geosci. Remote Sens.},
	volume       = 52,
	number       = 11,
	pages        = {6844--6857},
	doi          = {10.1109/TGRS.2014.2303895}
}

@article{Gu2008anom,
	title        = {A Selective {KPCA} Algorithm Based on High-Order Statistics for Anomaly Detection in Hyperspectral Imagery},
	author       = {Gu, Yanfeng and Liu, Ying and Zhang, Ye},
	year         = 2008,
	journal      = {IEEE Geosci. Remote Sens. Lett.},
	volume       = 5,
	number       = 1,
	pages        = {43--47},
	doi          = {10.1109/LGRS.2007.907304}
}

@article{LIU2019SAR,
	title        = {Stacked {F}isher autoencoder for {SAR} change detection},
	author       = {Ganchao Liu and Lingling Li and Licheng Jiao and Yongsheng Dong and Xuelong Li},
	year         = 2019,
	journal      = {Pattern Recognit.},
	volume       = 96,
	pages        = 106971,
	doi          = {https://doi.org/10.1016/j.patcog.2019.106971},
	issn         = {0031-3203},
	url          = {https://www.sciencedirect.com/science/article/pii/S0031320319302742}
}

@article{TanLRR2019,
	title        = {Anomaly Detection in Hyperspectral Imagery Based on Low-Rank Representation Incorporating a Spatial Constraint},
	author       = {Tan, Kun and Hou, Zengfu and Ma, Donglei and Chen, Yu and Du, Qian},
	year         = 2019,
	journal      = {Remote Sens.},
	volume       = 11,
	number       = 13,
	issn         = {2072-4292},
	url          = {https://www.mdpi.com/2072-4292/11/13/1578},
	article-number = 1578
}

@article{NiuLRR2016,
	title        = {Hyperspectral Anomaly Detection Based on Low-Rank Representation and Learned Dictionary},
	author       = {Niu, Yubin and Wang, Bin},
	year         = 2016,
	journal      = {Remote Sens.},
	volume       = 8,
	number       = 4,
	issn         = {2072-4292},
	url          = {https://www.mdpi.com/2072-4292/8/4/289},
	article-number = 289
}

@article{Shi2024change,
	title        = {Self-Guided Autoencoders for Unsupervised Change Detection in Heterogeneous Remote Sensing Images},
	author       = {Shi, Jiao and Wu, Tiancheng and Kai Qin, Alex and Lei, Yu and Jeon, Gwanggil},
	year         = 2024,
	journal      = {IEEE Trans. Artif. Intell.},
	volume       = 5,
	number       = 6,
	pages        = {2458--2471},
	doi          = {10.1109/TAI.2024.3357667}
}

@inproceedings{Touati2018,
	title        = {Change Detection in Heterogeneous Remote Sensing Images Based on an Imaging Modality-Invariant {MDS} Representation},
	author       = {Touati, Redha and Mignotte, Max and Dahmane, Mohamed},
	year         = 2018,
	booktitle    = {IEEE Int. Conf. Image Process},
	volume       = {},
	number       = {},
	pages        = {3998--4002},
	doi          = {10.1109/ICIP.2018.8451184}
}

@article{Cheng2020,
	title        = {Graph and Total Variation Regularized Low-Rank Representation for Hyperspectral Anomaly Detection},
	author       = {Cheng, Tongkai and Wang, Bin},
	year         = 2020,
	journal      = {IEEE Trans. Geosci. Remote Sens.},
	volume       = 58,
	number       = 1,
	pages        = {391--406},
	doi          = {10.1109/TGRS.2019.2936609}
}

@article{LRSM2014,
	title        = {Low-rank and sparse matrix decomposition-based anomaly detection for hyperspectral imagery},
	author       = {Weiwei Sun and Chun Liu and Jialin Li and Yenming Mark Lai and Weiyue Li},
	year         = 2014,
	journal      = {J. Appl. Remote Sens.},
	publisher    = {SPIE},
	volume       = 8,
	number       = 1,
	pages        = {083641},
	doi          = {10.1117/1.JRS.8.083641},
	url          = {https://doi.org/10.1117/1.JRS.8.083641}
}

@article{Carvalho2019,
	title        = {Machine Learning Interpretability: {A} Survey on Methods and Metrics},
	author       = {Carvalho, Diogo V. and Pereira, Eduardo M. and Cardoso, Jaime S.},
	year         = 2019,
	journal      = {Electronics},
	volume       = 8,
	number       = 8,
	doi          = {10.3390/electronics8080832},
	issn         = {2079-9292},
	url          = {https://www.mdpi.com/2079-9292/8/8/832},
	article-number = 832
}

@article{fan2021rgae,
	title        = {Hyperspectral Anomaly Detection With Robust Graph Autoencoders},
	author       = {Fan, Ganghui and Ma, Yong and Mei, Xiaoguang and Fan, Fan and Huang, Jun and Ma, Jiayi},
	year         = 2021,
	journal      = {IEEE Trans. Geosci. Remote Sens.},
	volume       = 60,
	pages        = {1--15},
	doi          = {10.1109/TGRS.2021.3097097}
}

@article{wu2024taef,
  title={Transformer-Based Autoencoder Framework for Nonlinear Hyperspectral Anomaly Detection},
  author={Wu, Ziyu and Wang, Bin},
  journal={IEEE Trans. Geosci. Remote Sens.},
  volume={62},
  pages={1--15},
  year={2024},
  publisher={IEEE}
}

@article{huo2024maae,
	title        = {Memory-Augmented Autoencoder With Adaptive Reconstruction and Sample Attribution Mining for Hyperspectral Anomaly Detection},
	author       = {Huo, Yu and Cheng, Xi and Lin, Sheng and Zhang, Min and Wang, Hai},
	year         = 2024,
	journal      = {IEEE Trans. Geosci. Remote Sens.},
	volume       = 62,
	pages        = {99313--99325},
	doi          = {10.1109/TGRS.2024.3399313}
}

@article{xiang2021guidedAE,
	title        = {Hyperspectral Anomaly Detection With Guided Autoencoder},
	author       = {Xiang, Pei and Ali, Shahzad and Jung, Soon Ki and Zhou, Huixin},
	year         = 2021,
	journal      = {IEEE Trans. Geosci. Remote Sens.},
	volume       = 60,
	pages        = {1--18},
	doi          = {10.1109/TGRS.2021.3057721}
}

@article{wang2022autoAD,
	title        = {{A}uto-{AD}: {A}utonomous Hyperspectral Anomaly Detection Network Based on Fully Convolutional Autoencoder},
	author       = {Wang, Shaoyu and Wang, Xinyu and Zhang, Liangpei and Zhong, Yanfei},
	year         = 2021,
	journal      = {IEEE Trans. Geosci. Remote Sens.},
	volume       = 60,
	pages        = {1--14},
	doi          = {10.1109/TGRS.2022.3207165}
}

@article{YousifChange2013,
	title        = {Improving Urban Change Detection From Multitemporal {SAR} Images Using {PCA}-{NLM}},
	author       = {Yousif, Osama and Ban, Yifang},
	year         = 2013,
	journal      = {IEEE Trans. Geosci. Remote Sens.},
	volume       = 51,
	number       = 4,
	pages        = {2032--2041},
	doi          = {10.1109/TGRS.2013.2245900}
}

@article{Festa2023,
	title        = {Unsupervised detection of {I}n{SAR} time series patterns based on {PCA} and K-means clustering},
	author       = {Davide Festa and Alessandro Novellino and Ekbal Hussain and Luke Bateson and Nicola Casagli and Pierluigi Confuorto and Matteo {Del Soldato} and Federico Raspini},
	year         = 2023,
	journal      = {Int. J. Appl. Earth Obs. Geoinf.},
	volume       = 118,
	pages        = 103276,
	doi          = {https://doi.org/10.1016/j.jag.2023.103276},
	issn         = {1569-8432},
	url          = {https://www.sciencedirect.com/science/article/pii/S1569843223000985}
}

@article{qin2021image,
	title        = {Image inpainting based on deep learning: {A} review},
	author       = {Qin, Zhen and Zeng, Qingliang and Zong, Yixin and Xu, Fan},
	year         = 2021,
	journal      = {Displays},
	publisher    = {Elsevier},
	volume       = 69,
	pages        = 102028
}

@article{shen2014effective,
	title        = {An effective thin cloud removal procedure for visible remote sensing images},
	author       = {Shen, Huanfeng and Li, Huifang and Qian, Yan and Zhang, Liangpei and Yuan, Qiangqiang},
	year         = 2014,
	journal      = {ISPRS J. Photogramm. Remote Sens.},
	publisher    = {Elsevier},
	volume       = 96,
	pages        = {224--235}
}

@article{li2019cloud,
	title        = {Cloud removal in remote sensing images using nonnegative matrix factorization and error correction},
	author       = {Li, Xinghua and Wang, Liyuan and Cheng, Qing and Wu, Penghai and Gan, Wenxia and Fang, Lina},
	year         = 2019,
	journal      = {ISPRS J. Photogramm. Remote Sens.},
	publisher    = {Elsevier},
	volume       = 148,
	pages        = {103--113}
}

@article{shen2015missing,
	title        = {Missing information reconstruction of remote sensing data: {A} technical review},
	author       = {Shen, Huanfeng and Li, Xinghua and Cheng, Qing and Zeng, Chao and Yang, Gang and Li, Huifang and Zhang, Liangpei},
	year         = 2015,
	journal      = {IEEE Geosci. Remote Sens. Mag.},
	publisher    = {IEEE},
	volume       = 3,
	number       = 3,
	pages        = {61--85}
}

@article{xu2016cloud,
	title        = {Cloud removal based on sparse representation via multitemporal dictionary learning},
	author       = {Xu, Meng and Jia, Xiuping and Pickering, Mark and Plaza, Antonio J},
	year         = 2016,
	journal      = {IEEE Trans. Geosci. Remote Sens.},
	publisher    = {IEEE},
	volume       = 54,
	number       = 5,
	pages        = {2998--3006}
}

@article{kandasamy2013comparison,
	title        = {A comparison of methods for smoothing and gap filling time series of remote sensing observations--application to {MODIS} {LAI} products},
	author       = {Kandasamy, Sivasathivel and Baret, Frederic and Verger, Aleixandre and Neveux, Philippe and Weiss, Marie},
	year         = 2013,
	journal      = {Biogeosciences},
	publisher    = {Copernicus GmbH},
	volume       = 10,
	number       = 6,
	pages        = {4055--4071}
}

@article{sirjacobs2011cloud,
	title        = {Cloud filling of ocean colour and sea surface temperature remote sensing products over the {S}outhern {N}orth {S}ea by the {D}ata {I}nterpolating {E}mpirical {O}rthogonal {F}unctions methodology},
	author       = {Sirjacobs, Damien and Alvera-Azc{\'a}rate, Aida and Barth, Alexander and Lacroix, Genevi{\`e}ve and Park, YoungJe and Nechad, Bouchra and Ruddick, Kevin and Beckers, Jean-Marie},
	year         = 2011,
	journal      = {Journal of Sea Research},
	publisher    = {Elsevier},
	volume       = 65,
	number       = 1,
	pages        = {114--130}
}

@article{wang2012three,
	title        = {A three-dimensional gap filling method for large geophysical datasets: {A}pplication to global satellite soil moisture observations},
	author       = {Wang, Guojie and Garcia, Damien and Liu, Yi and De Jeu, Richard and Dolman, A Johannes},
	year         = 2012,
	journal      = {Environmental Modelling \& Software},
	publisher    = {Elsevier},
	volume       = 30,
	pages        = {139--142}
}

@article{ding2024robust,
	title        = {Robust Haze and Thin Cloud Removal via Conditional Variational Autoencoders},
	author       = {Ding, Haidong and Xie, Fengying and Qiu, Linwei and Zhang, Xiaozhe and Shi, Zhenwei},
	year         = 2024,
	journal      = {IEEE Trans. Geosci. Remote Sens.},
	publisher    = {IEEE}
}

@article{brooks2012fitting,
	title        = {Fitting the multitemporal curve: {A} {F}ourier series approach to the missing data problem in remote sensing analysis},
	author       = {Brooks, Evan B and Thomas, Valerie A and Wynne, Randolph H and Coulston, John W},
	year         = 2012,
	journal      = {IEEE Trans. Geosci. Remote Sens.},
	publisher    = {IEEE},
	volume       = 50,
	number       = 9,
	pages        = {3340--3353}
}

@article{Kang_2021,
	title        = {Comparative study of different dimensionality reduction methods in hyperspectral image classification},
	author       = {Lei Kang and Xiaoqing Hu and Chengcheng Zhong and Kai Zhang and Yanan Jiang},
	year         = 2021,
	month        = {sep},
	journal      = {Journal of Physics: Conference Series},
	publisher    = {IOP Publishing},
	volume       = 2024,
	number       = 1,
	pages        = {012009},
	doi          = {10.1088/1742-6596/2024/1/012009},
	url          = {https://dx.doi.org/10.1088/1742-6596/2024/1/012009}
}

@article{dong2018inpainting,
	title        = {Inpainting of remote sensing {SST} images with deep convolutional generative adversarial network},
	author       = {Dong, Junyu and Yin, Ruiying and Sun, Xin and Li, Qiong and Yang, Yuting and Qin, Xukun},
	year         = 2018,
	journal      = {IEEE Geosci. Remote Sens. Lett.},
	publisher    = {IEEE},
	volume       = 16,
	number       = 2,
	pages        = {173--177}
}

@article{candes2011robust,
	title        = {Robust principal component analysis?},
	author       = {Cand{\`e}s, Emmanuel J and Li, Xiaodong and Ma, Yi and Wright, John},
	year         = 2011,
	journal      = {Journal of the ACM (JACM)},
	publisher    = {ACM New York, NY, USA},
	volume       = 58,
	number       = 3,
	pages        = {1--37}
}

@article{tsakiris2018dual,
	title        = {Dual principal component pursuit},
	author       = {Tsakiris, Manolis C and Vidal, Ren{\'e}},
	year         = 2018,
	journal      = {J. Mach. Learn. Res.},
	volume       = 19,
	number       = 18,
	pages        = {1--50}
}

@article{horel1984complex,
	title        = {Complex principal component analysis: {T}heory and examples},
	author       = {Horel, John D},
	year         = 1984,
	journal      = {J. Appl. Meteorol. Climatol.},
	volume       = 23,
	number       = 12,
	pages        = {1660--1673}
}

@article{kreutz2003dictionary,
	title        = {Dictionary learning algorithms for sparse representation},
	author       = {Kreutz-Delgado, Kenneth and Murray, Joseph F and Rao, Bhaskar D and Engan, Kjersti and Lee, Te-Won and Sejnowski, Terrence J},
	year         = 2003,
	journal      = {Neural Comput.},
	publisher    = {MIT Press One Rogers Street, Cambridge, MA 02142-1209, USA journals-info~…},
	volume       = 15,
	number       = 2,
	pages        = {349--396}
}

@article{ahmed1974discrete,
	title        = {Discrete cosine transform},
	author       = {Ahmed, Nasir and Natarajan, T\_ and Rao, Kamisetty R},
	year         = 1974,
	journal      = {IEEE Trans. Comp.},
	publisher    = {IEEE},
	volume       = 100,
	number       = 1,
	pages        = {90--93}
}

@article{baudat2000generalized,
	title        = {Generalized discriminant analysis using a kernel approach},
	author       = {Baudat, Gaston and Anouar, Fatiha},
	year         = 2000,
	journal      = {Neural Comput.},
	publisher    = {MIT Press One Rogers Street, Cambridge, MA 02142-1209, USA journals-info~…},
	volume       = 12,
	number       = 10,
	pages        = {2385--2404}
}

@inproceedings{scholkopf1997kernel,
	title        = {Kernel principal component analysis},
	author       = {Sch{\"o}lkopf, Bernhard and Smola, Alexander and M{\"u}ller, Klaus-Robert},
	year         = 1997,
	booktitle    = {Int. Conf. Artif. Neural Netw.},
	pages        = {583--588},
	organization = {Springer}
}

@article{Chandola2009anomaly,
	title        = {Anomaly detection: {A} survey},
	author       = {Chandola, Varun and Banerjee, Arindam and Kumar, Vipin},
	year         = 2009,
	month        = {jul},
	journal      = {ACM Comput. Surv.},
	publisher    = {Association for Computing Machinery},
	address      = {New York, NY, USA},
	volume       = 41,
	number       = 3,
	doi          = {10.1145/1541880.1541882},
	issn         = {0360-0300},
	url          = {https://doi.org/10.1145/1541880.1541882},
	issue_date   = {July 2009},
	articleno    = 15,
	numpages     = 58
}

@article{kohonen1990self,
	title        = {The self-organizing map},
	author       = {Kohonen, Teuvo},
	year         = 1990,
	journal      = {Proc. IEEE},
	publisher    = {IEEE},
	volume       = 78,
	number       = 9,
	pages        = {1464--1480}
}

@article{balasubramanian2002isomap,
	title        = {The isomap algorithm and topological stability},
	author       = {Balasubramanian, Mukund and Schwartz, Eric L},
	year         = 2002,
	journal      = {Science},
	publisher    = {American Association for the Advancement of Science},
	volume       = 295,
	number       = 5552,
	pages        = {7--7}
}

@article{keogh2001dimensionality,
	title        = {Dimensionality reduction for fast similarity search in large time series databases},
	author       = {Keogh, Eamonn and Chakrabarti, Kaushik and Pazzani, Michael and Mehrotra, Sharad},
	year         = 2001,
	journal      = {Knowl. Inf. Syst.},
	publisher    = {Springer},
	volume       = 3,
	pages        = {263--286}
}

@article{duhamel1990fast,
	title        = {Fast {F}ourier transforms: {A} tutorial review and a state of the art},
	author       = {Duhamel, Pierre and Vetterli, Martin},
	year         = 1990,
	journal      = {Signal Process.},
	publisher    = {Elsevier},
	volume       = 19,
	number       = 4,
	pages        = {259--299}
}

@book{broughton2018discrete,
	title        = {Discrete {F}ourier analysis and wavelets: applications to signal and image processing},
	author       = {Broughton, S Allen and Bryan, Kurt},
	year         = 2018,
	publisher    = {John Wiley \& Sons}
}

@article{rumelhart1986learning,
	title        = {Learning representations by back-propagating errors},
	author       = {Rumelhart, David E and Hinton, Geoffrey E and Williams, Ronald J},
	year         = 1986,
	journal      = {Nature},
	publisher    = {Nature Publishing Group UK London},
	volume       = 323,
	number       = 6088,
	pages        = {533--536}
}

@article{chalupka2017causal,
	title        = {Causal feature learning: {A}n overview},
	author       = {Chalupka, Krzysztof and Eberhardt, Frederick and Perona, Pietro},
	year         = 2017,
	journal      = {Behaviormetrika},
	publisher    = {Springer},
	volume       = 44,
	pages        = {137--164}
}

@article{szwagier2024curseisotropyprincipalcomponents,
  title   = {The Curse of Isotropy: From Principal Components to Principal Subspaces},
  author  = {Tom Szwagier and Xavier Pennec},
  journal = {Statistical Science},
  year    = {2023},
  note    = {In press},
  doi     = {10.48550/arXiv.2307.15348}
}

@article{saeed2018survey,
	title        = {A survey on multidimensional scaling},
	author       = {Saeed, Nasir and Nam, Haewoon and Haq, Mian Imtiaz Ul and Muhammad Saqib, Dost Bhatti},
	year         = 2018,
	journal      = {ACM Computing Surveys (CSUR)},
	publisher    = {ACM New York, NY, USA},
	volume       = 51,
	number       = 3,
	pages        = {1--25}
}

@inproceedings{ham2005semisupervised,
	title        = {Semisupervised alignment of manifolds},
	author       = {Ham, Jihun and Lee, Daniel and Saul, Lawrence},
	year         = 2005,
	booktitle    = {International Workshop on Artificial Intelligence and Statistics},
	pages        = {120--127},
	organization = {PMLR}
}

@article{tuia2014semisupervised,
	title        = {Semisupervised manifold alignment of multimodal remote sensing images},
	author       = {Tuia, Devis and Volpi, Michele and Trolliet, Maxime and Camps-Valls, Gustau},
	year         = 2014,
	journal      = {IEEE Trans. Geosci. Remote Sens.},
	publisher    = {IEEE},
	volume       = 52,
	number       = 12,
	pages        = {7708--7720}
}

@misc{hyperlabelme,
	title        = {HyperLabelMe: a Web Platform for Benchmarking Remote Sensing Image Classifiers},
	author       = {Jordi Mu\~noz-Mar\'i and Emma Izquierdo-Verdiguier and Manuel Campos-Taberner and Adri\'an P\'erez-Suay and Luis G\'omez-Chova and Gonzalo Mateo-Garc\'ia and Ana B. Ruescas and Valero Laparra and Jos\'e A. Padr\'on and Julia Amor\'os and Gustau Camps-Valls},
	year         = 2017,
	volume       = 5,
	number       = 4,
	pages        = {79--85},
	doi          = {https://doi.org/10.1109/MGRS.2017.2762476},
	url          = {http://hyperlabelme.uv.es/},
	note         = {V1.0}
}

@article{Du2007,
	title        = {Hyperspectral Image Compression Using {JPEG}2000 and Principal Component Analysis},
	author       = {Du, Qian and Fowler, James E.},
	year         = 2007,
	journal      = {IEEE Geosci. Remote Sens. Lett.},
	volume       = 4,
	number       = 2,
	pages        = {201--205},
	doi          = {10.1109/LGRS.2006.888109}
}

@article{damrich2022t,
	title        = {From t-{SNE} to {UMAP} with contrastive learning},
	author       = {Damrich, Sebastian and B{\"o}hm, Jan Niklas and Hamprecht, Fred A and Kobak, Dmitry},
	year         = 2022,
	journal      = {arXiv preprint arXiv:2206.01816}
}

@article{kobak2021initialization,
	title        = {Initialization is critical for preserving global data structure in both t-SNE and UMAP},
	author       = {Kobak, Dmitry and Linderman, George C},
	year         = 2021,
	journal      = {Nature biotechnology},
	publisher    = {Nature Publishing Group US New York},
	volume       = 39,
	number       = 2,
	pages        = {156--157}
}

@article{duderstadt2025nomad,
	title        = {{NOMAD} Projection},
	author       = {Duderstadt, Brandon and Nussbaum, Zach and Van der Maaten, Laurens},
	year         = 2025,
	journal      = {arXiv preprint arXiv:2505.15511}
}

@article{ascenso2023jpeg,
	title        = {The JPEG AI standard: Providing efficient human and machine visual data consumption},
	author       = {Ascenso, Jo{\~a}o and Alshina, Elena and Ebrahimi, Touradj},
	year         = 2023,
	journal      = {Ieee Multimedia},
	publisher    = {IEEE},
	volume       = 30,
	number       = 1,
	pages        = {100--111}
}

@article{pellicer2025video,
	title        = {Video Compression for Spatiotemporal {E}arth System Data},
	author       = {Pellicer-Valero, Oscar J and Aybar, Cesar and Valls, Gustau Camps},
	year         = 2025,
	journal      = {arXiv preprint arXiv:2506.19656}
}

@article{Karami2012,
	title        = {Compression of Hyperspectral Images Using Discerete Wavelet Transform and Tucker Decomposition},
	author       = {Karami, Azam and Yazdi, Mehran and Mercier, Grégoire},
	year         = 2012,
	journal      = {IEEE J. Sel. Top. Appl. Earth Obs. Remote Sens.},
	volume       = 5,
	number       = 2,
	pages        = {444--450},
	doi          = {10.1109/JSTARS.2012.2189200}
}

@article{Yang2013,
	title        = {A Novel Algorithm for Satellite Images Fusion Based on Compressed Sensing and {PCA}},
	author       = {Yang, Wenkao and Wang, Jing and Guo, Jing},
	year         = 2013,
	journal      = {Mathematical Problems in Engineering},
	volume       = 2013,
	number       = 1,
	pages        = 708985,
	doi          = {https://doi.org/10.1155/2013/708985}
}

@article{Ghamisi2019,
	title        = {Multisource and Multitemporal Data Fusion in Remote Sensing: {A} Comprehensive Review of the State of the Art},
	author       = {Ghamisi, Pedram and Rasti, Behnood and Yokoya, Naoto and Wang, Qunming and Hofle, Bernhard and Bruzzone, Lorenzo and Bovolo, Francesca and Chi, Mingmin and Anders, Katharina and Gloaguen, Richard and Atkinson, Peter M. and Benediktsson, Jon Atli},
	year         = 2019,
	journal      = {IEEE Geosci. Remote Sens. Mag.},
	volume       = 7,
	number       = 1,
	pages        = {6--39},
	doi          = {10.1109/MGRS.2018.2890023}
}

@inproceedings{6460533,
	title        = {Locally linear embedding based example learning for pan-sharpening},
	author       = {Liu, Qingjie and Liu, Lining and Wang, Yunhong and Zhang, Zhaoxiang},
	year         = 2012,
	booktitle    = {Proceedings of the 21st Int. Conf. Pattern Recognit. (ICPR2012)},
	volume       = {},
	number       = {},
	pages        = {1928--1931},
	doi          = {}
}

@article{XING2018165,
	title        = {Pan-sharpening via deep metric learning},
	author       = {Yinghui Xing and Min Wang and Shuyuan Yang and Licheng Jiao},
	year         = 2018,
	journal      = {ISPRS J. Photogramm. Remote Sens.},
	volume       = 145,
	pages        = {165--183},
	doi          = {https://doi.org/10.1016/j.isprsjprs.2018.01.016},
	issn         = {0924-2716},
	url          = {https://www.sciencedirect.com/science/article/pii/S0924271618300212},
	note         = {Deep Learning RS Data}
}

@article{lerman2018overview,
	title        = {An overview of robust subspace recovery},
	author       = {Lerman, Gilad and Maunu, Tyler},
	year         = 2018,
	journal      = {Proc. IEEE},
	publisher    = {IEEE},
	volume       = 106,
	number       = 8,
	pages        = {1380--1410}
}

@article{HUANG2020115850,
	title        = {Pan-sharpening via multi-scale and multiple deep neural networks},
	author       = {Wei Huang and Xuan Fei and Jingjing Feng and Hua Wang and Yan Liu and Yao Huang},
	year         = 2020,
	journal      = {Signal Processing: Image Communication},
	volume       = 85,
	pages        = 115850,
	doi          = {https://doi.org/10.1016/j.image.2020.115850},
	issn         = {0923-5965},
	url          = {https://www.sciencedirect.com/science/article/pii/S0923596520300710}
}

@article{9013047,
	title        = {Band-Independent Encoder–Decoder Network for Pan-Sharpening of Remote Sensing Images},
	author       = {Liu, Chi and Zhang, Yongjun and Wang, Shugen and Sun, Mingwei and Ou, Yangjun and Wan, Yi and Liu, Xiu},
	year         = 2020,
	journal      = {IEEE Trans. Geosci. Remote Sens.},
	volume       = 58,
	number       = 7,
	pages        = {5208--5223},
	doi          = {10.1109/TGRS.2020.2975230}
}

@article{Debes2014,
	title        = {Hyperspectral and {L}i{DAR} Data Fusion: {O}utcome of the 2013 GRSS Data Fusion Contest},
	author       = {Christian Debes and Andreas Merentitis and Roel Heremans and J{\"u}rgen T. Hahn and Nikolaos Frangiadakis and Tim van Kasteren and Wenzi Liao and Rik Bellens and Aleksandra Pivzurica and Sidharta Gautama and Wilfried Philips and Saurabh Prasad and Qian Du and Fabio Pacifici},
	year         = 2014,
	journal      = {IEEE J. Sel. Top. Appl. Earth Obs. Remote Sens.},
	volume       = 7,
	pages        = {2405--2418},
	url          = {https://api.semanticscholar.org/CorpusID:18245569}
}

@article{Huang2015,
	title        = {A New Pan-Sharpening Method With Deep Neural Networks},
	author       = {Huang, Wei and Xiao, Liang and Wei, Zhihui and Liu, Hongyi and Tang, Songze},
	year         = 2015,
	journal      = {IEEE Geosci. Remote Sens. Lett.},
	volume       = 12,
	number       = 5,
	pages        = {1037--1041},
	doi          = {10.1109/LGRS.2014.2376034}
}

@article{Shao2017,
	title        = {Stacked Sparse Autoencoder Modeling Using the Synergy of Airborne {LiDAR} and Satellite Optical and {SAR} Data to Map Forest Above-Ground Biomass},
	author       = {Shao, Zhenfeng and Zhang, Linjing and Wang, Lei},
	year         = 2017,
	journal      = {IEEE J. Sel. Top. Appl. Earth Obs. Remote Sens.},
	volume       = 10,
	number       = 12,
	pages        = {5569--5582},
	doi          = {10.1109/JSTARS.2017.2748341}
}

@article{Liao2015,
	title        = {Processing of Multiresolution Thermal Hyperspectral and Digital Color Data: {O}utcome of the 2014 {IEEE} {GRSS} Data Fusion Contest},
	author       = {Liao, Wenzhi and Huang, Xin and Van Coillie, Frieke and Gautama, Sidharta and Pižurica, Aleksandra and Philips, Wilfried and Liu, Hui and Zhu, Tingting and Shimoni, Michal and Moser, Gabriele and Tuia, Devis},
	year         = 2015,
	journal      = {IEEE J. Sel. Top. Appl. Earth Obs. Remote Sens.},
	volume       = 8,
	number       = 6,
	pages        = {2984--2996},
	doi          = {10.1109/JSTARS.2015.2420582}
}

@article{Ghamisi2017,
	title        = {Hyperspectral and {LiDAR} Data Fusion Using Extinction Profiles and Deep Convolutional Neural Network},
	author       = {Ghamisi, Pedram and Höfle, Bernhard and Zhu, Xiao Xiang},
	year         = 2017,
	journal      = {IEEE J. Sel. Top. Appl. Earth Obs. Remote Sens.},
	volume       = 10,
	number       = 6,
	pages        = {3011--3024},
	doi          = {10.1109/JSTARS.2016.2634863}
}

@article{Manzanera2016,
	title        = {Fusion of airborne {LiDAR} and multispectral sensors reveals synergic capabilities in forest structure characterization},
	author       = {Jose A. Manzanera and Antonio García-Abril and Cristina Pascual, Rosario Tejera and Susana Martín-Fernández and Timo Tokola and Ruben Valbuena},
	year         = 2016,
	journal      = {GIScience \& Remote Sensing},
	publisher    = {Taylor \& Francis},
	volume       = 53,
	number       = 6,
	pages        = {723--738},
	doi          = {10.1080/15481603.2016.1231605},
	url          = {https://doi.org/10.1080/15481603.2016.1231605},
	eprint       = {https://doi.org/10.1080/15481603.2016.1231605}
}

@article{rivera2017hyperspectral,
	title        = {Hyperspectral dimensionality reduction for biophysical variable statistical retrieval},
	author       = {Rivera-Caicedo, Juan Pablo and Verrelst, Jochem and Mu{\~n}oz-Mar{\'\i}, Jordi and Camps-Valls, Gustau and Moreno, Jos{\'e}},
	year         = 2017,
	journal      = {ISPRS J. Photogramm. Remote Sens.},
	publisher    = {Elsevier},
	volume       = 132,
	pages        = {88--101}
}

@article{li2018discriminant,
	title        = {Discriminant analysis-based dimension reduction for hyperspectral image classification: {A} survey of the most recent advances and an experimental comparison of different techniques},
	author       = {Li, Wei and Feng, Fubiao and Li, Hengchao and Du, Qian},
	year         = 2018,
	journal      = {IEEE Geosci. Remote Sens. Mag.},
	publisher    = {IEEE},
	volume       = 6,
	number       = 1,
	pages        = {15--34}
}

@article{liu2016synthetic,
	title        = {Synthetic aperture radar target configuration recognition using locality-preserving property and the Gamma distribution},
	author       = {Liu, Ming and Wu, Yan and Zhang, Qiang and Wang, Fan and Li, Ming},
	year         = 2016,
	journal      = {IET Radar, Sonar \& Navigation},
	publisher    = {Wiley Online Library},
	volume       = 10,
	number       = 2,
	pages        = {256--263}
}

@article{bachmann2005exploiting,
	title        = {Exploiting manifold geometry in hyperspectral imagery},
	author       = {Bachmann, Charles M and Ainsworth, Thomas L and Fusina, Robert A},
	year         = 2005,
	journal      = {IEEE Trans. Geosci. Remote Sens.},
	publisher    = {IEEE},
	volume       = 43,
	number       = 3,
	pages        = {441--454}
}

@article{harsanyi1994,
	title        = {Hyperspectral image classification and dimensionality reduction: an orthogonal subspace projection approach},
	author       = {Harsanyi, J.C. and Chang, C.-I.},
	year         = 1994,
	journal      = {IEEE Trans. Geosci. Remote Sens.},
	volume       = 32,
	number       = 4,
	pages        = {779--785},
	doi          = {10.1109/36.298007}
}

@article{chatterjee2000introduction,
	title        = {An introduction to the proper orthogonal decomposition},
	author       = {Chatterjee, Anindya},
	year         = 2000,
	journal      = {Current science},
	publisher    = {JSTOR},
	pages        = {808--817}
}

@article{yang2019survey,
	title        = {A survey on canonical correlation analysis},
	author       = {Yang, Xinghao and Liu, Weifeng and Liu, Wei and Tao, Dacheng},
	year         = 2019,
	journal      = {IEEE Trans. Knowl. Data Eng.},
	publisher    = {IEEE},
	volume       = 33,
	number       = 6,
	pages        = {2349--2368}
}

@article{xu2024texture,
	title        = {Texture-Aware Causal Feature Extraction Network for Multimodal Remote Sensing Data Classification},
	author       = {Xu, Zhengyi and Jiang, Wen and Geng, Jie},
	year         = 2024,
	journal      = {IEEE Trans. Geosci. Remote Sens.},
	publisher    = {IEEE}
}

@article{yan2006graph,
	title        = {Graph embedding and extensions: {A} general framework for dimensionality reduction},
	author       = {Yan, Shuicheng and Xu, Dong and Zhang, Benyu and Zhang, Hong-Jiang and Yang, Qiang and Lin, Stephen},
	year         = 2006,
	journal      = {IEEE Trans. Pattern Anal. Mach. Intell.},
	publisher    = {IEEE},
	volume       = 29,
	number       = 1,
	pages        = {40--51}
}

@article{qu2018udas,
	title        = {u{DAS}: An untied denoising autoencoder with sparsity for spectral unmixing},
	author       = {Qu, Ying and Qi, Hairong},
	year         = 2018,
	journal      = {IEEE Trans. Geosci. Remote Sens.},
	publisher    = {IEEE},
	volume       = 57,
	number       = 3,
	pages        = {1698--1712}
}

@article{nielsen2010kernel,
	title        = {Kernel maximum autocorrelation factor and minimum noise fraction transformations},
	author       = {Nielsen, Allan Aasbjerg},
	year         = 2010,
	journal      = {IEEE Trans. Image Process.},
	publisher    = {IEEE},
	volume       = 20,
	number       = 3,
	pages        = {612--624}
}

@article{luo2016minimum,
	title        = {Minimum noise fraction versus principal component analysis as a preprocessing step for hyperspectral imagery denoising},
	author       = {Luo, Guangchun and Chen, Guangyi and Tian, Ling and Qin, Ke and Qian, Shen-En},
	year         = 2016,
	journal      = {Canadian Journal of Remote Sensing},
	publisher    = {Taylor \& Francis},
	volume       = 42,
	number       = 2,
	pages        = {106--116}
}

@article{green1988transformation,
	title        = {A transformation for ordering multispectral data in terms of image quality with implications for noise removal},
	author       = {Green, Andrew A and Berman, Mark and Switzer, Paul and Craig, Maurice D},
	year         = 1988,
	journal      = {IEEE Trans. Geosci. Remote Sens.},
	publisher    = {IEEE},
	volume       = 26,
	number       = 1,
	pages        = {65--74}
}

@article{luo2017local,
	title        = {Local geometric structure feature for dimensionality reduction of hyperspectral imagery},
	author       = {Luo, Fulin and Huang, Hong and Duan, Yule and Liu, Jiamin and Liao, Yinghua},
	year         = 2017,
	journal      = {Remote Sens.},
	publisher    = {MDPI},
	volume       = 9,
	number       = 8,
	pages        = 790
}

@article{song2021learning,
	title        = {Learning to generate {SAR} images with adversarial autoencoder},
	author       = {Song, Qian and Xu, Feng and Zhu, Xiao Xiang and Jin, Ya-Qiu},
	year         = 2021,
	journal      = {IEEE Trans. Geosci. Remote Sens.},
	publisher    = {IEEE},
	volume       = 60,
	pages        = {1--15}
}

@article{rosipal2001kernel,
	title        = {Kernel partial least squares regression in reproducing kernel {H}ilbert space},
	author       = {Rosipal, Roman and Trejo, Leonard J},
	year         = 2001,
	journal      = {J. Mach. Learn. Res.},
	volume       = 2,
	number       = {Dec},
	pages        = {97--123}
}

@article{Nalepa2019,
	title        = {Training- and Test-Time Data Augmentation for Hyperspectral Image Segmentation},
	author       = {Nalepa, Jakub and Myller, Michał and Kawulok, Michal},
	year         = 2019,
	month        = {06},
	journal      = {IEEE Geosci. Remote Sens. Lett.},
	volume       = {PP},
	pages        = {1--5},
	doi          = {10.1109/LGRS.2019.2921011}
}

@article{klemmer2023satclip,
	title        = {Sat{CLIP}: {G}lobal, general-purpose location embeddings with satellite imagery},
	author       = {Klemmer, Konstantin and Rolf, Esther and Robinson, Caleb and Mackey, Lester and Ru{\ss}wurm, Marc},
	year         = 2023,
	journal      = {arXiv preprint arXiv:2311.17179}
}

@inproceedings{radford2021learning,
	title        = {Learning transferable visual models from natural language supervision},
	author       = {Radford, Alec and Kim, Jong Wook and Hallacy, Chris and Ramesh, Aditya and Goh, Gabriel and Agarwal, Sandhini and Sastry, Girish and Askell, Amanda and Mishkin, Pamela and Clark, Jack and others},
	year         = 2021,
	booktitle    = {Int. Conf. Mach. Learn.},
	pages        = {8748--8763},
	organization = {PMLR}
}

@article{li2011locality,
	title        = {Locality-preserving dimensionality reduction and classification for hyperspectral image analysis},
	author       = {Li, Wei and Prasad, Saurabh and Fowler, James E and Bruce, Lori Mann},
	year         = 2011,
	journal      = {IEEE Trans. Geosci. Remote Sens.},
	publisher    = {IEEE},
	volume       = 50,
	number       = 4,
	pages        = {1185--1198}
}

@article{benz1995comparison,
	title        = {A comparison of several algorithms for {SAR} raw data compression},
	author       = {Benz, Ursula and Strodl, Klaus and Moreira, Alberto},
	year         = 1995,
	journal      = {IEEE Trans. Geosci. Remote Sens.},
	publisher    = {IEEE},
	volume       = 33,
	number       = 5,
	pages        = {1266--1276}
}

@article{duan2021low,
	title        = {Low-complexity point cloud denoising for {L}i{DAR} by {PCA}-based dimension reduction},
	author       = {Duan, Yao and Yang, Chuanchuan and Chen, Hao and Yan, Weizhen and Li, Hongbin},
	year         = 2021,
	journal      = {Optics Comm.},
	publisher    = {Elsevier},
	volume       = 482,
	pages        = 126567
}

@article{lalitha2022review,
	title        = {A review on remote sensing imagery augmentation using deep learning},
	author       = {Lalitha, V and Latha, B},
	year         = 2022,
	journal      = {Mater. Today},
	publisher    = {Elsevier},
	volume       = 62,
	pages        = {4772--4778}
}

@article{hao2023review,
	title        = {A review of data augmentation methods of remote sensing image target recognition},
	author       = {Hao, Xuejie and Liu, Lu and Yang, Rongjin and Yin, Lizeyan and Zhang, Le and Li, Xiuhong},
	year         = 2023,
	journal      = {Remote Sens.},
	publisher    = {MDPI},
	volume       = 15,
	number       = 3,
	pages        = 827
}

@article{Song2019,
	title        = {Improved t-{SNE} based manifold dimensional reduction for remote sensing data processing},
	author       = {Song, Weijing and Wang, Lizhe and Liu, Peng and Choo, Kim-Kwang Raymond},
	year         = 2019,
	month        = {02},
	journal      = {Multimed. Tools Appl.},
	volume       = 78,
	pages        = {},
	doi          = {10.1007/s11042-018-5715-0}
}

@article{Najim2023,
	title        = {Insightful Visualization of Remote Sensing Images},
	author       = {Najim, Safa A. and Ahmed, Basaeir Y.},
	year         = 2023,
	journal      = {IEEE Geosci. Remote Sens. Lett.},
	volume       = 20,
	number       = {},
	pages        = {1--4},
	doi          = {10.1109/LGRS.2022.3228874}
}

@article{wang2012nonnegative,
	title        = {Nonnegative matrix factorization: {A} comprehensive review},
	author       = {Wang, Yu-Xiong and Zhang, Yu-Jin},
	year         = 2012,
	journal      = {IEEE Trans. Knowl. Data Eng.},
	publisher    = {IEEE},
	volume       = 25,
	number       = 6,
	pages        = {1336--1353}
}

@article{kaarna2000compression,
	title        = {Compression of multispectral remote sensing images using clustering and spectral reduction},
	author       = {Kaarna, Arto and Zemcik, Pavel and Kalviainen, Heikki and Parkkinen, Jussi},
	year         = 2000,
	journal      = {IEEE Trans. Geosci. Remote Sens.},
	publisher    = {IEEE},
	volume       = 38,
	number       = 2,
	pages        = {1073--1082}
}

@article{wang2006independent,
	title        = {Independent component analysis-based dimensionality reduction with applications in hyperspectral image analysis},
	author       = {Wang, Jing and Chang, Chein-I},
	year         = 2006,
	journal      = {IEEE Trans. Geosci. Remote Sens.},
	publisher    = {IEEE},
	volume       = 44,
	number       = 6,
	pages        = {1586--1600}
}

@article{xiang2024remote,
	title        = {Remote Sensing Image Compression Based on High-frequency and Low-frequency Components},
	author       = {Xiang, Shao and Liang, Qiaokang},
	year         = 2024,
	journal      = {IEEE Trans. Geosci. Remote Sens.},
	publisher    = {IEEE}
}

@article{penna2007transform,
	title        = {Transform coding techniques for lossy hyperspectral data compression},
	author       = {Penna, Barbara and Tillo, Tammam and Magli, Enrico and Olmo, Gabriella},
	year         = 2007,
	journal      = {IEEE Trans. Geosci. Remote Sens.},
	publisher    = {IEEE},
	volume       = 45,
	number       = 5,
	pages        = {1408--1421}
}

@article{penna2006progressive,
	title        = {Progressive 3-{D} coding of hyperspectral images based on {JPEG}2000},
	author       = {Penna, Barbara and Tillo, Tammam and Magli, Enrico and Olmo, Gabriella},
	year         = 2006,
	journal      = {IEEE Geosci. Remote Sens. Lett.},
	publisher    = {IEEE},
	volume       = 3,
	number       = 1,
	pages        = {125--129}
}

@article{garcia2019improved,
	title        = {Improved statistically based retrievals via spatial-spectral data compression for {IASI} data},
	author       = {Garcia-Sobrino, Joaquin and Laparra, Valero and Serra-Sagrist{\`a}, Joan and Calbet, Xavier and Camps-Valls, Gustau},
	year         = 2019,
	journal      = {IEEE Trans. Geosci. Remote Sens.},
	publisher    = {IEEE},
	volume       = 57,
	number       = 8,
	pages        = {5651--5668}
}

@article{Liu2021,
	title        = {Using t-distributed Stochastic Neighbor Embedding (t-SNE) for cluster analysis and spatial zone delineation of groundwater geochemistry data},
	author       = {Honghua Liu and Jing Yang and Ming Ye and Scott C. James and Zhonghua Tang and Jie Dong and Tongju Xing},
	year         = 2021,
	journal      = {Journal of Hydrology},
	volume       = 597,
	pages        = 126146,
	doi          = {https://doi.org/10.1016/j.jhydrol.2021.126146},
	issn         = {0022-1694},
	url          = {https://www.sciencedirect.com/science/article/pii/S0022169421001931}
}

@article{Canas1985,
	title        = {The generation and interpretation of false-colour composite principal component images},
	author       = {A. A. D. Canas and M. E. Barnett},
	year         = 1985,
	journal      = {Int. J. Remote Sens.},
	publisher    = {Taylor \& Francis},
	volume       = 6,
	number       = 6,
	pages        = {867--881},
	doi          = {10.1080/01431168508948510},
	url          = {https://doi.org/10.1080/01431168508948510},
	eprint       = {https://doi.org/10.1080/01431168508948510}
}

@article{Horel1981,
	title        = {Planetary-Scale Atmospheric Phenomena Associated with the {S}outhern {O}scillation},
	author       = {John D.  Horel and John M.  Wallace},
	year         = 1981,
	journal      = {Mon. Weather Rev.},
	publisher    = {American Meteorological Society},
	address      = {Boston MA, USA},
	volume       = 109,
	number       = 4,
	pages        = {813--829},
	doi          = {10.1175/1520-0493(1981)109<0813:PSAPAW>2.0.CO;2},
	url          = {https://journals.ametsoc.org/view/journals/mwre/109/4/1520-0493_1981_109_0813_psapaw_2_0_co_2.xml}
}

@article{Barnston1987,
	title        = {Classification, Seasonality and Persistence of Low-Frequency Atmospheric Circulation Patterns},
	author       = {Anthony G.  Barnston and Robert E.  Livezey},
	year         = 1987,
	journal      = {Mon. Weather Rev.},
	publisher    = {American Meteorological Society},
	address      = {Boston MA, USA},
	volume       = 115,
	number       = 6,
	pages        = {1083--1126},
	doi          = {10.1175/1520-0493(1987)115<1083:CSAPOL>2.0.CO;2},
	url          = {https://journals.ametsoc.org/view/journals/mwre/115/6/1520-0493_1987_115_1083_csapol_2_0_co_2.xml}
}

@article{Ibebuchi,
	title        = {Redefining the {N}orth {A}tlantic {O}scillation Index Generation using Autoencoder Neural Network},
	author       = {Ibebuchi, Chibuike},
	year         = 2024,
	month        = {01},
	journal      = {Mach. Learn.:-Sci. Technol.},
	pages        = {},
	doi          = {10.1088/2632-2153/ad1c32}
}

@article{rasti2021image,
	title        = {Image restoration for remote sensing: {O}verview and toolbox},
	author       = {Rasti, Behnood and Chang, Yi and Dalsasso, Emanuele and Denis, Loic and Ghamisi, Pedram},
	year         = 2021,
	journal      = {IEEE Geosci. Remote Sens. Mag.},
	publisher    = {IEEE},
	volume       = 10,
	number       = 2,
	pages        = {201--230}
}

@article{gross1993visualization,
	title        = {Visualization of multidimensional image data sets using a neural network},
	author       = {Gross, Markus H and Seibert, Frank},
	year         = 1993,
	journal      = {Vis. Comput.},
	publisher    = {Springer},
	volume       = 10,
	pages        = {145--159}
}

@article{tasdemir2009exploiting,
	title        = {Exploiting data topology in visualization and clustering of self-organizing maps},
	author       = {Tasdemir, Kadim and Mer{\'e}nyi, Erzs{\'e}bet},
	year         = 2009,
	journal      = {IEEE Trans. Neural Netw.},
	publisher    = {IEEE},
	volume       = 20,
	number       = 4,
	pages        = {549--562}
}

@article{zhang2020three,
	title        = {Three-dimensional convolutional neural network model for tree species classification using airborne hyperspectral images},
	author       = {Zhang, Bin and Zhao, Lin and Zhang, Xiaoli},
	year         = 2020,
	journal      = {Remote Sens. Environ.},
	publisher    = {Elsevier},
	volume       = 247,
	pages        = 111938
}

@article{haenlein2004beginner,
	title        = {A beginner's guide to partial least squares analysis},
	author       = {Haenlein, Michael and Kaplan, Andreas M},
	year         = 2004,
	journal      = {Underst. Stat.},
	publisher    = {Taylor \& Francis},
	volume       = 3,
	number       = 4,
	pages        = {283--297}
}

@article{akaho2006kernel,
	title        = {A kernel method for canonical correlation analysis},
	author       = {Akaho, Shotaro},
	year         = 2006,
	journal      = {arXiv preprint cs/0609071}
}

@article{Payandeh2023,
	title        = {Deep Representation Learning: Fundamentals, Technologies, Applications, and Open Challenges},
	author       = {Payandeh, Amirreza and Baghaei, Kourosh T. and Fayyazsanavi, Pooya and Ramezani, Somayeh Bakhtiari and Chen, Zhiqian and Rahimi, Shahram},
	year         = 2023,
	journal      = {IEEE Access},
	volume       = 11,
	number       = {},
	pages        = {137621--137659},
	doi          = {10.1109/ACCESS.2023.3335196}
}

@misc{zhao2025surveylargelanguagemodels,
	title        = {A Survey of Large Language Models},
	author       = {Wayne Xin Zhao and Kun Zhou and Junyi Li and Tianyi Tang and Xiaolei Wang and Yupeng Hou and Yingqian Min and Beichen Zhang and Junjie Zhang and Zican Dong and Yifan Du and Chen Yang and Yushuo Chen and Zhipeng Chen and Jinhao Jiang and Ruiyang Ren and Yifan Li and Xinyu Tang and Zikang Liu and Peiyu Liu and Jian-Yun Nie and Ji-Rong Wen},
	year         = 2023,
	url          = {https://arxiv.org/abs/2303.18223},
	eprint       = {2303.18223},
	archiveprefix = {arXiv},
	primaryclass = {cs.CL}
}

@article{Lu2025,
	title        = {Vision Foundation Models in Remote Sensing: A survey},
	author       = {Lu, Siqi and Guo, Junlin and Zimmer-Dauphinee, James R. and Nieusma, Jordan M. and Wang, Xiao and vanValkenburgh, Parker and Wernke, Steven A. and Huo, Yuankai},
	year         = 2025,
	journal      = {IEEE Geosci. Remote Sens. Mag.},
	volume       = {},
	number       = {},
	pages        = {2--27},
	doi          = {10.1109/MGRS.2025.3541952}
}

@misc{Szwarcman2025,
	title        = {Prithvi-EO-2.0: A Versatile Multi-Temporal Foundation Model for {E}arth Observation Applications},
	author       = {Daniela Szwarcman and Sujit Roy and Paolo Fraccaro and {\TH}orsteinn Elí Gíslason and Benedikt Blumenstiel and Rinki Ghosal and Pedro Henrique de Oliveira and Joao Lucas de Sousa Almeida and Rocco Sedona and Yanghui Kang and Srija Chakraborty and Sizhe Wang and Carlos Gomes and Ankur Kumar and Myscon Truong and Denys Godwin and Hyunho Lee and Chia-Yu Hsu and Ata Akbari Asanjan and Besart Mujeci and Disha Shidham and Trevor Keenan and Paulo Arevalo and Wenwen Li and Hamed Alemohammad and Pontus Olofsson and Christopher Hain and Robert Kennedy and Bianca Zadrozny and David Bell and Gabriele Cavallaro and Campbell Watson and Manil Maskey and Rahul Ramachandran and Juan Bernabe Moreno},
	year         = 2025,
	url          = {https://arxiv.org/abs/2412.02732},
	eprint       = {2412.02732},
	archiveprefix = {arXiv},
	primaryclass = {cs.CV}
}

@inproceedings{fuller2023croma,
	title        = {{CROMA}: Remote Sensing Representations with Contrastive Radar-Optical Masked Autoencoders},
	author       = {Anthony Fuller and Koreen Millard and James R Green},
	year         = 2023,
	booktitle    = {Neural Inf. Process. Syst.},
	url          = {https://openreview.net/forum?id=ezqI5WgGvY}
}

@misc{waldmann2025,
	title        = {Panopticon: Advancing Any-Sensor Foundation Models for {E}arth Observation},
	author       = {Leonard Waldmann and Ando Shah and Yi Wang and Nils Lehmann and Adam J. Stewart and Zhitong Xiong and Xiao Xiang Zhu and Stefan Bauer and John Chuang},
	year         = 2025,
	url          = {https://arxiv.org/abs/2503.10845},
	eprint       = {2503.10845},
	archiveprefix = {arXiv},
	primaryclass = {cs.LG}
}

@inproceedings{balestriero2022contrastive,
	title        = {Contrastive and Non-Contrastive Self-Supervised Learning Recover Global and Local Spectral Embedding Methods},
	author       = {Balestriero, Randall and LeCun, Yann},
    pages        = {26671--26685},
	year         = 2022,
	booktitle    = {Adv. Neural Inf. Process. Syst.}
}

@misc{ramospollan2024,
	title        = {Uncertainty and Generalizability in Foundation Models for {E}arth Observation},
	author       = {Raul Ramos-Pollan and Freddie Kalaitzis and Karthick Panner Selvam},
	year         = 2024,
	url          = {https://arxiv.org/abs/2409.08744},
	eprint       = {2409.08744},
	archiveprefix = {arXiv},
	primaryclass = {cs.CV}
}

@misc{czerkawski2024,
	title        = {Global and Dense Embeddings of {E}arth: {M}ajor {TOM} Floating in the Latent Space},
	author       = {Mikolaj Czerkawski and Marcin Kluczek and Jędrzej S. Bojanowski},
	year         = 2024,
	url          = {https://arxiv.org/abs/2412.05600},
	eprint       = {2412.05600},
	archiveprefix = {arXiv},
	primaryclass = {cs.CV}
}

@article{Wang2022,
	title        = {Self-Supervised Learning in Remote Sensing: A review},
	author       = {Wang, Yi and Albrecht, Conrad M. and Braham, Nassim Ait Ali and Mou, Lichao and Zhu, Xiao Xiang},
	year         = 2022,
	journal      = {IEEE Geosci. Remote Sens. Mag.},
	volume       = 10,
	number       = 4,
	pages        = {213--247},
	doi          = {10.1109/MGRS.2022.3198244}
}

@article{Kang2021,
	title        = {Deep Unsupervised Embedding for Remotely Sensed Images Based on Spatially Augmented Momentum Contrast},
	author       = {Kang, Jian and Fernandez-Beltran, Ruben and Duan, Puhong and Liu, Sicong and Plaza, Antonio J.},
	year         = 2021,
	journal      = {IEEE Trans. Geosci. Remote Sens.},
	volume       = 59,
	number       = 3,
	pages        = {2598--2610},
	doi          = {10.1109/TGRS.2020.3007029}
}

@inproceedings{Manas_2021_ICCV,
	title        = {Seasonal Contrast: Unsupervised Pre-Training From Uncurated Remote Sensing Data},
	author       = {Ma\~nas, Oscar and Lacoste, Alexandre and Gir\'o-i-Nieto, Xavier and Vazquez, David and Rodr{\'\i}guez, Pau},
	year         = 2021,
	month        = {October},
	booktitle    = {IEEE Int. Conf. Comput. Vis.},
	pages        = {9414--9423}
}

@inproceedings{Ayush2021,
	title        = {Geography-Aware Self-Supervised Learning},
	author       = {Ayush, Kumar and Uzkent, Burak and Meng, Chenlin and Tanmay, Kumar and Burke, Marshall and Lobell, David and Ermon, Stefano},
	year         = 2021,
	month        = {October},
	booktitle    = {IEEE Int. Conf. Comput. Vis.},
	pages        = {10181--10190}
}

@misc{Marsocci2025,
	title        = {{PANGAEA}: A Global and Inclusive Benchmark for Geospatial Foundation Models},
	author       = {Valerio Marsocci and Yuru Jia and Georges Le Bellier and David Kerekes and Liang Zeng and Sebastian Hafner and Sebastian Gerard and Eric Brune and Ritu Yadav and Ali Shibli and Heng Fang and Yifang Ban and Maarten Vergauwen and Nicolas Audebert and Andrea Nascetti},
	year         = 2025,
	url          = {https://arxiv.org/abs/2412.04204},
	eprint       = {2412.04204},
	archiveprefix = {arXiv},
	primaryclass = {cs.CV}
}

@article{Wang2024,
	title        = {RSID-CR: {R}emote Sensing Image Denoising Based on Contrastive Learning},
	author       = {Wang, Zhibao and He, Xiaoqing and Xiao, Bin and Chen, Liangfu and Bi, Xiuli},
	year         = 2024,
	journal      = {IEEE J. Sel. Top. Appl. Earth Obs. Remote Sens.},
	volume       = 17,
	number       = {},
	pages        = {18784--18799},
	doi          = {10.1109/JSTARS.2024.3476566}
}

@misc{gupta2025mosaicmultimodalmultilabelsupervisionaware,
	title        = {MoSAiC: Multi-Modal Multi-Label Supervision-Aware Contrastive Learning for Remote Sensing},
	author       = {Debashis Gupta and Aditi Golder and Rongkhun Zhu and Kangning Cui and Wei Tang and Fan Yang and Ovidiu Csillik and Sarra Alaqahtani and V. Paul Pauca},
	year         = 2025,
	url          = {https://arxiv.org/abs/2507.08683},
	eprint       = {2507.08683},
	archiveprefix = {arXiv},
	primaryclass = {cs.CV}
}

@inproceedings{bohm2023unsupervised,
	title        = {Unsupervised visualization of image datasets using contrastive learning},
	author       = {Niklas B{\"o}hm and Philipp Berens and Dmitry Kobak},
	year         = 2023,
	booktitle    = {The Eleventh Int. Conf. Learn. Represent.}
}

@article{Garcia-Vilchez2011253,
	title        = {{On the impact of lossy compression on hyperspectral image classification and unmixing}},
	author       = {Garc\'ia-V\'ilchez, F. and Mu\~{n}oz-Mar\'i, J. and Zortea, M. and Blanes, I. and Gonz\'alez-Ruiz, V. and Camps-Valls, G. and Plaza, A. and Serra-Sagrist\`{a}, J.},
	year         = 2011,
	journal      = {IEEE Geosci. Remote Sens. Lett.},
	publisher    = {IEEE},
	volume       = 8,
	number       = 2,
	pages        = {253--257},
	doi          = {10.1109/LGRS.2010.2062484}
}

@article{buhlmann2020invariance,
	title        = {Invariance, causality and robustness},
	author       = {B{\"u}hlmann, Peter},
	year         = 2020,
	journal      = {Stat. Sci.},
	publisher    = {JSTOR},
	volume       = 35,
	number       = 3,
	pages        = {404--426}
}

@misc{tseng2024lightweightpretrainedtransformersremote,
	title        = {Lightweight, Pre-trained Transformers for Remote Sensing Timeseries},
	author       = {Gabriel Tseng and Ruben Cartuyvels and Ivan Zvonkov and Mirali Purohit and David Rolnick and Hannah Kerner},
	year         = 2024,
	url          = {https://arxiv.org/abs/2304.14065},
	eprint       = {2304.14065},
	archiveprefix = {arXiv},
	primaryclass = {cs.CV}
}

@inproceedings{Chen2020,
  title={A Simple Framework for Contrastive Learning of Visual Representations},
  author={Chen, Ting and Kornblith, Simon and Norouzi, Mohammad and Hinton, Geoffrey},
  booktitle={Int. Conf. Mach. Learn.},
  pages={1597--1607},
  year={2020},
  organization={PmLR}
}

@article{Awais2025,
	title        = {Foundation Models Defining a New Era in Vision: A Survey and Outlook},
	author       = {Awais, Muhammad and Naseer, Muzammal and Khan, Salman and Anwer, Rao Muhammad and Cholakkal, Hisham and Shah, Mubarak and Yang, Ming-Hsuan and Khan, Fahad Shahbaz},
	year         = 2025,
	journal      = {IEEE Trans. Pattern Anal. Mach. Intell.},
	volume       = 47,
	number       = 4,
	pages        = {2245--2264},
	doi          = {10.1109/TPAMI.2024.3506283}
}

@misc{zhu2024foundationsearthclimatefoundation,
	title        = {On the Foundations of {E}arth and Climate Foundation Models},
	author       = {Xiao Xiang Zhu and Zhitong Xiong and Yi Wang and Adam J. Stewart and Konrad Heidler and Yuanyuan Wang and Zhenghang Yuan and Thomas Dujardin and Qingsong Xu and Yilei Shi},
	year         = 2024,
	url          = {https://arxiv.org/abs/2405.04285},
	eprint       = {2405.04285},
	archiveprefix = {arXiv},
	primaryclass = {cs.AI}
}

@misc{xiong2024,
	title        = {Neural Plasticity-Inspired Multimodal Foundation Model for {E}arth Observation},
	author       = {Zhitong Xiong and Yi Wang and Fahong Zhang and Adam J. Stewart and Joëlle Hanna and Damian Borth and Ioannis Papoutsis and Bertrand Le Saux and Gustau Camps-Valls and Xiao Xiang Zhu},
	year         = 2024,
	url          = {https://arxiv.org/abs/2403.15356},
	eprint       = {2403.15356},
	archiveprefix = {arXiv},
	primaryclass = {cs.CV}
}

@misc{jakubik2025terramindlargescalegenerativemultimodality,
	title        = {{T}erra{M}ind: {L}arge-Scale Generative Multimodality for {E}arth Observation},
	author       = {Johannes Jakubik and Felix Yang and Benedikt Blumenstiel and Erik Scheurer and Rocco Sedona and Stefano Maurogiovanni and Jente Bosmans and Nikolaos Dionelis and Valerio Marsocci and Niklas Kopp and Rahul Ramachandran and Paolo Fraccaro and Thomas Brunschwiler and Gabriele Cavallaro and Juan Bernabe-Moreno and Nicolas Longépé},
	year         = 2025,
	url          = {https://arxiv.org/abs/2504.11171},
	eprint       = {2504.11171},
	archiveprefix = {arXiv},
	primaryclass = {cs.CV}
}

@misc{SHRUG-FM,
	title        = {SHRUG-FM: Reliability-Aware Foundation Models for Earth Observation},
	author       = {Kai-Hendrik Cohrs and Zuzanna Osika and Maria Gonzalez-Calabuig and Vishal Nedungadi and Ruben Cartuyvels and Steffen Knoblauch and Joppe Massant and Shruti Nath and Patrick Ebel and Vasileios Sitokonstantinou},
	year         = 2025,
	url          = {https://arxiv.org/abs/2511.10370},
	eprint       = {2511.10370},
	archiveprefix = {arXiv},
	primaryclass = {cs.CV}
}

@inproceedings{rezende2015variational,
  title={Variational inference with normalizing flows},
  author={Rezende, Danilo and Mohamed, Shakir},
  booktitle={International conference on machine learning},
  pages={1530--1538},
  year={2015},
  organization={PMLR}
}

@article{goodfellow2020generative,
  title={Generative adversarial networks},
  author={Goodfellow, Ian and Pouget-Abadie, Jean and Mirza, Mehdi and Xu, Bing and Warde-Farley, David and Ozair, Sherjil and Courville, Aaron and Bengio, Yoshua},
  journal={Communications of the {ACM}},
  volume={63},
  number={11},
  pages={139--144},
  year={2020},
  publisher={ACM New York, NY, USA}
}

@inproceedings{astruc2025anysat,
  title={Anysat: {O}ne earth observation model for many resolutions, scales, and modalities},
  author={Astruc, Guillaume and Gonthier, Nicolas and Mallet, Clement and Landrieu, Loic},
  booktitle={IEEE Conf. Comput. Vis. Pattern Recognit.},
  pages={19530--19540},
  year={2025}
}

@inproceedings{wang2025unifiedcopernicusfoundationmodel,
  title={Towards a Unified Copernicus Foundation Model for Earth Vision},
  author={Yi Wang and Zhitong Xiong and Chenying Liu and Adam J. Stewart and Thomas Dujardin and Nikolaos Ioannis Bountos and Angelos Zavras and Franziska Gerken and Ioannis Papoutsis and Laura Leal-Taixé and Xiao Xiang Zhu},
  booktitle={IEEE Int. Conf. Comput. Vis.},
  year={2025}
}

@article{kipf2016variational,
  title={Variational graph auto-encoders},
  author={Kipf, Thomas N and Welling, Max},
  journal={arXiv preprint arXiv:1611.07308},
  year={2016}
}

@article{fan2021heterogeneous,
  title={Heterogeneous hypergraph variational autoencoder for link prediction},
  author={Fan, Haoyi and Zhang, Fengbin and Wei, Yuxuan and Li, Zuoyong and Zou, Changqing and Gao, Yue and Dai, Qionghai},
  journal={IEEE transactions on pattern analysis and machine intelligence},
  volume={44},
  number={8},
  pages={4125--4138},
  year={2021},
  publisher={IEEE}
}

@article{keshava2002spectral,
  title={Spectral unmixing},
  author={Keshava, Nirmal and Mustard, John F},
  journal={IEEE signal processing magazine},
  volume={19},
  number={1},
  pages={44--57},
  year={2002},
  publisher={IEEE}
}

@article{papamarkou2024position,
  title={Position: {T}opological deep learning is the new frontier for relational learning},
  author={Papamarkou, Theodore and Birdal, Tolga and Bronstein, Michael and Carlsson, Gunnar and Curry, Justin and Gao, Yue and Hajij, Mustafa and Kwitt, Roland and Lio, Pietro and Di Lorenzo, Paolo and others},
  journal={Proceedings of machine learning research},
  volume={235},
  pages={39529},
  year={2024}
}

@article{karniadakis2021physics,
  title={Physics-informed machine learning},
  author={Karniadakis, George Em and Kevrekidis, Ioannis G and Lu, Lu and Perdikaris, Paris and Wang, Sifan and Yang, Liu},
  journal={Nature Reviews Physics},
  volume={3},
  number={6},
  pages={422--440},
  year={2021},
  publisher={Nature Publishing Group UK London}
}

@article{camps2026ai,
  title={{AI} needs a new philosophy of science},
  author={Camps-Valls, Gustau},
  journal={The Innovation},
  pages={101311},
  year={2026},
  publisher={Elsevier}
}

@article{hohl2024opening,
  title={Opening the black box: {A} systematic review on explainable artificial intelligence in remote sensing},
  author={H{\"o}hl, Adrian and Obadic, Ivica and Fernandez-Torres, Miguel-Angel and Najjar, Hiba and Oliveira, Dario Augusto Borges and Akata, Zeynep and Dengel, Andreas and Zhu, Xiao Xiang},
  journal={IEEE Geoscience and Remote Sensing Magazine},
  volume={12},
  number={4},
  pages={261--304},
  year={2024},
  publisher={IEEE}
}

@article{ince2020superpixel,
  title={Superpixel-based graph {L}aplacian regularization for sparse hyperspectral unmixing},
  author={Ince, Taner},
  journal={IEEE Geoscience and Remote Sensing Letters},
  volume={19},
  pages={1--5},
  year={2020},
  publisher={IEEE}
}
